\documentclass[final,3p,times]{elsarticle}
\usepackage{graphicx}
\usepackage[caption=false]{subfig}
\usepackage{amsmath,bm}
\usepackage{amsthm,amsmath,amssymb} \usepackage{mathrsfs}
\usepackage{url}
\usepackage{lineno}
\usepackage{color}
\usepackage{algorithmic}
\usepackage{algorithm}
\usepackage{diagbox}
\usepackage{makecell}
\usepackage{booktabs}
\usepackage{multirow}
\usepackage[figuresright]{rotating}
\usepackage[normalem]{ulem} 
\usepackage{siunitx} 
\usepackage{xcolor}
\usepackage{colortbl} 
\usepackage{hyperref}

\definecolor{bestblue}{RGB}{0, 119, 187}
\definecolor{rowgray}{gray}{0.92} 
\definecolor{impgreen}{RGB}{0, 150, 0} 

\newcommand{\best}[1]{\textbf{\tablenum[table-format=1.2e-1]{#1}}}
\newcommand{\second}[1]{\color{bestblue}\bfseries\tablenum[table-format=1.2e-1]{#1}}
\newcommand{\third}[1]{\uline{\tablenum[table-format=1.2e-1]{#1}}}

\newcommand{\imp}[1]{\multicolumn{1}{c}{\cellcolor{impgreen!10}\bfseries\color{impgreen!80!black}$\uparrow$#1\%}}

\journal{}

\begin{document}

\begin{frontmatter}

\title{From Complex Dynamics to DynFormer: Rethinking Transformers for PDEs}

\author[aff1]{Pengyu Lai}
\author[aff1]{Yixiao Chen}
\author[aff1]{Dewu Yang}
\author[aff1]{Rui Wang}
\author[aff1]{Feng Wang}
\author[aff1]{Hui Xu\corref{cor1}}

\cortext[cor1]{Corresponding author: dr.hxu@sjtu.edu.cn}

\affiliation[aff1]{%
organization={School of Aeronautics and Astronautics, Shanghai Jiao Tong University},
city={Shanghai},
postcode={200240},
country={China}
}

\begin{abstract}
Partial differential equations (PDEs) are fundamental for modeling complex physical systems, yet classical numerical solvers face prohibitive computational costs in high-dimensional and multi-scale regimes. 
While Transformer-based neural operators have emerged as powerful data-driven alternatives, they conventionally treat all discretized spatial points as uniform, independent tokens. This monolithic approach ignores the intrinsic scale separation of physical fields, applying computationally prohibitive $\mathcal{O}(N^4)$ global attention that redundantly mixes smooth large-scale dynamics with high-frequency fluctuations. Rethinking Transformers through the lens of complex dynamics, we propose DynFormer, a novel dynamics-informed neural operator. Rather than applying a uniform attention mechanism across all scales, DynFormer explicitly assigns specialized network modules to distinct physical scales. It leverages a Spectral Embedding to isolate low-frequency modes, enabling a Kronecker-structured attention mechanism to efficiently capture large-scale global interactions with reduced $\mathcal{O}(N^3)$ complexity. Concurrently, guided by the slaving principle of complex dynamics, we introduce a Local-Global-Mixing (LGM) transformation. This module utilizes nonlinear multiplicative frequency mixing to implicitly reconstruct the small-scale, fast-varying turbulent cascades that are slaved to the macroscopic state, without incurring the cost of global attention.
Integrating these modules into a hybrid evolutionary architecture ensures robust long-term temporal stability. Extensive memory-aligned evaluations across chaotic, elliptic, parabolic, and hyperbolic PDE benchmarks demonstrate that DynFormer achieves up to a 95\% reduction in relative error compared to state-of-the-art baselines, while significantly reducing GPU memory consumption. Our results establish that embedding first-principles physical dynamics into Transformer architectures yields a highly scalable, theoretically grounded blueprint for PDE surrogate modeling.
\end{abstract}


\end{frontmatter}

\section{Introduction}
PDEs play a fundamental role in modeling a wide range of natural and engineered phenomena \cite{evans2022partial, farlow1993partial}, including fluid dynamics\cite{landau1987fluid}, heat transfer \cite{pletcher2012computational}, electromagnetism \cite{griffiths2023introduction}, and quantum mechanics \cite{griffiths2023introduction}. Solving PDEs is essential for understanding the underlying physical laws and for predicting the behavior of complex systems under various conditions. Analytical solutions, when available, provide exact descriptions of system dynamics; however, most real-world problems involve nonlinearities, irregular domains, or intricate boundary conditions that preclude closed-form solutions. Consequently, numerical and computational methods for solving PDEs have become indispensable across science and engineering. 

While classical discretization-based techniques such as finite difference \cite{smith1985numerical}, finite element \cite{zienkiewicz1971finite}, and finite volume \cite{versteeg2007introduction} methods—has achieved remarkable success, they face inherent challenges in high-dimensional, strongly nonlinear, and multi-scale regimes. Grid resolution requirements rise sharply when capturing localized features, discontinuities, or thin boundary layers, driving up both memory footprint and wall-clock cost. The resulting computational burden is exacerbated in parametric, inverse, or design-loop settings where repeated solves are required. These limitations have spurred intense interest in data-driven surrogates for PDEs.

Data-driven neural approximators provide one such alternative. A broad range of architectures have been explored, including fully connected neural networks (FCNNs) \cite{psichogios1992hybrid}, long short-term memory networks (LSTMs) \cite{xuan2023reconstruction}, generative adversarial networks \cite{kim2024prediction} and autoencoder \cite{shi2015convolutional}. To incorporate the prior physical knowledge to facilitate the training and generalization, physics-informed neural networks (PINNs) were proposed, leveraging automatic differentiation to efficiently compute derivatives and enforce PDE constraints \cite{raissi2019physics, pang2019fpinns, lu2021physics}. 
By combining the expressiveness of deep learning with physical priors, these approaches offer flexibility, improved data efficiency, and in some cases real-time inference, making them an attractive complement or even replacement for classical numerical solvers in many complex scenarios \cite{cho2023separable, raissi2018hidden}.

Despite their promise, many classical neural formulations remain problem-specific, demanding re-implementation when dealing with adjustable parameters, irregular geometries or evolving domains. These limitations motivated the presence of the neural operator approximating the operator mapping between function spaces \cite{kovachki2023neural, lu2021learning}. Neural operator families now underpin a growing range of applications including weather forecasting
 \cite{bonev2023spherical, lam2023learning, kontolati2024learning, bonev2025fourcastnet}, seismology \cite{sun2023phase, shi2024broadband, haghighat2024deeponet},  turbomachinery \cite{li2025attention} and addictive manufacturing \cite{liu2024deep}. Methodological advances continue to broaden capability: contrastive learning for PDE generalization \cite{zhang2024deciphering, jiang2023training} ; transfer learning for heterogeneous task adaptation \cite{goswami2022deep, lyu2023multi}; in-context learning for out-of-distribution generalization \cite{yang2023context}; prior physics knowledge incorporation for network architectures \cite{long2018pde, long2019pde} and loss functions \cite{jiang2023training, li2024physics}; operator learning on manifolds and curved geometries \cite{seidman2022nomad, bonev2023spherical, li2023fourier}; and generalized spectral kernels beyond Fourier bases \cite{cao2024laplace, gupta2021multiwavelet, tripura2023wavelet, li2020neural}. 

Given the effectiveness of neural operators, researchers have explored diverse architectural backbones within the operator-learning paradigm, including U-Net architecture \cite{wen2022u}, large-kernel convolutional designs \cite{raonic2023convolutional}, graph-based PDE formers for unstructured meshes \cite{ye2024pdeformer}, and attention/transformer-based models \cite{cao2021choose, ye2024pdeformer}. Especially, transformers are particularly appealing, because their global receptive field and flexible tokenization align naturally with discretized function evaluations. Their success in sequence modeling and large language models has catalyzed rapid cross-fertilization with scientific machine learning. The Operator Transformer (OFormer) employs stacked self- and cross-attention blocks \cite{li2022transformer}. The General Neural Operator Transformer (GNOT) extends this idea with heterogeneous normalized attention and geometric gating \cite{hao2023gnot}. Vision-inspired hybrids also appear: the Vision Transformer-Operator (ViTO) embeds a ViT module within a U-Net \cite{ovadia2024vito}, while the Continuous Vision Transformer Operator (CViT) adapts ViT-style encoders with grid-aware positional embeddings and query-wise cross-attention \cite{dosovitskiy2020image}. 
In parallel, large pretrained operator models for PDEs are emerging \cite{chen2024data, subramanian2023towards, mccabe2024multiple, herde2024poseidon}, further amplifying the need for scalable and efficient transformer backbones that can serve as shared surrogates across tasks, resolutions, and physical regimes.

Despite this progress, transformer-based neural operators face practical barriers. This monolithic, "one-size-fits-all" attention mechanism ignores the inherent geometric and physical priors of PDE systems, indiscriminately mixing smooth large-scale flows with highly oscillatory small-scale turbulence,
incurring quadratic space–time complexity in the number of tokens and driving GPU memory usage to prohibitive levels on fine grids of dynamics. Even approximate or nominally linear-complexity attention variants can become numerically fragile or computationally heavy at high resolution. Consequently, efficiency has become a central research theme. Representative efforts include axial factorization of attention along coordinate directions in the Factorized Transformer (FactFormer) \cite{li2023scalable}; orthogonality-regularized attention pathways in the Orthogonal Neural Operator (ONO) \cite{xiao2023improved}; and low-rank physics-aware tokenization in the fast transformer solver (Transolver) \cite{wu2024transolver}. 

To break this computational and conceptual bottleneck, we draw inspiration from the hierarchical energy cascades and the slaving principle inherent in complex dynamical systems. Many high-dimensional physical systems exhibit scale-dependent coupling between large- and small-scale features \cite{kraichnan1965inertial}. Large-scale components are often smoother and lower-dimensional, whereas small-scale fluctuations can be intermittent yet strongly slaved to the coarse state in certain regimes \cite{carr2012applications}. This suggests that if a model can efficiently capture correlations in the large-scale dynamics, it may be possible to reconstruct, or at least approximate, relevant small-scale behavior from that coarse representation. Such ideas underpin long traditions in turbulence modeling and reduced-order modeling of multi-scale systems. They also resonate with emerging evidence in operator learning that fine-scale structure can, in part, be inferred from coarsened inputs \cite{lai2024neural}.

Guided by this physical insight, we propose DynFormer, which fundamentally rethinks the Transformer for PDEs. Rather than forcing a single attention mechanism to process all scales, DynFormer follows a simple but powerful philosophy: the neural architecture should mirror how nature evolves. We explicitly abandon uniform tokenization and establish a framework where network modules are strictly bound to physical scales.
Our main contributions are summarized as follows:
\begin{itemize}
    \item \textbf{Rethinking Transformers via the Slaving Principle.} We expose the inefficiency of treating physical fields as uniform tokens. By bridging neural architecture with the slaving principle of complex dynamics, we establish a novel framework that fundamentally shifts from monolithic attention to a scale-specialized processing paradigm.
    \item \textbf{Large-Scale Dynamics via Spectral Separation and Kronecker Attention} We design a spectral embedding mechanism to truncate high frequencies and explicitly isolate the large-scale, low-frequency modes. On this highly predictable latent manifold, we propose a Kronecker-structured attention mechanism that leverages axis-wise factorization, drastically reducing spatial complexity from $\mathcal{O}(N^4)$ to $\mathcal{O}(N^3)$ while preserving long-range physical coupling.
    \item \textbf{Small-Scale Reconstruction via LGM.} Because small-scale dynamics are slaved to macroscopic states, we bypass expensive full-grid attention for high frequencies. Instead, we introduce the LGM transformation, which uses multiplicative frequency mixing—analogous to nonlinear convective transport—to implicitly reconstruct the sub-grid turbulent cascades discarded during initial spectral truncation.
    \item \textbf{Comprehensive Evaluation on Multi-Scale PDE Benchmarks.} Through extensive experiments on challenging 1D to 3D multi-scale PDE systems, DynFormer achieves state-of-the-art predictive accuracy while operating at a fraction of the GPU memory footprint of existing transformer-based operators.
\end{itemize}

\begin{figure}[htbp]
  \centering
  \includegraphics[width=0.9\linewidth]{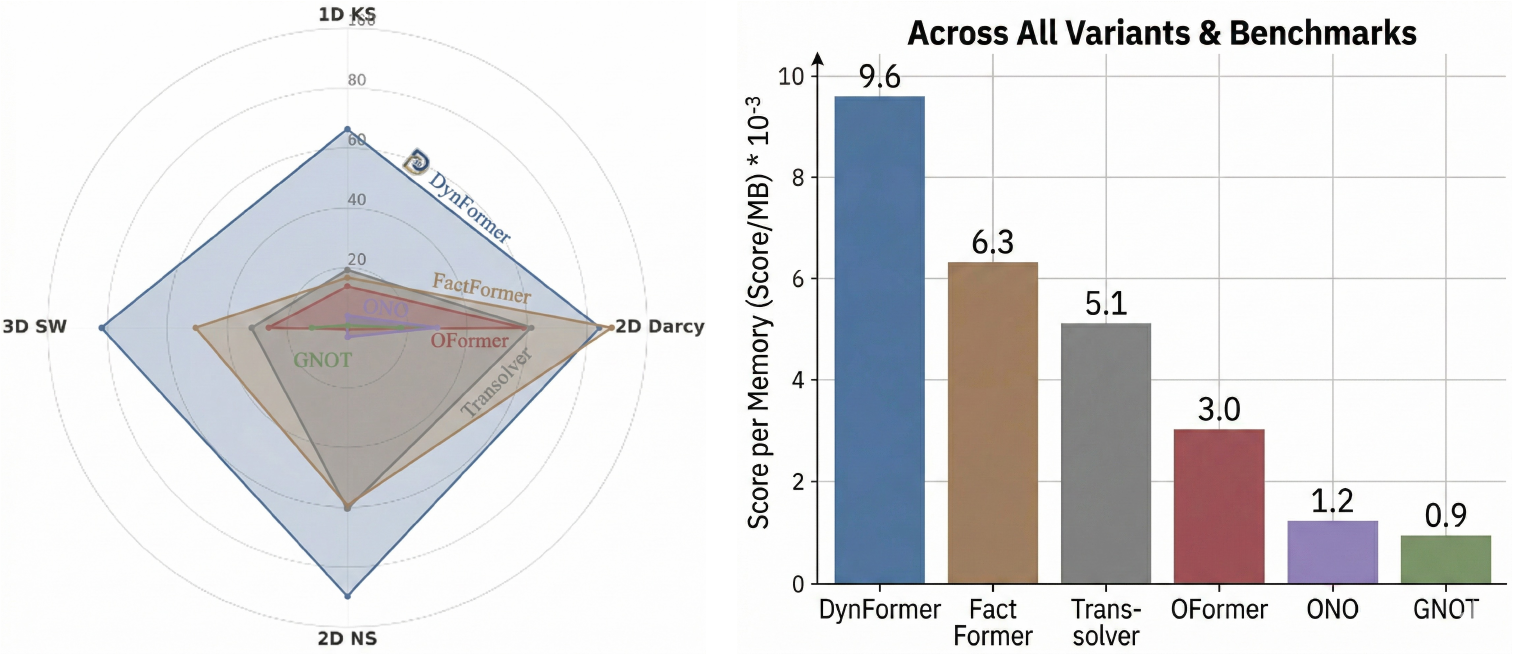}
  \caption{Overview of DynFormer's performance across four PDE benchmarks and baselines' peak GPU memory consumption.}
  \label{fig:radar_overview}
\end{figure}

\section{Methodology}\label{Methodology}

\subsection{Scale Decomposition of Complex Dynamics}\label{sec:scale_decomposition}
Unlike standard vision or language transformers that process all spatial tokens as uniform, independent entities, physical fields governed by PDEs exhibit distinct, hierarchically structured multiscale behaviors. To effectively capture this and "rethink" the transformer for physical systems, we begin with a formal decomposition of the dynamics into large-scale and small-scale components. This physically grounded decomposition provides the theoretical foundation for our architecture, allowing us to assign global attention mechanisms strictly to large-scale correlations while handling local nonlinear interactions separately.

\paragraph{Large-Scale and Small-Scale Dynamics}
Consider a general nonlinear dynamical system governed by the evolution equation:
\begin{equation}
\frac{\partial u(t)}{\partial t} = \mathscr{A}u(t) + f,
\label{eq:general_dynamics}
\end{equation}
where $u(t) \in \mathscr{X}$ represents the system state at time $t$, $\mathscr{A} = \mathscr{L} + \mathscr{N}$ denotes a composite differential operator composed of linear ($\mathscr{L}$) and nonlinear ($\mathscr{N}$) components, and $f$ is an external forcing term. The solution space $\mathscr{X}$ is assumed to be a Hilbert space equipped with an appropriate inner product structure.

Following the nonlinear Galerkin methodology \cite{foias1988inertial, marion1989nonlinear}, we introduce a functional space decomposition $\mathscr{X} = \mathscr{X}^M \oplus \mathscr{X}^{M^\perp}$, where $\mathscr{X}^M$ represents an $M$-dimensional coarse-scale subspace and $\mathscr{X}^{M^\perp}$ denotes its orthogonal fine-scale complement. Let $\mathscr{P}_M: \mathscr{X} \rightarrow \mathscr{X}^M$ and $\mathscr{Q}_M: \mathscr{X} \rightarrow \mathscr{X}^{M^\perp}$ be the corresponding projection operators satisfying $\mathscr{P}_M + \mathscr{Q}_M = \mathcal{I}$ and $\mathscr{P}_M \mathscr{Q}_M = \mathscr{Q}_M \mathscr{P}_M = 0$. The system dynamics can then be decomposed as:
\begin{subequations}
\label{eq:decomposed_dynamics}
\begin{align}
\frac{\partial p_m(t)}{\partial t} &= \mathscr{L}p_m(t) + \mathscr{P}_M\mathscr{N}\big(p_m(t) + q_m(t)\big) + \mathscr{P}_M f, \label{eq:large_scale} \\
\frac{\partial q_m(t)}{\partial t} &= \mathscr{L}q_m(t) + \mathscr{Q}_M\mathscr{N}\big(p_m(t) + q_m(t)\big) + \mathscr{Q}_M f, \label{eq:small_scale}
\end{align}
\end{subequations}
where $p_m(t) = \mathscr{P}_M u(t) \in \mathscr{X}^M$ and $q_m(t) = \mathscr{Q}_M u(t) \in \mathscr{X}^{M^\perp}$ represent the large-scale and small-scale components, respectively.

\paragraph{Mapping from Large- to Small-Scale Dynamics}
This decomposition mathematically aligns with the energy cascade observed in complex dynamics and renormalization group theory \cite{chen1994}, where $p_m(t)$ and $q_m(t)$ correspond to slow carriers and fast fluctuations, respectively. Due to the fast-relaxation nature of small-scale dynamics and the typically smaller magnitude of $q_m$ compared to $p_m$, we can neglect higher-order nonlinearities involving $q_m$ under the quasi-steady-state assumption. Equation \eqref{eq:small_scale} then reduces to:
\begin{equation}
\mathscr{L}q_m(t) + \mathscr{Q}_M \mathscr{N}p_m(t) + \mathscr{Q}_M f \approx 0,
\label{eq:quasi_steady}
\end{equation}
yielding an approximate mapping:
\begin{equation}
q_m(t) \approx \mathscr{L}^{-1}\Big(-\mathscr{Q}_M \mathscr{N}p_m(t) - \mathscr{Q}_M f\Big).
\label{eq:small_scale_mapping}
\end{equation}
This approximation implies the existence of a graph mapping $\Phi: \mathscr{X}^M \rightarrow \mathscr{X}^{M^\perp}$ such that $q_m(t) = \Phi\big(p_m(t)\big)$. This theoretical insight is the exact foundation of our architectural "rethink": because $q_m(t)$ is slaved to $p_m(t)$, we absolutely do not need to subject the high-frequency residuals to the $\mathcal{O}(N^4)$ global attention mechanism. Instead, the small-scale dynamics can be efficiently and implicitly reconstructed from the processed large-scale representation via learned nonlinear transformations (realized subsequently through the LGM transformation).

\paragraph{Scale decomposition via Fourier Modes}
To realize this continuous decomposition practically within a neural architecture, we design a \emph{Spectral Dynamics Embedding} based on Fourier decomposition. This offers both computational efficiency (via FFT with $\mathcal{O}(N \log N)$ complexity) and physical interpretability. Given a discretized field $u \in \mathbb{R}^{N_1 \times N_2 \times \cdots \times N_d \times d_{\text{in}}}$ over a $d$-dimensional spatial domain, each spatial point $u_{\mathbf{i}}$ can be projected onto a finite-dimensional Fourier space:
\begin{equation}
u_{\mathbf{i}} = \sum_{\|\mathbf{k}\| \leq N_{\max}} \hat{u}_{\mathbf{i},\mathbf{k}} \exp\left(\imath \frac{2\pi \mathbf{k} \cdot \mathbf{i}}{N}\right),
\label{eq:fourier_expansion}
\end{equation}
where $\mathbf{k}$ denotes the frequency index, $N$ represents the grid resolution, and $\hat{u}_{\mathbf{i},\mathbf{k}}$ are the Fourier coefficients.
The corresponding projections onto large-scale and small-scale subspaces are defined as:
\begin{align}
{p_m}_{\mathbf{i}} &= \mathcal{F}^{-1}\big[\mathscr{P}_M \mathcal{F}(u_{\mathbf{i}})\big] = \sum_{\|\mathbf{k}\| \leq M} \hat{u}_{\mathbf{i},\mathbf{k}} \exp\left(\imath \frac{2\pi \mathbf{k} \cdot \mathbf{i}}{N}\right) \in \mathscr{X}^{M}, \label{eq:large_scale_projection} \\
{q_m}_{\mathbf{i}} &= \mathcal{F}^{-1}\big[\mathscr{Q}_M \mathcal{F}(u_{\mathbf{i}})\big] = \sum_{M < \|\mathbf{k}\| \leq N_{\max}} \hat{u}_{\mathbf{i},\mathbf{k}} \exp\left(\imath \frac{2\pi \mathbf{k} \cdot \mathbf{i}}{N}\right) \in \mathscr{X}^{M^\perp}, \label{eq:small_scale_projection}
\end{align}
where $\mathcal{F}$ and $\mathcal{F}^{-1}$ denote the forward and inverse Fourier transforms. The truncation operator $\mathscr{P}_M$ acts as an architectural filter, retaining only the lowest $M$ Fourier modes to isolate $p_m$. 

By isolating the low-dimensional, long-range features ($p_m$), we construct an optimal latent space where global attention mechanisms can operate efficiently without being overwhelmed by the high-frequency residuals ($q_m$). This physically motivated spectral separation directly enables the $\mathcal{O}(N^3)$ attention scheme introduced in the following section.

\subsection{Modeling Large-Scale Interaction via Kronecker-Structured Attention}\label{sec:kronecker_attention}
Having established the spectral isolation of the large-scale component $p_m$ in Section \ref{sec:scale_decomposition}, we now focus on efficiently modeling its evolution. Understanding large-scale dynamics in complex systems requires capturing long-range dependencies and global interactions. Attention mechanisms naturally support such global modeling; however, standard implementations flatten spatial grids into 1D sequences, destroying geometric priors and incurring quadratic complexity. 
Crucially, large-scale physical components are typically smooth and dominated by low-frequency modes. In classical PDE theory, such smooth fields on orthogonal domains frequently admit a \emph{separation of variables} \cite{evans2022partial}, implying their interaction kernels possess a \emph{structured low-rank representation} along coordinate axes. 

Guided by this physical property, we assume the latent space projected by $\mathbb{R}^{N_1 \times N_2 \times d_{\text{in}}} \rightarrow \mathbb{R}^{N_1 \times N_2 \times d_n}$ is separable and thus propse a \textit{Kronecker-structured attention} backbone that leverages axis-wise separability. This drastically reduces computational cost while preserving the multi-dimensional global coupling essential for coarse-grained evolution.

\paragraph{Attention as a Learnable Kernel Operator}
From the integral-operator perspective, attention mechanisms serve as learnable kernel transforms capable of expressing nonlocal coupling \cite{cao2021choose}. We model the attention operator $\mathcal{A}$ as a data-dependent kernel integral operator acting on the large-scale state $p_m$:
\begin{equation}
(\mathcal{A}(p_m;\theta)v)(x) = \int_{D} \kappa_{\theta}(x, y, p_m(x), p_m(y)) v(y) \, dy, \quad \forall x \in D,
\label{eq:attention_integral}
\end{equation}
where $D \subset \mathbb{R}^d$ is the spatial domain, $v$ is the value function, and $\kappa_{\theta}$ is a learnable kernel parametrized by $\theta$. In the discretized setting, a linear attention kernel $\kappa_{\theta}(x, y) = q(x) k^{\top}(y)$ aligns rigorously with the Petrov–Galerkin projections of transformer operators \cite{cao2021choose}. However, computing this full-rank kernel scales quadratically with spatial resolution ($\mathcal{O}(N^4)$ for 2D grids), which is prohibitive. We address this by completing our \textit{spectral dynamics embedding} and applying the \textit{Kronecker-structured attention} mechanism.

\paragraph{Spectral Dynamics Embedding}
To prepare the state for global interaction, we propose the dynamics embedding. While NLP tokens represent semantic concepts, our "tokens" represent the continuous physical state. To optimally embed this state, we operate in the frequency domain. 
Building upon the spectral truncation that isolates the large-scale modes, let $\hat{p}_m \in \mathbb{C}^{M_1 \times M_2 \times d_{\text{in}}}$ be the truncated Fourier representation from Eq. (\ref{eq:large_scale_projection}). We enrich this large-scale representation by applying learnable complex-valued weights in the frequency domain, which is analogous to linear projections in standard attention but constrained by spectral priors:
\begin{equation}
\tilde{u} = \mathcal{F}^{-1}\left( \mathcal{W}_{\text{spec}} \star \hat{p}_m \right),
\label{eq:spectral_embedding}
\end{equation}
where $\mathcal{W}_{\text{spec}} \in \mathbb{C}^{d_{\text{in}} \times d_{n} \times M_1 \times M_2}$ denotes the learnable spectral kernels, $\star$ denotes channel-wise multiplication, $d_n$ is the hidden dimension, and $\mathcal{F}^{-1}$ maps the features back to the physical domain (via zero-padding to $N_1 \times N_2$). This two-step embedding explicitly guarantees that the subsequent attention block operates strictly on a scale-separated, low-frequency latent state $\tilde{u} \in \mathbb{R}^{N_1 \times N_2 \times d_n}$.

\paragraph{Kronecker-Structured Attention Mechanism}
Following the embedding, we apply Kronecker-structured attention to mix features spatially. By leveraging the separability of the physical coordinate axes, we decompose the kernel function. For a 2D spatial domain with coordinates $\mathbf{x} = (x_1, x_2)$, we assume the kernel admits a Kronecker-type factorization:
\begin{equation}
\kappa_{\theta}(\mathbf{x}, \mathbf{y}) \approx \kappa_{\theta_1}(x_1, y_1) \cdot \kappa_{\theta_2}(x_2, y_2).
\label{eq:kronecker_kernel}
\end{equation}
To realize this, we generate 1D features for each spatial dimension via axis-wise reduction. We apply global mean pooling along orthogonal axes followed by linear projections $\text{Proj}_x, \text{Proj}_y$:
\begin{equation}
\begin{aligned}
\tilde{u}_{x}  &= \text{Proj}_x\left(\text{Mean}_{y}(\tilde{u})\right) \in \mathbb{R}^{N_1 \times d_{n}}, \\
\tilde{u}_{y}  &= \text{Proj}_y\left(\text{Mean}_{x}(\tilde{u})\right) \in \mathbb{R}^{N_2 \times d_{n}}.
\end{aligned}
\label{eq:axis_contraction}
\end{equation}
These reduced features are processed by a position-aware module. We employ separate query ($\phi_Q$) and key ($\phi_K$) networks, followed by rotary position embeddings (RoPE) \cite{su2024roformer} to inject relative positional information. For coordinates $i \in \{1, \dots, N_1\}$ and $j \in \{1, \dots, N_2\}$, and positional frequencies $\theta_x^{(i)}, \theta_y^{(j)}$, the rotary transformations are:
\begin{equation}
\begin{aligned}
\tilde{q}_x[i]  &= \mathcal{R}_{\theta_x^{(i)}} \cdot \phi_Q\left(\tilde{u}_x[i]\right), \quad \tilde{k}_x[i]  = \mathcal{R}_{\theta_x^{(i)}} \cdot \phi_K\left(\tilde{u}_x[i]\right), \\
\tilde{q}_y[j]  &= \mathcal{R}_{\theta_y^{(j)}} \cdot \phi_Q\left(\tilde{u}_y[j]\right), \quad \tilde{k}_y[j]  = \mathcal{R}_{\theta_y^{(j)}} \cdot \phi_K\left(\tilde{u}_y[j]\right),
\end{aligned}
\label{eq:rope_transformation}
\end{equation}
where $\mathcal{R}_{\theta} \in \mathbb{R}^{d_k \times d_k}$ is the standard block-diagonal rotation matrix. The axis-specific kernel matrices are computed via dot products:
\begin{equation}
\mathcal{K}_1[i, j] = \tilde{q}_x[i]^\top \tilde{k}_x[j], \quad \mathcal{K}_2[i, j] = \tilde{q}_y[i]^\top \tilde{k}_y[j].
\label{eq:kernel_matrices_rope}
\end{equation}
Given the value tensor $\tilde{V} \in \mathbb{R}^{N_1 \times N_2 \times d_{n}}$, the final spatial mixing for channel $c$ is computed via sequential matrix chaining:
\begin{equation}
(\mathcal{A}\tilde{V})_c = \mathcal{K}_1 \tilde{V}_c \mathcal{K}_2^\top, \quad c=1,\dots,d_{n}.
\label{eq:kronecker_output}
\end{equation}
In vector form, this operation aggregates information sequentially along each axis:
\begin{equation}
\mathcal{A} \tilde{v}_c \left(x_{ij}\right)=\sum_{k=1}^{N_1} \kappa_{\theta_1}\left(x_i, y_k\right)\left(\sum_{l=1}^{N_2} \kappa_{\theta_2}\left(x_j, y_l\right) \tilde{v}_c\left(y_{kl}\right)\right).
\label{eq:kronecker_vector}
\end{equation}
This physics-informed factorization directly decouples the computational complexity from $\mathcal{O}(N^4)$ to $\mathcal{O}(N^3)$. Unlike purely empirical factorized attention \cite{li2023scalable}, our derivation is fundamentally grounded in the separation-of-variables assumption, ensuring that this aggressive complexity reduction is safely applied only to the low-frequency manifold.

\subsection{Recovering Full-Scale Dynamics via Mixing Transformations}\label{sec:mixing_transformations}
Having modeled the large-scale evolution $p_m$ via the efficient Kronecker-structured attention (Section \ref{sec:kronecker_attention}), we now address the reconstruction of the full-scale state $u = p_m + q_m$. As established in Section \ref{sec:scale_decomposition}, the small-scale component $q_m$ is implicitly slaved to the large-scale dynamics via a mapping $\Phi: \mathscr{X}^M \rightarrow \mathscr{X}^{M^\perp}$ (Eq. \eqref{eq:small_scale_mapping}). 
However, reconstructing $q_m$ solely through linear transformations of $p_m$ faces a fundamental bottleneck: because $p_m$ is constructed via spectral truncation (Eq. \eqref{eq:large_scale_projection}), it is strictly band-limited to frequencies $\|\mathbf{k}\| \leq M$. Linear operations confined to this subspace cannot generate high-frequency components beyond the cutoff, effectively preventing the recovery of fine-scale structures. To overcome this spectral limitation and approximate the nonlinear mapping $\Phi$, we adopt the \textit{LGM} transformation \cite{lai2025dymixop}. Inspired by multiscale energy cascades in physical systems, LGM leverages nonlinear multiplicative interactions to expand the spectral bandwidth and implicitly reconstruct the small-scale residuals.

\paragraph{Local-Global-Mixing Transformation}
Formally, the LGM transformation $\mathscr{M}_\theta$ is defined as the Hadamard product of a global integral operator $\mathscr{G}_\theta$ and a local transformation $\mathscr{L}_\theta$ acting on the latent state $u$:
\begin{equation}
\mathscr{M}_\theta(u)(x) = \mathscr{L}_\theta(u)(x) \odot \mathscr{G}_\theta(u)(x), \quad \forall x \in D.
\label{eq:lgm_definition}
\end{equation}
In our architecture, the global operator $\mathscr{G}_\theta$ is instantiated specifically as the \textit{Kronecker-structured attention} block, i.e., $\mathscr{G}_\theta(u) \equiv \mathcal{A}(\tilde{u})$. This ensures the global branch captures long-range dependencies using the efficient $\mathcal{O}(N^3)$ axis-wise decomposition derived in Eq. \eqref{eq:kronecker_vector}. The local operator $\mathscr{L}_\theta$, conversely, is designed to capture fine-grained spatial details and is implemented via pointwise MLPs to maintain computational efficiency. 

This multiplicative formulation mirrors the nonlinear convective interactions (e.g., $u \cdot \nabla u$) that drive frequency mixing in fluid dynamics, offering a distinct spectral advantage over standard additive compositions (e.g., residual connections). Let $\hat{g}$ and $\hat{l}$ denote the Fourier spectra of the global and local outputs, with effective bandwidths $K$ and $M$ respectively. By the Convolution Theorem, the spectrum of the mixed output $\hat{m}$ satisfies $\text{supp}(\hat{m}) \subseteq \{\mathbf{k} : \|\mathbf{k}\| \leq K + M\}$. Because the local operator typically possesses a much larger effective bandwidth ($M \gg K$), the LGM transformation systematically modulates the global low-frequency carrier with local high-frequency details. This enables the representation of frequencies strictly greater than the global truncation threshold $K$, allowing DynFormer to recover the small-scale residual $q_m$ discarded during the initial spectral decomposition.

\paragraph{Integration into Full-Scale Dynamics Layers (FSDL)}
The LGM transformation serves as the essential nonlinear bridge between hierarchical dynamic scales. By integrating the Spectral Dynamics Embedding, Kronecker-structured attention, and the LGM transformation into a cohesive block, we form the \emph{Full-Scale Dynamics Layer (FSDL)}. 

Within the DynFormer architecture, stacking multiple FSDLs enables the network to approximate the multiscale mapping $\Phi(p_m)$. By enforcing this mixing structure on the latent states produced by the large-scale dynamics learner, we implicitly model the residual fine-scale dynamics without incurring the $\mathcal{O}(N^4)$ cost of full-grid attention. This ensures that the network not only propagates physically consistent large-scale correlations but also synthesizes the small-scale textures required for accurate full-field reconstruction. The output of the FSDL, denoted $\hat{u}_{\text{FSDL}} \in \mathbb{R}^{N_1 \times N_2 \times d_{n}}$, serves as the enriched feature representation for subsequent evolutionary steps.

\subsection{Evolutionary Operator with DynFormer}\label{sec:dynformer_operator}
Having established the scale decomposition (Section \ref{sec:scale_decomposition}), the efficient large-scale learner (Section \ref{sec:kronecker_attention}), and the multiscale mixing transformation (Section \ref{sec:mixing_transformations}), we now assemble these components into the complete DynFormer evolutionary operator. The architecture is designed to simulate the time evolution of complex physical systems by stacking Full-Scale Dynamics Layers (FSDLs) within a dynamics-informed framework that explicitly mirrors the structural decomposition of Eq. \eqref{eq:decomposed_dynamics}.

\paragraph{Full-Scale Dynamics Layer (FSDL) Internal Routing}
While Section \ref{sec:mixing_transformations} established the LGM transformation as the core mechanism for full-scale recovery, the internal routing of the FSDL is designed to approximate the composite differential operator $\mathscr{A} = \mathscr{L} + \mathscr{N}$ (Eq. \eqref{eq:general_dynamics}). To resolve linear and nonlinear dynamics independently, which is proposed in \cite{lai2025dymixop} and analogous to \emph{operator splitting} techniques in classical numerical PDE solvers \cite{gottlieb1977numerical}, the FSDL employs a multi-branch structure. Specifically, the dynamic update $\mathscr{F}_\theta(v)$ acting on a latent state $v \in \mathbb{R}^{N_1 \times N_2 \times d_n}$ is approximated by summing $n_l$ linear branches and $n_n$ nonlinear branches:
\begin{align}
\mathscr{F}_\theta(v) = \underbrace{\sum_{a=1}^{n_l} \mathscr{M}^{\mathscr{L}^{a}}_\theta(v)}_{\text{Approximation of } \mathscr{L}v} + \underbrace{\Psi_\theta\left[\sum_{b=1}^{n_n} \mathscr{M}^{\mathscr{N}^{b}}_\theta(v)\right]}_{\text{Approximation of } \mathscr{N}(v)},
\label{eq:fsdl_definition}
\end{align}
where $\mathscr{M}^{\mathscr{L}}_\theta$ and $\mathscr{M}^{\mathscr{N}}_\theta$ represent LGM transformations dedicated to capturing linear and nonlinear dynamics, respectively. The operator $\Psi_\theta$ serves as a learnable residual refinement module, implemented as a pointwise MLP, to enhance the representation of high-frequency nonlinear interactions. This physically motivated decoupling allows the layer to handle stiff and non-stiff dynamic components separately. Notably, the external forcing term $f$ can be recovered implicitly by the operator $\Psi_\theta$, maintaining network simplicity without sacrificing expressiveness.

\paragraph{Dynamics-Informed Architecture}
To simulate long-term system evolution, multiple FSDLs are sequentially connected to form a deep architecture. The network depth $L$ can be interpreted as a discretized surrogate for the time evolution of the system. Let $u$ denote the physical input field and $v$ denote the internal latent state. Assuming a continuous-time transformation induced by the operator $\mathscr{F}_\theta$, the architecture treats the depth $L$ as temporal discretization steps, yielding the discrete evolutionary dynamics:
\begin{align}
v_l = v_{l-1} + \Delta t_{\theta_l} \mathscr{F}_{\theta_l}(v_{l-1}), \quad l=1, \dots, L,
\label{eq:arch_evolution}
\end{align}
where $v_0 = \mathscr{T}_{\text{up}}(u)$ is the lifted latent state, and $\Delta t_{\theta_l}$ denotes a learnable temporal scaling factor for layer $l$. This formulation mimics a nonuniform, learnable numerical integration scheme. Depending on the physical nature of the target system, Eq. \eqref{eq:arch_evolution} provides three architectural variants:

\begin{enumerate}[(i)]
    \item Hierarchical Variant: Each FSDL receives the output of the previous layer, progressively refining the state through sequential updates. This aligns with standard autoregressive time-stepping schemes for evolutionary PDEs, where the state at $t+\Delta t$ depends strictly on $t$.
    \item Parallel Variant: All FSDLs share the same input $v_0$ and evolve the state in parallel, accumulating the output with a shared temporal step:
    \begin{align}
    v_{L} = v_0 + \Delta t \sum_{l=1}^{L} \mathscr{F}_{\theta_l}(v_0).
    \label{eq:arch_parallel}
    \end{align}
    This variant is optimally suited for steady-state problems (e.g., Darcy flow) where the solution represents an equilibrium field rather than a temporal trajectory.
    \item Hybrid Variant: We decouple the update step $\Delta t_{\theta_l}$ across layers while maintaining sequential dependency:
    \begin{align}
    v_{L} = v_0 + \sum_{l=1}^{L}  \Delta t_{\theta_l} \mathscr{F}_{\theta_l}(v_{l-1}).
    \label{eq:arch_hybrid}
    \end{align}
    This formulation acts as a learnable, adaptive Runge-Kutta method. By allowing the network to dynamically adjust its "time step" per layer, it offers the optimal balance between numerical stability and expressiveness, making it highly effective for chaotic systems with extreme multiscale temporal dynamics (e.g., Navier-Stokes).
\end{enumerate}

\begin{figure}[hbtp]
  \centering
  \includegraphics[width=0.95\textwidth]{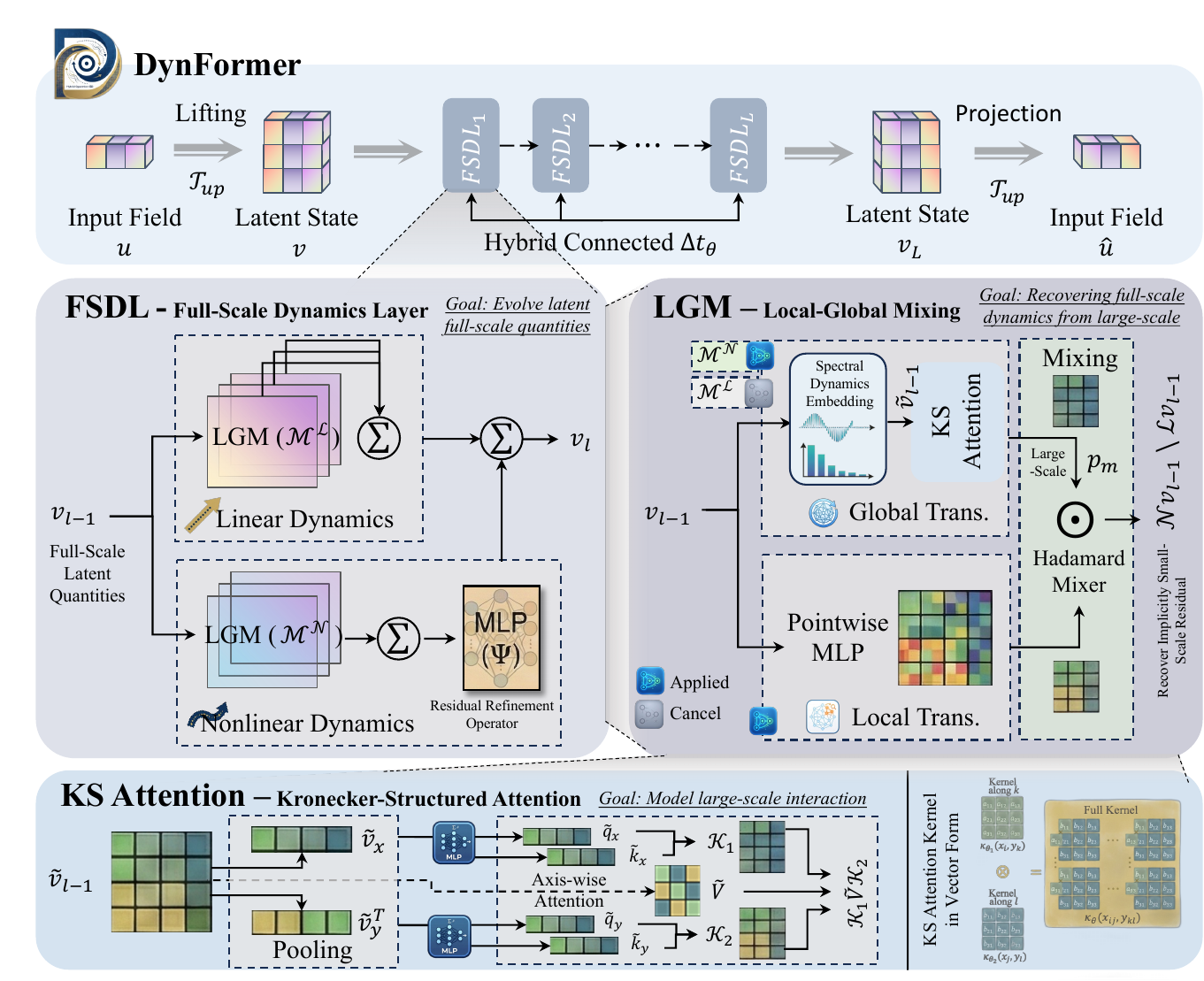}
  \caption{Illustration of the DynFormer architecture.}
  \label{fig:arch}
\end{figure}

\paragraph{Complete DynFormer Framework}
The overall DynFormer architecture (Fig. \ref{fig:arch}) forms an end-to-end operator learning framework consisting of three stages:
\begin{enumerate}[(i)]
    \item Lifting Transformation: A pointwise operator $\mathscr{T}_{\text{up}}: \mathbb{R}^{N_1 \times N_2 \times d_{\text{in}}} \rightarrow \mathbb{R}^{N_1 \times N_2 \times d_n}$ projects the input field $u$ into the expanded latent space $v_0$.
    \item Latent Evolution: A sequence of $L$ FSDLs evolves the latent state (e.g., via the Hybrid scheme in Eq. \eqref{eq:arch_hybrid}), applying the scale-aware dynamics approximation from Eq. \eqref{eq:fsdl_definition}.
    \item Projection Transformation: A pointwise operator $\mathscr{T}_{\text{down}}: \mathbb{R}^{N_1 \times N_2 \times d_n} \rightarrow \mathbb{R}^{N_1 \times N_2 \times d_{\text{out}}}$ maps the final latent state $v_L$ back to the physical solution space, producing the predicted field $\hat{u}$.
\end{enumerate}
The complete forward pass is summarized as:
\begin{align}
\hat{u} = \mathscr{T}_{\text{down}}\left( v_L \right), \quad \text{where } v_L = \text{FSDL}_L \circ \cdots \circ \text{FSDL}_1 \left( \mathscr{T}_{\text{up}}(u) \right).
\label{eq:dynformer_forward}
\end{align}

\paragraph{Computational Complexity}
The computational efficiency of DynFormer stems from two critical design choices. First, the Kronecker-structured attention reduces the spatial complexity from $\mathcal{O}(N^4)$ to $\mathcal{O}(N^3)$ for 2D grids of size $N \times N$. Second, the spectral truncation at $M$ modes bounds the frequency-domain operations to $\mathcal{O}(M^2 \log M)$ via FFT. For typical configurations where $M \ll N$ (e.g., $M = N/4$), the overall complexity per FSDL is vastly dominated by the $\mathcal{O}(N^3)$ Kronecker attention. This yields massive memory and computational savings compared to standard transformer-based neural operators, without sacrificing the expressive power required for multi-scale PDEs (Table \ref{tab:complexity_comparison}).

\begin{table}[h]
\centering
\caption{Computational complexity comparison for 2D spatial domains of size $N \times N$.}
\label{tab:complexity_comparison}
\begin{tabular}{l c}
\toprule
\textbf{Method} & \textbf{Complexity} \\
\midrule
Standard Self-Attention & $\mathcal{O}(N^4)$ \\
Kronecker Attention & $\mathcal{O}(N^3)$ \\
Spectral Embedding & $\mathcal{O}(M^2 \log M)$ \\
\bottomrule
\end{tabular}
\end{table}

\section{Results}
\subsection{Baselines}
(i) \textbf{ONO} (Orthogonal Neural Operator) implements a novel orthogonal attention mechanism: rather than softmax, it orthogonalizes learned eigenfunction projections. It separates eigenfunction approximation and solution learning into two disentangled pathways \cite{xiao2023improved}.
(ii) \textbf{FactFormer} (Factorized Transformer) employs axial factorized attention: it decomposes high-dimensional inputs into multiple 1D sub-functions using a learnable projection, then computes attention among these sub-components \cite{li2023scalable}.
(iii) \textbf{OFormer} (Operator Transformer) combines self‑attention and cross‑attention layers, along with pointwise MLP processing \cite{li2022transformer}.
(iv) \textbf{GNOT} (General Neural Operator Transformer) introduces heterogeneous normalized attention paired with a geometric gating mechanism, enabling adaptive interaction over irregular meshes and capturing multi-scale behaviors \cite{hao2023gnot}.
(v) \textbf{Transolver} proposes Physics‑Attention by grouping mesh points into learnable slices (physical-state tokens) and attending over these tokens instead of raw points \cite{wu2024transolver}.

\subsection{Benchmarks}
To validate both the scalability and versatility of our method, we evaluate it across four classic PDE models that exhibit distinct mathematical behaviors and physical regimes. These benchmarks span parabolic, elliptic, and hyperbolic behaviors, from chaotic, dissipative, steady-state, to conservation-law regimes, allowing us to thoroughly assess the range and robustness of our operator learner:
(i) \textbf{1D Kuramoto Sivashinsky} (KS), a fourth-order nonlinear parabolic PDE originally derived to model diffusive‑thermal instabilities. The KS equation provides a stringent test of a model’s ability to capture chaotic dissipative dynamics and scale separation in one spatial dimension.
(ii) \textbf{2D Darcy Flow}, a second-order elliptic PDE describing steady-state porous‑media flow with spatially varying permeability. As a model for steady elliptic behavior, Darcy flow tests how well an operator‑learning method generalizes across sharp spatial heterogeneities under global equilibrium conditions.
(iii) \textbf{2D Navier-Stokes} (NS), a parabolic PDE system governing two-dimensional incompressible viscous fluid motion. It introduces highly nonlinear advection and diffusion, offering a rigorous test of a model's capacity to resolve fine-scale turbulent cascades.
(iv) \textbf{3D Shallow Water} (SW), a set of hyperbolic conservation‑law PDEs. These equations model large‑scale geophysical wave dynamics. Their nonlinearity, shock‐like features, and global spherical geometry serve as benchmarks for transferring learned operators to complex, large‑scale curvilinear domains.

\subsection{Experimental Setup and Evaluation Metrics}\label{sec:experimental_setup}
Comparative evaluations in the neural operator literature often suffer from performance gains attributed to meticulous hyperparameter tuning or task-specific architectural adjustments rather than fundamental algorithmic superiority. This reliance on over-optimization creates significant challenges for establishing fair benchmarks \cite{mcgreivy2024weak}. To address this, we evaluate all baselines using standardized configurations, strictly avoiding task-specific parameter engineering.

\paragraph{Capacity Scaling and Memory Alignment}
Our scaling strategy focuses principally on adjusting the hidden dimension and network depth to vary model capacity. Crucially, we align these hyperparameter settings based on \textbf{peak GPU memory usage} rather than parameter count. Because parameter count often scales nonlinearly with actual computational costs (especially in attention-based models), memory usage serves as a more practical metric for fairness. Under these strict constraints, we construct three model variants—Tiny, Medium, and Large—spanning a diverse range of GPU memory budgets. This ensures our assessment reflects the intrinsic potential of the architecture. The results reported in Section \ref{sec:main_results} correspond to the optimal configuration found within these memory-aligned constraints.

\paragraph{Training Hyperparameters}
To ensure reproducibility and mitigate stochastic variations, all models are trained using a uniform configuration. Training proceeds for $500$ epochs using the AdamW optimizer \cite{loshchilov2017decoupled} with an initial learning rate of $10^{-3}$, a weight decay of $0.97$, and a step size of $7$. A StepLR scheduler adjusts the learning rate dynamically. The batch size is fixed at $64$ across all experiments on NVIDIA A800 GPUs. Furthermore, all experiments are conducted using two distinct random seeds $\{123, 456\}$, and the averaged results are reported to evaluate robustness.

\paragraph{Data Normalization and Loss Function}
Let $\mathcal{D} = \{ (a_i, u_i) \}_{i=1}^{|\mathcal{D}|}$ denote the dataset, where $a_i$ is the input field and $u_i \in \mathbb{R}^{n}$ is the ground truth solution. All datasets are normalized using min-max normalization to stabilize training:
\begin{equation}
u^* = \frac{u - u_{\text{min}}}{u_{\text{max}} - u_{\text{min}}}.
\label{eq:minmax_norm}
\end{equation}
For optimization and evaluation, we adopt the relative mean squared error (MSE) to address scenarios with extremely minimal solution magnitudes and to prevent gradient vanishing. The relative MSE $\epsilon$ is computed as:
\begin{equation}
\epsilon = \frac{1}{|\mathcal{D}|} \sum_{i=1}^{|\mathcal{D}|} \frac{\| \hat{u}_i - u_i \|_2^2}{\| u_i \|_2^2},
\label{eq:relative_mse}
\end{equation}
where $\hat{u}_i$ denotes the predicted solution.

\paragraph{Log-Min-Max Normalized Score}
To enable an intuitive and fair comparison across diverse benchmarks—which exhibit vastly different loss magnitudes due to physical scales and problem complexity—we introduce a Log-Min-Max normalized score mapped to the $[0, 100]$ interval. Let $\epsilon_{\text{min}}$ and $\epsilon_{\text{max}}$ denote the smallest and largest relative MSE values observed among all evaluated models on a specific benchmark. The score is computed as:
\begin{equation}
\text{Score} = 100 \cdot \left(1 - \frac{\log(\epsilon) - \log(\epsilon_{\text{min}})}{\log(\epsilon_{\text{max}}) - \log(\epsilon_{\text{min}})}\right).
\label{eq:log_min_max_score}
\end{equation}
This logarithmic transformation ensures uniform sensitivity across multiple orders of magnitude, emphasizing relative improvement over absolute error scale. 

\subsection{Main Comparison Results}
\label{sec:main_results}

\begin{figure}[htbp]
  \centering
  \includegraphics[width=\linewidth]{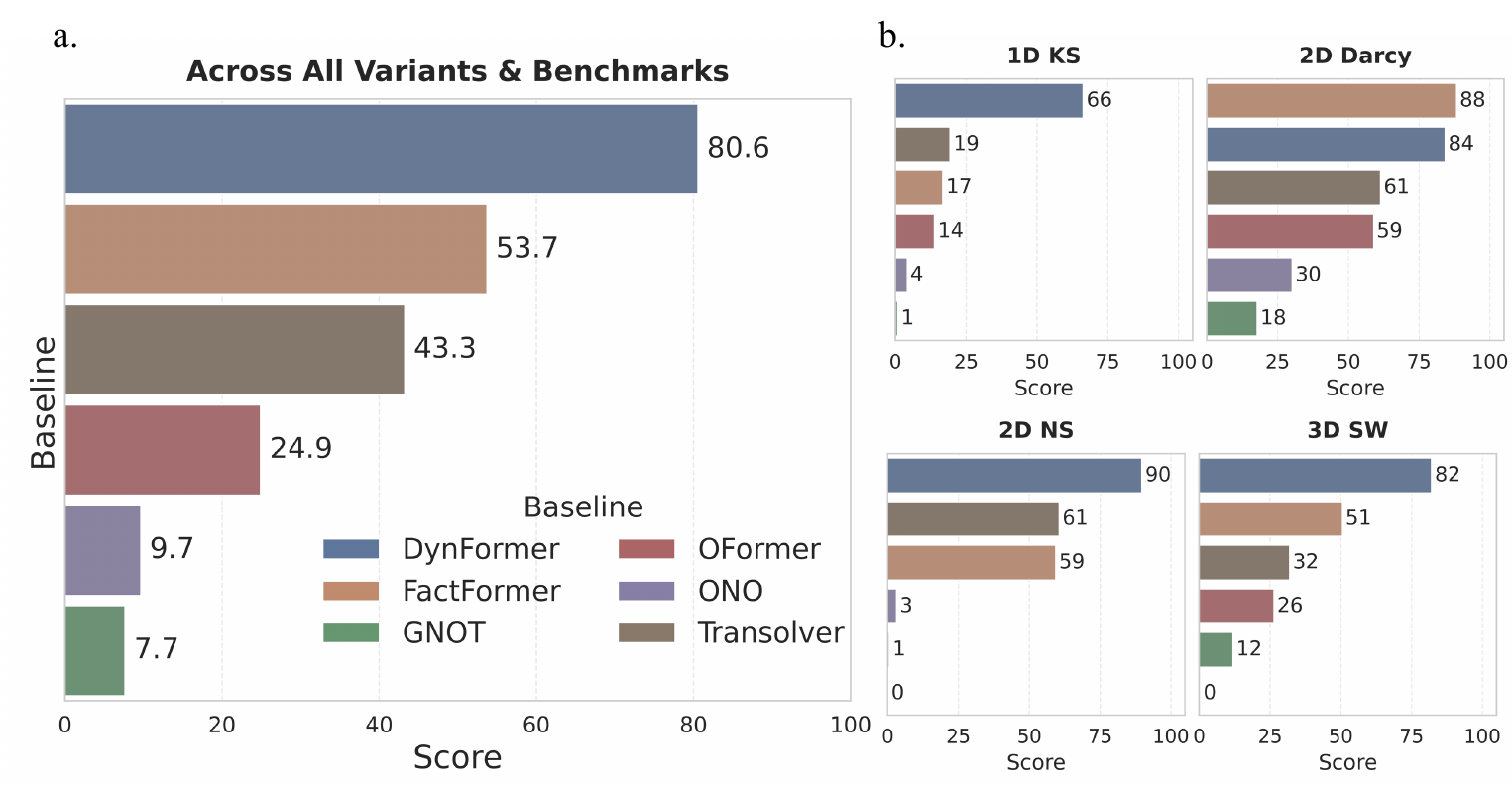}
\caption{Performance comparison of neural operator architectures across diverse PDE benchmarks. 
(a) Aggregated Log-Min-Max normalized scores (0--100) averaged over all benchmarks, model variants (Tiny/Medium/Large), and random seeds, highlighting DynFormer's substantial lead over state-of-the-art baselines. 
(b) Per-benchmark breakdown showing DynFormer's dominance across 1D Kuramoto-Sivashinsky, 2D Darcy, 2D Navier-Stokes, and 3D Shallow Water equations.}
\label{fig:performance}
\end{figure}

\begin{figure}[htbp]
  \centering
  \includegraphics[width=\linewidth]{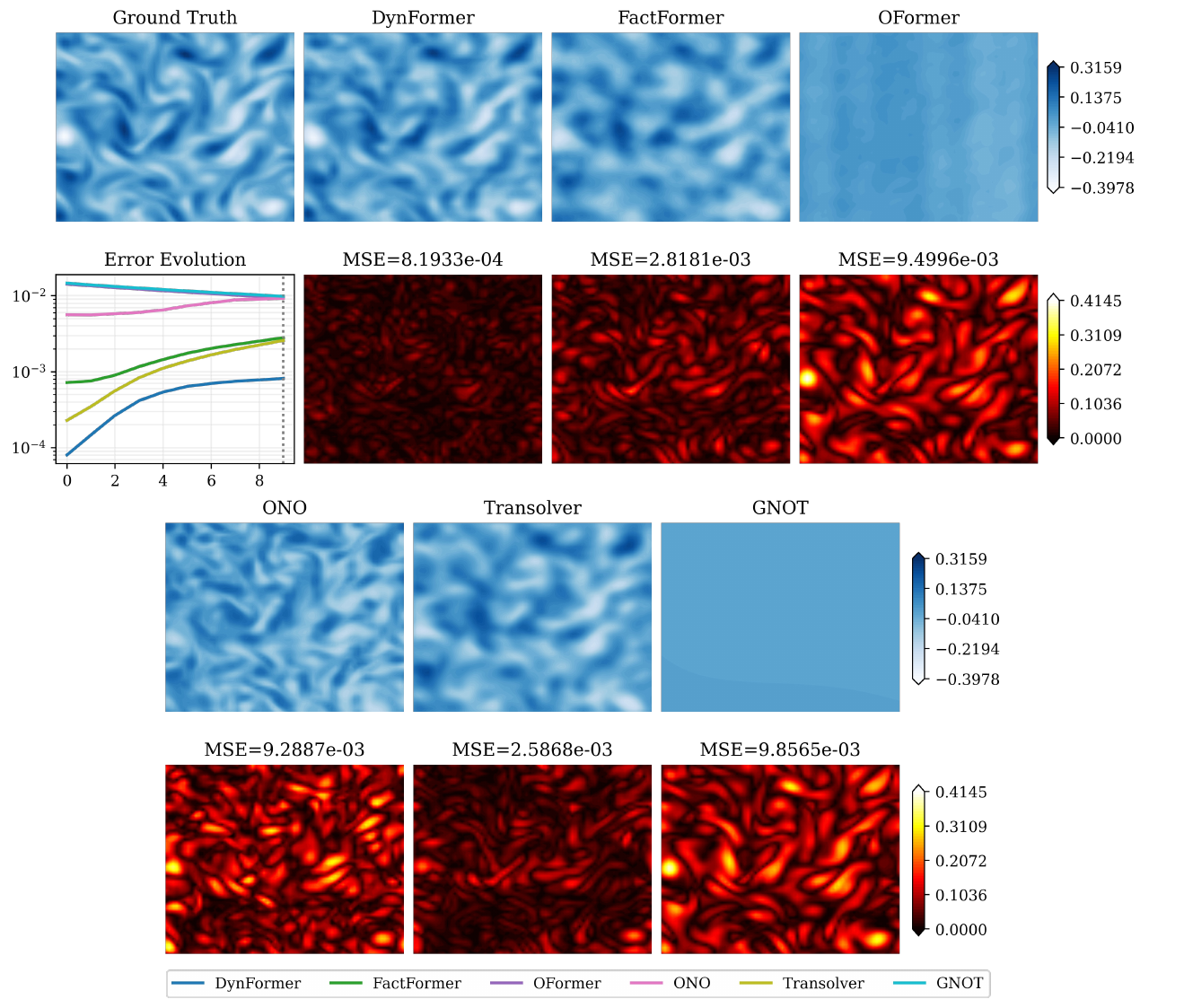}
\caption{Qualitative visualization and error analysis on the 2D Navier-Stokes benchmark. 
Top row: Predicted vorticity fields compared against Ground Truth. DynFormer captures fine-scale turbulent structures with high fidelity, whereas baselines exhibit smoothing or artifacts. 
Bottom left: Error evolution over simulation timesteps, demonstrating DynFormer's stability. 
Bottom right: Mean Squared Error (MSE) comparison, where DynFormer achieves an order-of-magnitude reduction compared to competing methods.}
\label{fig:visualization}
\end{figure}

\begin{figure}[htbp]
  \centering
  \includegraphics[width=\linewidth]{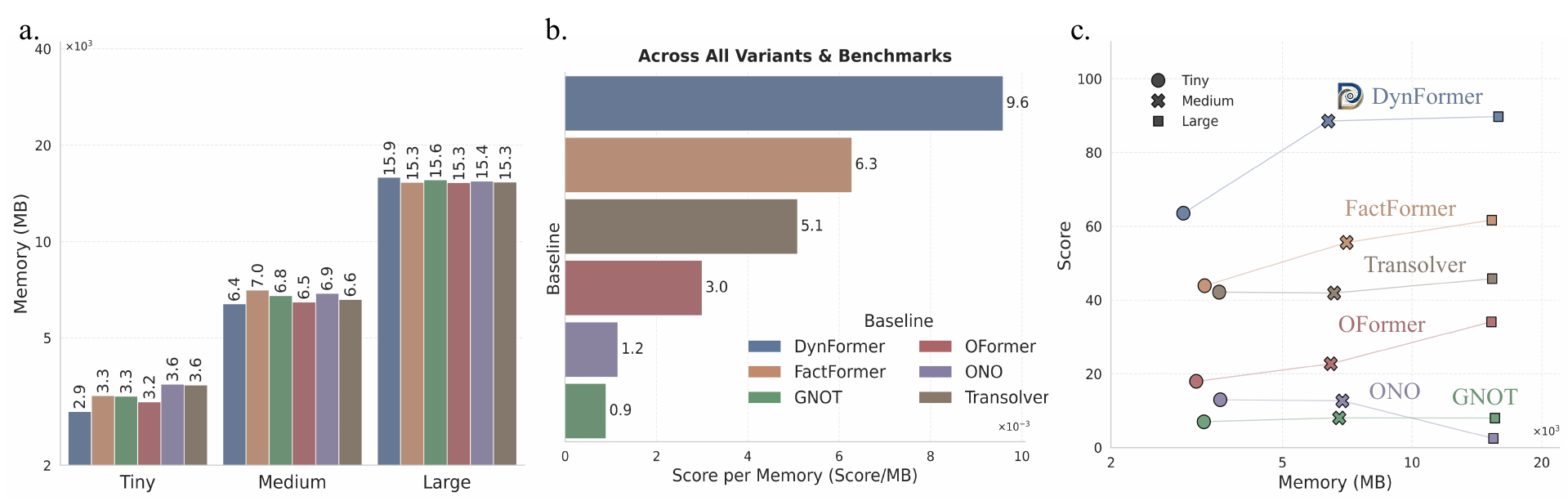}
\caption{Computational efficiency and memory--performance trade-offs.
(a) Peak GPU memory consumption (MB) for Tiny, Medium, and Large variants. All baselines are built with comparable consumptions.
(b) Score per Memory ratio, illustrating DynFormer's superior efficiency by reaching 9.6. 
(a) Scalability analysis across Tiny, Medium, and Large model variants. DynFormer consistently achieves higher performance scores for equivalent or lower memory usage, with its Tiny variant outperforming the Large variants of several baselines.}
\label{fig:efficiency}
\end{figure}

\begin{table}[htbp]
  \centering
  \small
  \setlength{\tabcolsep}{3pt}
  \renewcommand{\arraystretch}{1.2}
  \caption{Comprehensive performance comparison across PDE benchmarks. Metrics reported per model size variant (Tiny/Medium/Large): peak GPU memory usage (GB), relative MSE loss on test dataset, and averaged training time (seconds) per epoch. The \textbf{Improvement} row indicates the percentage reduction in error achieved by DynFormer compared to the previous state-of-the-art (SOTA). \textbf{Bold}, \textbf{\color{bestblue}blue}, and \uline{underline} indicate the 1\textsuperscript{st}, 2\textsuperscript{nd}, and 3\textsuperscript{rd} best results among competitor models. DynFormer results are highlighted with gray background.}
  \label{tab:results_comparison_detailed}
  \sisetup{
    output-exponent-marker = \ensuremath{\mathrm{e}},
    retain-explicit-plus = false,
    mode = match,
    reset-text-series = false,
    reset-text-family = false,
    reset-text-shape = false,
    text-series-to-math = true,
    exponent-mode = scientific,
    table-format = 1.2e-2
  }
  \resizebox{\textwidth}{!}{
  \begin{tabular}{l c *{4}{c c c c}}
    \toprule
    & & \multicolumn{4}{c}{\textbf{1D}} & \multicolumn{8}{c}{\textbf{2D}} & \multicolumn{4}{c}{\textbf{3D}} \\
    \cmidrule(lr){3-6} \cmidrule(lr){7-14} \cmidrule(lr){15-18}
    \multirow{2}{*}{\textbf{Model}} & \multirow{2}{*}{\textbf{Sz}} & \multicolumn{4}{c}{\textbf{Kuramoto-Sivashinsky}} & \multicolumn{4}{c}{\textbf{Darcy}} & \multicolumn{4}{c}{\textbf{Navier-Stokes}} & \multicolumn{4}{c}{\textbf{Shallow Water}} \\
    \cmidrule(lr){3-6} \cmidrule(lr){7-10} \cmidrule(lr){11-14} \cmidrule(lr){15-18}
    & & \textbf{Mem} & \textbf{Params} & {\textbf{Loss}} & \textbf{Time} & \textbf{Mem} & \textbf{Params} & {\textbf{Loss}} & \textbf{Time} & \textbf{Mem} & \textbf{Params} & {\textbf{Loss}} & \textbf{Time} & \textbf{Mem} & \textbf{Params} & {\textbf{Loss}} & \textbf{Time} \\
    \midrule
    \multirow{3}{*}{FactFormer \cite{li2023scalable}}
      & T & 0.8 & 6.6K & 5.12e-01 & 2.0 & 0.7 & 38.9K & {\third{8.12e-04}} & 0.4 & 7.6 & 25.3K & 1.85e-01 & 12.3 & 4.1 & 25.6K & 1.24e-01 & 13.3 \\
      & M & 2.9 & 51.4K & 3.74e-01 & 3.6 & 2.9 & 476.1K & {\second{1.40e-04}} & 1.0 & 14.8 & 99.2K & 1.52e-01 & 13.5 & 7.7 & 99.9K & {\second{5.58e-02}} & 13.4 \\
      & L & 12.3 & 638.0K & {\third{3.38e-01}} & 6.8 & 13.0 & 7.3M & {\best{7.67e-05}} & 4.8 & 23.7 & 202.6K & {\second{1.41e-01}} & 16.1 & 12.3 & 203.3K & {\best{2.66e-02}} & 15.1 \\
    \addlinespace[0.3em]
    \multirow{3}{*}{GNOT \cite{hao2023gnot}}
      & T & 0.9 & 14.1K & 1.00e+00 & 2.5 & 0.8 & 8.0K & 7.60e-02 & 0.5 & 7.5 & 3.8K & 9.96e-01 & 11.1 & 3.9 & 3.9K & 5.94e-01 & 12.1 \\
      & M & 3.0 & 59.2K & 1.00e+00 & 4.7 & 3.1 & 103.5K & 7.60e-02 & 1.2 & 14.0 & 7.3K & 9.99e-01 & 16.8 & 7.0 & 7.4K & 4.65e-01 & 15.6 \\
      & L & 14.6 & 993.3K & 1.00e+00 & 9.0 & 11.6 & 786.1K & 7.62e-02 & 4.1 & 23.8 & 16.4K & 9.94e-01 & 28.1 & 12.3 & 16.5K & 4.75e-01 & 21.9 \\
    \addlinespace[0.3em]
    \multirow{3}{*}{OFormer \cite{li2022transformer}}
      & T & 0.9 & 11.6K & 9.49e-01 & 2.2 & 0.9 & 11.6K & 4.54e-03 & 0.4 & 7.1 & 3.0K & 9.91e-01 & 12.4 & 3.7 & 3.1K & 3.52e-01 & 12.9 \\
      & M & 2.7 & 92.0K & 6.47e-01 & 3.1 & 2.6 & 91.9K & 3.14e-03 & 0.9 & 13.6 & 11.6K & 9.86e-01 & 13.8 & 7.0 & 11.7K & 2.31e-01 & 13.0 \\
      & L & 11.3 & 1.4M & {\best{1.77e-01}} & 5.3 & 12.3 & 1.7M & 9.84e-04 & 4.5 & 24.9 & 33.5K & 9.72e-01 & 17.4 & 12.6 & 33.7K & 1.39e-01 & 14.6 \\
    \addlinespace[0.3em]
    \multirow{3}{*}{ONO \cite{xiao2023improved}}
      & T & 0.8 & 21.1K & 7.71e-01 & 2.3 & 0.9 & 25.6K & 9.45e-03 & 0.5 & 8.2 & 8.5K & 9.99e-01 & 13.2 & 4.4 & 8.9K & 9.94e-01 & 12.9 \\
      & M & 3.0 & 152.1K & 7.89e-01 & 5.1 & 3.1 & 151.0K & 9.42e-03 & 1.2 & 14.0 & 17.3K & 9.78e-01 & 16.3 & 7.5 & 17.8K & 9.50e-01 & 15.0 \\
      & L & 13.2 & 1.2M & 1.05e+00 & 10.8 & 11.3 & 1.2M & 3.37e-01 & 4.0 & 23.8 & 38.7K & 7.93e-01 & 21.9 & 12.3 & 39.3K & 9.89e-01 & 17.9 \\
    \addlinespace[0.3em]
    \multirow{3}{*}{Transolver \cite{wu2024transolver}}
      & T & 0.8 & 80.4K & 4.66e-01 & 2.0 & 0.8 & 65.1K & 1.32e-03 & 0.4 & 8.0 & 22.0K & {\third{1.51e-01}} & 10.9 & 4.7 & 22.4K & 2.12e-01 & 10.5 \\
      & M & 2.7 & 369.7K & 5.06e-01 & 3.7 & 2.6 & 283.6K & 9.61e-04 & 1.0 & 13.9 & 42.9K & 1.92e-01 & 13.8 & 7.2 & 43.4K & 1.73e-01 & 11.7 \\
      & L & 13.5 & 2.8M & {\second{1.93e-01}} & 7.2 & 12.3 & 2.8M & 6.88e-03 & 4.6 & 23.5 & 95.8K & {\best{1.27e-01}} & 18.6 & 12.1 & 96.5K & {\third{1.20e-01}} & 16.9 \\
    \midrule
    \rowcolor{rowgray}
    & T & 1.0 & 1.2K & 1.69e-01 & 3.9 & 0.7 & 32.6K & 3.64e-04 & 0.7 & 6.9 & 15.1K & 9.66e-02 & 12.9 & 3.2 & 7.0K & 3.23e-02 & 13.8 \\
    \rowcolor{rowgray}
    & M & 2.8 & 5.9K & {\best{8.63e-03}} & 5.7 & 2.4 & 162.9K & 3.06e-04 & 1.4 & 13.3 & 57.9K & 5.03e-02 & 12.8 & 7.1 & 39.7K & 6.18e-03 & 14.9 \\
    \rowcolor{rowgray}
    \multirow{-3}{*}{\textbf{DynFormer(Ours)}} & L & 12.9 & 163.5K & 2.30e-02 & 11.4 & 14.1 & 3.4M & {\best{2.19e-04}} & 3.6 & 24.6 & 110.7K & {\best{4.79e-02}} & 17.2 & 11.9 & 82.9K & {\best{4.57e-03}} & 18.9 \\
    \midrule
    \multicolumn{2}{@{}l}{\textbf{\textit{Improvement vs. SOTA}}} & & & \imp{95.1} & & & & {-} & & & & \imp{62.3} & & & & \imp{82.8} & \\
    \bottomrule
  \end{tabular}
  }
\end{table}

To comprehensively evaluate the efficacy of DynFormer, we conducted comparative experiments against five state-of-the-art neural operator baselines across four distinct PDE benchmarks. The aggregated performance metrics are summarized in Figure~\ref{fig:performance}, while detailed quantitative results are provided in Table~\ref{tab:results_comparison_detailed}. 

\paragraph{Performance on Various Benchmarks}
As illustrated in Figure~\ref{fig:performance}(a), DynFormer achieves a dominant aggregate score of 80.6 across all variants and benchmarks, substantially outperforming the nearest competitor, FactFormer (scoring 53.7). This aggregate advantage confirms that DynFormer's dynamics-informed architecture successfully generalizes across diverse physical regimes. Crucially, the memory-aligned evaluation protocol ensures these gains are driven by our theoretical architectural advantages, specifically spectral scale separation and mixing, rather than computational inflation.
Examining performance at the individual benchmark level (Figure~\ref{fig:performance}(b) and Table~\ref{tab:results_comparison_detailed}) reveals how DynFormer's theoretical mechanisms translate into empirical success:
\begin{itemize}
    \item 1D Kuramoto-Sivashinsky (KS): DynFormer achieves an unparalleled score of 66, demonstrating a 95.1$\%$ error reduction over the SOTA. This highlights the effectiveness of our \textit{Scale Decomposition}. By isolating the large-scale chaotic carriers from the stiff, fourth-order hyperdiffusive noise, DynFormer maintains long-term temporal stability where models without scale separation fail entirely (e.g., Transolver: scoring 19).
    
    \item 2D Darcy Flow: DynFormer achieves a highly competitive score of 84, while FactFormer attains 88. Darcy flow is an elliptic, linear diffusion-dominated system that strictly relies on steady-state spatial interpolation rather than nonlinear frequency mixing. This result suggests that while \textit{LGM Transformation} is highly optimized for transient, nonlinear dynamics, DynFormer still maintains robust spatial operator approximation capabilities under highly constrained global equilibrium conditions.

    \item 2D Navier-Stokes (NS): DynFormer achieves a score of 90, opening a massive performance gap over FactFormer (scoring 59). This explicitly validates the \textit{LGM Transformation}. While other models suffer from numerical diffusion, DynFormer's multiplicative mixing expands the spectral bandwidth, allowing it to accurately recover sub-grid scale turbulent cascades and nonlinear advective interactions.
    
    \item 3D Shallow Water (SW): DynFormer demonstrates robust generalization to hyperbolic wave dynamics, scoring 82 and achieving an 82.8$\%$ error reduction over the SOTA. The efficient \textit{Kronecker-Structured Attention} explicitly captures the global planetary-scale geostrophic balances, while the nonlinear branches resolve the localized wave-breaking phenomena.
\end{itemize}

\paragraph{Qualitative Assessment and Error Stability}
Complementing the quantitative findings, Figure~\ref{fig:visualization} provides visual validation of DynFormer's theoretical claims on the 2D Navier-Stokes benchmark. Visually, DynFormer is the only model that avoids artificial numerical diffusion, preserving the crisp, high-frequency filamentary structures of the ground truth. This is the direct empirical result of the LGM module overcoming the spectral truncation limits of the global attention mechanism. FactFormer and Transolver exhibit severe smoothing artifacts, confirming that without an explicit nonlinear mixing mechanism, small-scale residuals ($q_m$) are lost. Quantitatively, the error evolution trajectory (Figure~\ref{fig:visualization}, bottom left) confirms DynFormer's stability, reflecting the architectural benefit of decoupling linear and nonlinear dynamic branches (Eq. \eqref{eq:fsdl_definition}).

\paragraph{Computational Efficiency Trade-offs}
Figure~\ref{fig:efficiency} proves that the reduction of spatial complexity from $\mathcal{O}(N^4)$ to $\mathcal{O}(N^3)$ via Kronecker-structured attention translate directly into superior hardware efficiency. The score-per-memory metric (Figure~\ref{fig:efficiency}(b)) shows DynFormer achieving 9.6, nearly double that of FactFormer (scoring 5.1). 
The scalability analysis (Figure~\ref{fig:efficiency}(c)) reveals that even the \textit{Tiny} variant of DynFormer achieves competitive absolute scores while consuming substantially less memory than the \textit{Large} variants of baseline methods. For instance, on the 3D Shallow Water benchmark, DynFormer-Medium achieves a relative MSE of $6.18 \times 10^{-3}$, outperforming FactFormer-Large (MSE $2.66 \times 10^{-2}$) while using 42\% less memory. Furthermore, DynFormer maintains highly competitive training throughput (Table~\ref{tab:results_comparison_detailed}), confirming that the spectral embedding and Kronecker attention add minimal wall-clock overhead while drastically reducing memory footprint.

\subsection{Ablation Experiments}\label{sec:ablation_experiments}
To validate the specific theoretical design choices underlying DynFormer, we conduct a series of ablation studies across the 2D Darcy Flow and 2D Navier-Stokes benchmarks. These experiments systematically isolate the contributions of the Kronecker-structured attention, spectral embedding, mixing transformations, and evolutionary flow. 

\paragraph{Impact of Kronecker-Structured Attention (E1)}
We first evaluate the efficiency of the proposed Kronecker-structured (KS) attention mechanism (Table~\ref{tab:ablation_e1}). Classical full-rank attention incurs prohibitive computational costs, consuming 57.5 GB of GPU memory on the 2D NS benchmark due to its $\mathcal{O}(N^4)$ complexity, and fails to converge (Loss: $4.07 \times 10^{-1}$). Standard linear attention reduces memory usage but still suffers from significant accuracy degradation (Loss: $1.93 \times 10^{-1}$), as it struggles to capture the complex global dependencies inherent in fluid dynamics. DynFormer with KS attention achieves the lowest test loss ($5.60 \times 10^{-2}$) at a moderate 21.9 GB memory footprint, proving that axis-wise factorization successfully preserves global coupling while solving the quadratic scalability bottleneck.
\begin{table}[htbp]
  \centering
  \small
  \setlength{\tabcolsep}{4pt}
  \renewcommand{\arraystretch}{1.2}
  \caption{\textbf{Ablation Experiment 1 (E1): Attention Mechanisms.} Detailed comparison of variants on 2D Darcy and 2D NS. Memory (Mem) is reported in GB, and training time per epoch (Time) is in seconds. \textbf{Bold} indicates the best test loss.}
  \label{tab:ablation_e1}
  \begin{tabular}{l c c c c c c c c}
    \toprule
    & \multicolumn{4}{c}{\textbf{2D Darcy}} & \multicolumn{4}{c}{\textbf{2D NS}} \\
    \cmidrule(lr){2-5} \cmidrule(lr){6-9}
    \textbf{Variant} & \textbf{Mem} & \textbf{Params} & {\textbf{Loss}} & \textbf{Time} & \textbf{Mem} & \textbf{Params} & {\textbf{Loss}} & \textbf{Time} \\
    \midrule
    KS Attention & 13.1 & 3.0M & \textbf{1.68e-04} & 3.2 & 21.9 & 98.2K & \textbf{5.60e-02} & 16.9 \\
    Classical Attention & 17.0 & 1.3K & 1.25e-01 & 2.4 & 57.5 & 49.0B & 4.07e-01 & 9.8 \\
    Linear Attention & 8.7 & 857.2K & 3.25e-04 & 3.7 & 22.2 & 46.7K & 1.93e-01 & 29.0 \\
    \bottomrule
  \end{tabular}
\end{table}

\paragraph{Impact of Spectral Embedding (E2)}
We investigate the benefit of establishing the scale decomposition latent space in the spectral domain (Eq. \eqref{eq:spectral_embedding}) versus the physical domain (Table~\ref{tab:ablation_e2}). While physical embedding performs marginally better on the linear-dominated Darcy benchmark ($7.60 \times 10^{-5}$), the Spectral variant significantly outperforms on the highly nonlinear Navier-Stokes task ($5.60 \times 10^{-2}$ vs. $1.54 \times 10^{-1}$). This confirms our hypothesis from Section 2.1: spectral truncation explicitly isolates non-stiff low-frequency modes from small-scale noise, providing a far more robust latent state for global processing in chaotic systems.
\begin{table}[htbp]
  \centering
  \small
  \setlength{\tabcolsep}{4pt}
  \renewcommand{\arraystretch}{1.2}
  \caption{\textbf{Ablation Experiment 2 (E2): Decomposition Strategy.} Detailed comparison of variants on 2D Darcy and 2D NS. Memory (Mem) is reported in GB, and training time per epoch (Time) is in seconds. \textbf{Bold} indicates the best test loss.}
  \label{tab:ablation_e2}
  \begin{tabular}{l c c c c c c c c}
    \toprule
    & \multicolumn{4}{c}{\textbf{2D Darcy}} & \multicolumn{4}{c}{\textbf{2D NS}} \\
    \cmidrule(lr){2-5} \cmidrule(lr){6-9}
    \textbf{Variant} & \textbf{Mem} & \textbf{Params} & {\textbf{Loss}} & \textbf{Time} & \textbf{Mem} & \textbf{Params} & {\textbf{Loss}} & \textbf{Time} \\
    \midrule
    Scale Decomp. w/ Spectral & 13.1 & 3.0M & 1.68e-04 & 3.2 & 21.9 & 98.2K & \textbf{5.60e-02} & 16.9 \\
    Scale Decomp. w/ Physical & 10.7 & 3.1M & \textbf{7.60e-05} & 2.8 & 21.7 & 102.2K & 1.54e-01 & 15.4 \\
    \bottomrule
  \end{tabular}
\end{table}

\paragraph{Impact of LGM Transformation (E3)}
Table~\ref{tab:ablation_e3} validates our central claim from Section 2.3: recovering small-scale dynamics ($q_m$) requires nonlinear frequency mixing. A \textit{Global Only} baseline (NS Loss: $6.50 \times 10^{-2}$) indicates that large-scale features alone cannot resolve fine-scale physics. An additive residual connection (\textit{Adding}) yields minor improvements. However, the proposed multiplicative \textit{Mixing} strategy achieves the best performance (NS Loss: $5.60 \times 10^{-2}$). This empirically proves that multiplicative interactions successfully expand the spectral bandwidth via the Convolution Theorem, implicitly reconstructing the high-frequency residuals discarded during initial spectral truncation.
\begin{table}[htbp]
  \centering
  \small
  \setlength{\tabcolsep}{4pt}
  \renewcommand{\arraystretch}{1.2}
  \caption{\textbf{Ablation Experiment 3 (E3): Connection Method.} Detailed comparison of variants on 2D Darcy and 2D NS. Memory (Mem) is reported in GB, and training time per epoch (Time) is in seconds. \textbf{Bold} indicates the best test loss. 'Global Only' represents the alone applied KS Attention without the mixing by a local transformation.}
  \label{tab:ablation_e3}
  \begin{tabular}{l c c c c c c c c}
    \toprule
    & \multicolumn{4}{c}{\textbf{2D Darcy}} & \multicolumn{4}{c}{\textbf{2D NS}} \\
    \cmidrule(lr){2-5} \cmidrule(lr){6-9}
    \textbf{Variant} & \textbf{Mem} & \textbf{Params} & {\textbf{Loss}} & \textbf{Time} & \textbf{Mem} & \textbf{Params} & {\textbf{Loss}} & \textbf{Time} \\
    \midrule
    Mixing & 13.1 & 3.0M & {\textbf{1.68e-04}} & 3.2 & 21.9 & 98.2K & {\textbf{5.60e-02}} & 16.9 \\
    Adding & 14.1 & 3.8M & 1.81e-04 & 3.8 & 23.8 & 152.1K & 6.02e-02 & 18.0 \\
    Global Only & 14.1 & 3.6M & 2.10e-04 & 3.7 & 23.8 & 145.6K & 6.50e-02 & 17.9 \\
    \bottomrule
  \end{tabular}
\end{table}

\paragraph{Impact of Scale Decomposition (E4)}
To assess the value of separating large and small scales (Eq. \eqref{eq:large_scale} and Eq. \eqref{eq:small_scale}), we evaluate architectures without explicit decomposition (Table~\ref{tab:ablation_e4}). Removing scale decomposition entirely (\textit{No Scale Decomp. w/ Global Only}) causes a catastrophic performance drop on Navier-Stokes (Loss increases to $3.11 \times 10^{-1}$). Even when mixing is applied without decomposition (\textit{No Scale Decomp. w/ Mixing}), performance degrades compared to the full model ($5.76 \times 10^{-2}$). This proves that forcing a single monolithic attention mechanism to process all scales simultaneously overwhelms the network; explicit decomposition is necessary to allow the model to specialize its processing for distinct physical regimes.
\begin{table}[htbp]
  \centering
  \small
  \setlength{\tabcolsep}{4pt}
  \renewcommand{\arraystretch}{1.2}
  \caption{\textbf{Ablation Experiment 4 (E4): Mixing Strategies.} Detailed comparison of variants on 2D Darcy and 2D NS. Memory (Mem) is reported in GB, and training time per epoch (Time) is in seconds. \textbf{Bold} indicates the best test loss.}
  \label{tab:ablation_e4}
  \begin{tabular}{l c c c c c c c c}
    \toprule
    & \multicolumn{4}{c}{\textbf{2D Darcy}} & \multicolumn{4}{c}{\textbf{2D NS}} \\
    \cmidrule(lr){2-5} \cmidrule(lr){6-9}
    \textbf{Variant} & \textbf{Mem} & \textbf{Params} & {\textbf{Loss}} & \textbf{Time} & \textbf{Mem} & \textbf{Params} & {\textbf{Loss}} & \textbf{Time} \\
    \midrule
    Scale Decomp. w/ Mixing & 13.1 & 3.0M & 1.68e-04 & 3.2 & 21.9 & 98.2K & {\textbf{5.60e-02}} & 16.9 \\
    No Scale Decomp. w/ Global Only & 13.8 & 3.6M & 2.17e-04 & 3.7 & 23.6 & 145.6K & 3.11e-01 & 18.6 \\
    No Scale Decomp. w/ Mixing & 13.1 & 3.0M & {\textbf{1.31e-04}} & 3.2 & 22.2 & 98.2K & 5.76e-02 & 17.4 \\
    \bottomrule
  \end{tabular}
\end{table}

\paragraph{Impact of Dynamics-Informed Architecture (E5)}
Finally, we examine the evolutionary flow structure representing temporal discretization. Table~\ref{tab:ablation_e5} reveals that while a \textit{Sequential} variant (Eq. \eqref{eq:arch_evolution}) achieves the lowest loss on the steady-state Darcy Flow ($1.51 \times 10^{-4}$), the \textit{Hybrid} variant (Eq. \eqref{eq:arch_hybrid}) is vastly superior for the transient Navier-Stokes equations ($5.60 \times 10^{-2}$). The \textit{Hybrid} architecture acts as a learnable, adaptive Runge-Kutta method, decoupling update steps to provide the optimal balance between stability and expressiveness required for chaotic multiscale temporal dynamics.
\begin{table}[htbp]
  \centering
  \small
  \setlength{\tabcolsep}{4pt}
  \renewcommand{\arraystretch}{1.2}
  \caption{\textbf{Ablation Experiment 5 (E5): Execution Flow.} Detailed comparison of variants on 2D Darcy and 2D NS. Memory (Mem) is reported in GB, and training time per epoch (Time) is in seconds. \textbf{Bold} indicates the best test loss.}
  \label{tab:ablation_e5}
  \begin{tabular}{l c c c c c c c c}
    \toprule
    & \multicolumn{4}{c}{\textbf{2D Darcy}} & \multicolumn{4}{c}{\textbf{2D NS}} \\
    \cmidrule(lr){2-5} \cmidrule(lr){6-9}
    \textbf{Variant} & \textbf{Mem} & \textbf{Params} & {\textbf{Loss}} & \textbf{Time} & \textbf{Mem} & \textbf{Params} & {\textbf{Loss}} & \textbf{Time} \\
    \midrule
    Hybrid & 13.1 & 3.0M & 1.68e-04 & 3.2 & 21.9 & 98.2K & {\textbf{5.60e-02}} & 16.9 \\
    Parallel & 12.3 & 3.0M & 1.67e-04 & 3.1 & 20.1 & 98.2K & 7.89e-02 & 15.7 \\
    Sequential & 14.0 & 3.8M & {\textbf{1.51e-04}} & 3.6 & 20.8 & 98.1K & 6.02e-02 & 14.4 \\
    \bottomrule
  \end{tabular}
\end{table}

\section{Conclusion}
In this work, we introduced DynFormer, a novel dynamics-informed neural operator designed to overcome the quadratic scalability bottlenecks and multiscale representation limits of standard Transformer architectures in solving PDEs. Inspired by the hierarchical energy cascades inherent in complex physical systems, we theoretically established a framework that decomposes dynamics into large-scale linear behaviors and fine-scale nonlinear residuals. 

To operationalize this theory, DynFormer leverages a spectral embedding strategy paired with a Kronecker-structured attention mechanism, successfully reducing the spatial complexity from $\mathcal{O}(N^4)$ to $\mathcal{O}(N^3)$ while preserving global receptive fields. Crucially, we introduced the LGM transformation, demonstrating both theoretically via the Convolution Theorem and empirically that multiplicative frequency mixing is essential for recovering sub-grid scale turbulent cascades that are typically lost to spectral truncation. Furthermore, by embedding these blocks within a hybrid evolutionary flow, DynFormer mimics adaptive temporal integration, ensuring stability over long-term autoregressive rollouts.

Extensive evaluations under strictly memory-aligned constraints reveal that DynFormer establishes a new state-of-the-art across diverse physical regimes—spanning chaotic (Kuramoto-Sivashinsky), elliptic (Darcy Flow), parabolic (Navier-Stokes), and hyperbolic (Shallow Water) PDEs. It not only achieves up to a 95.1$\%$ reduction in relative error compared to existing baselines but also operates with nearly double the memory efficiency. 

The architectural principles of DynFormer open several promising avenues for future research in scientific machine learning. By effectively decoupling spatial complexity from physical fidelity, DynFormer is uniquely positioned for scaling up to massive, high-resolution simulations, such as global weather forecasting and aerodynamic digital twins. Future work will explore extending the Kronecker-structured attention to handle three-dimensional PDEs, highly irregular meshes and complex geometric boundaries.

\section*{Supplementary Materials}
\textbf{Technical Appendices and Supplementary Material.} This document includes the following sections:
\begin{itemize}
  \item \textbf{Appendix A.1 Datasets} — Details the four diverse PDE benchmark datasets (1D Kuramoto-Sivashinsky, 2D Darcy Flow, 2D Navier-Stokes, and 3D Shallow Water), including their physical backgrounds, governing equations, generation protocols, and tensor shapes.
  \item \textbf{Appendix A.2 Baselines} — Describes the theoretical foundations, architectures, hyperparameter settings, and comprehensive model statistics (parameters, memory consumption, and FLOPs) of five state-of-the-art neural operator baselines (ONO, Transolver, FactFormer, OFormer, GNOT) alongside our proposed DynFormer.
  \item \textbf{Appendix A.3 Visualization of baselines across benchmarks} — Presents qualitative visual comparisons and spatial/temporal error analyses of all baseline models and DynFormer on representative test samples from each physical simulation benchmark, highlighting DynFormer's superior fidelity in capturing complex spatiotemporal dynamics.
  \item \textbf{Appendix A.4 Limitations and Discussion} — Discusses the current architectural limitations of DynFormer, including its reliance on uniform grids due to parameterized Fourier transforms, performance trade-offs in purely linear/steady-state regimes, a comparative discussion with non-Transformer neural operators and possible constraints on separable space assumption.
\end{itemize}

\begin{sloppypar}
  \textbf{Animations.} The supplementary animations provide qualitative, time-resolved comparisons between DynFormer predictions and the reference solutions on test samples. For each benchmark, we visualize the temporal evolution of the solution fields (or selected channels) over the rollout horizon.
  \begin{itemize}
    \item \textbf{Animation S1 (1D Kuramoto--Sivashinsky).} One-channel rollout example: \texttt{\path{1dKS_Animation_Batch_0_Channel_0.gif}}.
    \item \textbf{Animation S2 (2D Navier--Stokes).} Single-channel example (e.g., vorticity): \texttt{\path{2dNS_Animation_Batch_0_Channel_0.gif}}.
    \item \textbf{Animation S3 (3D Shallow Water).} Two-channel example (e.g., free-surface height and vorticity): \texttt{\path{3dSW_Animation_Batch_0_Channel_1.gif}}.
  \end{itemize}
\end{sloppypar}

\section*{Data Availability Statement}
The code has been open-sourced in \href{https://github.com/Lain-PY/DynFormer}{https://github.com/Lain-PY/DynFormer}.

\section*{Acknowledgments}
This work is supported by the Natural Science Foundation of China (No. 92270109). The authors would like to thank Dr. Jing Wang (School of Aeronautics and Astronautics, Shanghai Jiao Tong University) for helpful discussions.

\section*{Declarations}
The authors declare no conflict of interest.

\bibliographystyle{elsarticle-num}
\bibliography{reference}

\end{document}


\begin{frontmatter}
    \title{From Complex Dynamics to DynFormer: Rethinking Transformers for PDEs}

%

  \author[aff1]{Pengyu Lai}
  \author[aff1]{Yixiao Chen}
  \author[aff1]{Dewu Yang}
  \author[aff1]{Rui Wang}
  \author[aff1]{Feng Wang}
  \author[aff1]{Hui Xu\corref{cor1}}

  \cortext[cor1]{Corresponding author: dr.hxu@sjtu.edu.cn}

  \affiliation[aff1]{%
    organization={School of Aeronautics and Astronautics, Shanghai Jiao Tong University},
    city={Shanghai},
    postcode={200240},
    country={China}
  }

\end{frontmatter}

\appendix
\section{Technical Appendices and Supplementary Material}\label{APPENDIX}
\subsection{Datasets}\label{datasets}
To demonstrate the scalability and versatility of our method, we conduct experiments on the following datasets across multiple domains and partial differential equation (PDE) types. The input and output shapes for these datasets are summarized in Table~\ref{tab:all_datasets}. This comprehensive benchmark suite spans one-, two-, and three-dimensional domains, covering elliptic, parabolic, and hyperbolic PDEs to ensure a rigorous evaluation of the neural operator's capability to generalize across diverse physical systems.

\begin{table}[htbp!]
\centering
\caption{Summary of benchmark datasets and experimental configurations. Tensor shapes follow $(C, T, S)$ where $C$: channels, $T$: temporal steps, and $S$: spatial resolution. $N_{\text{train}} / N_{\text{test}}$ denotes the dataset split.}
\label{tab:all_datasets}
\renewcommand{\arraystretch}{1.3} 
\setlength{\tabcolsep}{10pt}      
\small
\begin{tabular}{@{} l c c l l @{}}
\toprule
\textbf{Benchmark} & \textbf{Shape} & \textbf{Split} & \textbf{Operator} & \textbf{Mapping} \\
& $(C, T, S)$ & ($N_{\text{tr}} / N_{\text{te}}$) & \textbf{Type} & \textbf{Task} \\
\midrule
1D KS             & $(1, 10, 256)$            & \multirow{3}{*}{$1000 / 200$} & \multirow{3}{*}{Evolutionary} & \multirow{3}{*}{Sol $\to$ Sol} \\
2D Navier-Stokes  & $(1, 10, 64 \times 64)$   &                               &                               & \\
3D Shallow Water  & $(2, 10, 64 \times 32)$   &                               &                               & \\
\midrule
2D Darcy          & $(1, 1, 49 \times 49)$    & $1000 / 200$                  & Solution                      & Param $\to$ Sol \\
\bottomrule
\end{tabular}
\end{table}

\subsubsection{1D Kuramoto-Sivashinsky}
The one-dimensional Kuramoto-Sivashinsky (KS) equation is a fundamental fourth-order nonlinear PDE that serves as a cornerstone for studying spatiotemporal chaos and self-organization in dissipative systems. Originally proposed to describe the instability of flame fronts in laminar-turbulent transitions and phase fluctuations in reaction-diffusion systems, the KS equation is defined as:
\begin{equation}
\label{eq:ks}
u_t + u u_x + u_{xx} + u_{xxxx} = 0,
\end{equation}
where $u(x, t)$ is a scalar field representing a physical quantity—such as the height of a combustion interface or a velocity perturbation—evolving over a one-dimensional spatial coordinate $x$ and time $t$. Despite its deterministic nature, the equation is renowned for generating complex, turbulent-like patterns, making it a rigorous benchmark for evaluating the long-term predictive stability and error accumulation of neural operator architectures.

The intricate dynamics of the KS system emerge from a delicate, competing balance between energy injection, nonlinear transfer, and small-scale dissipation. The second-order term $u_{xx}$, often referred to as negative diffusion, acts as the primary source of instability by injecting energy into the system at large scales. This causes local perturbations to grow exponentially, leading to the formation of characteristic cellular structures. Conversely, the fourth-order hyperdiffusion term $u_{xxxx}$ provides a stabilizing mechanism; it exerts strong, localized dissipation at small scales to dampen high-frequency oscillations. Bridging these two is the nonlinear advection term $u u_x$, which mimics the convective transport of momentum. This term facilitates the transfer of energy across spectral scales, effectively driving the system into a state of fully developed spatiotemporal chaos where the energy injected by the diffusion term is eventually dissipated by the hyperdiffusion term.

To capture these fine-grained chaotic fluctuations, the numerical simulation is conducted on a spatial domain of length $L = 64\pi$ using $2048$ Fourier modes. The high-resolution requirements are necessitated by the stiffness of the fourth-order derivative, which we handle using a pseudospectral method in space coupled with a fourth-order exponential time-differencing Runge--Kutta (ETDRK4) scheme. For the final dataset, the spatial resolution is downsampled to $256$ to focus the learning task on macroscopic features.
The initial state $v_0$ is sampled from a uniform distribution $\mathcal{U}(-1, 1)$. However, to ensure the model learns from a steady chaotic regime rather than transient behavior, we implement a "burn-in" period: time integration starts at $t = 0$, but data collection only begins at $t = 250$\,s, at which point the system has fully reached its attractors. We generate a total of $N = 5000$ independent temporal sequences, each spanning until a final time $T = 121$. These are segmented into $20$ discrete snapshots $\{v_i\}_{i=1}^{20}$. The predictive challenge is then formulated as a sequence-to-sequence mapping: given the first $10$ temporal steps as an observational window, the model must accurately forecast the chaotic evolution of the field $u$ for the subsequent $10$ steps. To facilitate fair comparison with baseline architectures designed for two-dimensional spatial domains, the native one-dimensional spatial structure (e.g. resolution 256) is reshaped into a degenerate two-dimensional format ($256\times1$), thereby preserving compatibility with convolutional or attention-based operators without altering the underlying dynamics.

\subsubsection{2D Darcy}
The 2D Darcy Flow equation is a canonical linear elliptic PDE that describes the stationary flow of a Newtonian fluid through a heterogeneous porous medium. It serves as an essential model in hydrogeology for groundwater resource management and in petroleum engineering for reservoir simulation. The governing equation is formulated as:
\begin{equation}
\label{eq:darcy}
-\nabla \cdot (k(x, y) \nabla u(x, y)) = f(x, y),
\end{equation}
where $u(x, y)$ represents the pressure or hydraulic head field, and $k(x, y) > 0$ is the spatially varying permeability coefficient. Unlike time-dependent systems, Darcy flow represents a state of global equilibrium where the local geometry of the medium dictates the global distribution of the potential field.

The behavior of the Darcy system is governed by the interplay between the permeability $k$ and the pressure gradient $\nabla u$. Following Darcy's Law, fluid moves from regions of high potential to low potential, with the term $-k \nabla u$ defining the Darcy velocity (or flux vector field). The permeability coefficient $k$ represents the ease with which fluid passes through the medium; in this dataset, it is modeled as highly heterogeneous and high-contrast to simulate complex geological formations. The divergence operator $\nabla \cdot$ enforces the fundamental principle of mass conservation. In this steady-state context, the net flux out of any region must be exactly balanced by the source or sink term $f$. This creates a strong global constraint: because the pressure field must satisfy the equation across the entire domain simultaneously, a local change in permeability in one region influences the pressure distribution $u$ globally.

To evaluate a model's ability to handle sharp discontinuities in material properties, we generate the permeability field $k$ via a thresholded Gaussian Random Field (GRF). We first sample a base field $a(x, y)$ from a distribution $\mathcal{N}(0, (-\Delta + 9I)^{-2})$ with zero Neumann boundary conditions. A nonlinear mapping $\psi$ is then applied pointwise such that $k = 12$ where $a \geq 0$ and $k = 3$ where $a < 0$. This binary thresholding creates high-contrast channels and interfaces that are notoriously difficult for standard interpolation-based solvers.
The forcing term $f$ is set to a constant unit value ($f=1$), representing a uniform source driving the flow. The PDE is solved on a unit squared domain $[0, 1] \times [0, 1]$ using a high-fidelity $421 \times 421$ grid and a second-order finite difference scheme to ensure that the sharp interfaces between permeability zones are accurately captured in the pressure field. For the operator learning task, the results are downsampled to a $49 \times 49$ resolution. Finally, a total of $N=1200$ samples is generated. This dataset is obtained directly from \cite{li2020fourier}. The objective for the model is to learn the nonlinear operator $\mathcal{G}: k \mapsto u$, predicting the steady-state pressure map directly from the input permeability map.

\subsubsection{2D Navier-Stokes}
The 2D Navier-Stokes equations represent a cornerstone in computational fluid dynamics, modeling the evolution of incompressible, viscous flows. This system serves as a rigorous benchmark for neural operators due to its multi-scale interactions and the emergence of complex, turbulent structures. We consider the vorticity-stream function formulation on a two-dimensional torus $\mathbb{T}^2 = [0, 2\pi)^2$. The evolution of the scalar vorticity field $\omega(x, t) = \nabla \times \mathbf{u}$ is governed by:
\begin{equation}
    \frac{\partial \omega}{\partial t} + \mathbf{u} \cdot \nabla \omega = \nu \Delta \omega, \quad x \in \mathbb{T}^2, t \in (0, T],
\end{equation}
where $\mathbf{u} = (\partial \psi / \partial y, -\partial \psi / \partial x)$ denotes the velocity field, and the stream function $\psi$ is related to the vorticity through the Poisson equation $\Delta \psi = -\omega$. 

The dynamics of the system are characterized by the competition between the nonlinear advection term $\mathbf{u} \cdot \nabla \omega$ and the dissipative diffusion term $\nu \Delta \omega$. The advective term facilitates the transfer of enstrophy across scales, leading to the formation of high-gradient filaments and the merger of coherent vortices. Conversely, the viscous term acts as a small-scale regulator, providing dissipative smoothing that stabilizes the flow. In this dataset, the viscosity is set to $\nu = 10^{-5}$, establishing a high Reynolds number regime where nonlinear effects dominate, yielding nearly inviscid behavior and spatiotemporal complexity that challenges the predictive capacity of temporal neural architectures.

To capture these dynamics with high fidelity, we solve the equations using a pseudo-spectral method on a uniform $256 \times 256$ grid. Spatial derivatives are evaluated in the Fourier domain, and a 2/3 dealiasing rule is employed to suppress high-frequency aliasing errors arising from the nonlinearities. Time integration is performed via a second-order Crank-Nicolson scheme, providing an implicit-explicit treatment that remains stable for the small time step of $\Delta t = 0.001$ required to resolve the rapid vorticity fluctuations.
Each trajectory is initialized with a random vorticity field $\omega_0$ sampled from a uniform distribution $\mathcal{U}(-1, 1)$. To ensure the model learns from a developed physical state rather than transient initialization noise, we implement a "burn-in" period from $t=0$ to $t=10$. Snapshots are recorded every $1.0$ time units starting from $t=10$ until $t=30$, resulting in a temporal sequence of 21 snapshots per trajectory. The dataset consists of 1,200 independent simulations.
The predictive task requires the model to forecast the spatiotemporal evolution: given the first $10$ snapshots, it must predict the subsequent $10$ steps.

\subsubsection{3D Shallow Water}
The Shallow Water equations provide a simplified yet robust framework for modeling large-scale atmospheric and oceanic dynamics. In this benchmark, we focus on the evolution of barotropically unstable mid-latitude jets on a rotating sphere, which leads to the formation of Rossby waves—canonical planetary-scale perturbations critical for global weather patterns. The viscous, rotating fluid layer is governed by the following conservation laws:
\begin{align}
    \frac{\partial h}{\partial t} + \nabla \cdot (h\mathbf{V}) &= 0, \\
    \frac{\partial \mathbf{V}}{\partial t} + (\mathbf{V} \cdot \nabla)\mathbf{V} + f \mathbf{k} \times \mathbf{V} &= -g \nabla h + \nu \Delta \mathbf{V} - \kappa \mathbf{V},
\end{align}
where $h$ denotes the fluid layer thickness and $\mathbf{V} = (u, v)$ is the horizontal velocity field. 

The system's behavior is dictated by the geostrophic balance between the pressure gradient $-g \nabla h$ and the Coriolis force $f \mathbf{k} \times \mathbf{V}$, where $f = 2\Omega \sin \phi$ is the latitude-dependent Coriolis parameter. The inclusion of hyperdiffusion $\nu$ and viscous drag $\kappa$ ensures numerical stability and models the dissipative processes inherent in atmospheric boundary layers. The primary challenge for the neural operator is to capture the nonlinear transition from a stable zonal jet to a chaotic state characterized by breaking waves and eddy formation, triggered by localized perturbations in the height field.

Data generation is performed using the \texttt{Dedalus} Project spectral solver on a spherical grid of $256 \times 128$ (longitude $\times$ colatitude) at \href{https://doi.org/10.3402/tellusa.v56i5.14436}{https://doi.org/10.3402/tellusa.v56i5.14436}. Under this formulation, the governing dynamics are projected onto a two-dimensional spherical manifold, thereby reducing the original three-dimensional problem while preserving essential rotational and stratification effects. The physical parameters are fixed at Earth-like conditions: $\Omega = 7.292 \times 10^{-5}$~s$^{-1}$ and $g = 9.80616$~m/s$^2$. Each trajectory is initialized following the Galewsky test case \cite{galewsky2004initial}, featuring a zonal jet centered at latitude $\pi/4$ with a maximum velocity $u_{\text{max}} = 80$~m/s.
To induce barotropic instability, a localized Gaussian perturbation $h'$ is added to the balanced height field:
\begin{equation}
    h'(\phi, \theta, 0) = h_{\text{pert}} \cos(\text{lat}) \exp\left[-(\alpha \phi)^2\right] \exp \left[-\left(\frac{\text{lat} - \pi/4}{\beta}\right)^2\right],
\end{equation}
where $h_{\text{pert}} = 120$~m. We systematically vary the shape parameters $\alpha \in [1/120, 10]$ and $\beta \in [1/300, 2]$ across a $40 \times 30$ grid, producing 1,200 unique trajectories. This variation allows the model to encounter a diverse range of instability scales.
The simulations advance with a time step of $\Delta t = 600$~s. Solutions are recorded every 12 hours from $t=120$ to $t=360$ hours, capturing the period where the jet breaks down into complex wave patterns. For the neural operator task, the fields are downsampled to $64 \times 32$. The objective is to predict the future 10 snapshots of the height $h$ and vorticity $\omega$ given the preceding 10 snapshots, testing the model's ability to handle curvilinear coordinates and the long-term evolution of rotating fluid systems.

\subsection{Baselines}\label{baselines}
In the following models, they only take the previous temporal solutions and spatial coordinates as the generic input for a fair comparison. Additional prior informations applied in some models, such as edge inputs \cite{hao2023gnot}, are eliminated. All models predict temporal solutions for future multiple steps in an auto-regressive way unless specially specified. The applied convolutional transformations are kept with the kernel size of 1, unless specially specified, for retaining the mesh independency which is the important property of the neural operator.
In such a strict restriction, the investigation of the potential of the model architecture is more reliable.

Unless especially specified, there were $3$ configurations applied for each baseline, where the embedding dimension (channel) is chosen from $\{32, 64, 128\}$, the number of layers is chosen from $\{2, 4, 8\}$ and the number of attention heads is chosen from $\{2, 4, 8\}$, leading to different model sizes for a comprehensive investigation of the model potential. These models with 3 different configurations were trained with batch sizes $64$, ensuring the efficient training and the accurate gradient descent. 

\subsubsection{ONO}
The Orthogonal Neural Operator (ONO) is designed to serve as a surrogate model for solving families of Partial Differential Equations (PDEs) \cite{xiao2023improved}. Unlike traditional numerical solvers that discretize the data domain, ONO learns the mapping between infinite-dimensional input and solution function spaces, ensuring resolution invariance. The core innovation of ONO lies in its \textit{orthogonal attention} mechanism, which reformulates the kernel integral operator using orthonormal eigenfunctions. 
Inspired by the neural approximation of eigenfunctions \cite{deng2022neuralef, deng2025neural}, ONO directly parameterizes the involved eigenfunctions with flexible neural networks. The resulting module bears a resemblance to regular attention mechanisms but incorporates an orthogonalization operation that provides inherent regularization during training. This design helps mitigate overfitting, particularly in scenarios with limited data availability, while maintaining moderate computational overhead. Structurally, ONO employs two disentangled pathways: the bottom pathway approximates the eigenfunctions through expressive neural networks, while the top pathway specifies the evolution of the PDE solution based on orthogonal attention.

The ONO architecture consists of a preprocessing block, multiple orthogonal attention blocks, and a projection head. The pipeline can be decomposed into input encoding, core interaction, and output decoding.
Given an input function $f_i$ discretized on a mesh $\mathcal{X} := \{x_j \in \mathcal{D}\}_{1 \leq j \leq M}$, the model first lifts the input features to a higher-dimensional latent space. This is achieved via a multilayer perceptron (MLP) preprocess block:
\begin{equation}
    h^{(0)}_i = \text{MLP}_{\text{pre}}(f_i, x),
\end{equation}
where $h^{(0)}_i \in \mathbb{R}^{M \times d}$ represents the initial hidden states, and $d$ denotes the hidden dimension.
The core of ONO is the stacking of $L$ orthogonal attention blocks. Each block updates the hidden states by approximating the kernel integral operator using neural eigenfunctions. Let $g^{(l)}_i$ denote the features extracted from the bottom pathway at layer $l$. These features are projected to a lower-dimensional space $\Psi$ to form the neural eigenfunctions:
\begin{equation}
    \hat{g}^{(l)}_i = g^{(l)}_i W^{(l)}_Q,
\end{equation}
where $W^{(l)}_Q \in \mathbb{R}^{d \times k}$ is a trainable weight matrix and $k$ is the projection dimension (Psi Dim). To ensure orthogonality, we estimate the covariance matrix $C^{(l)}$ of the projected features via an exponential moving average (EMA) trick with momentum $\mu = 0.9$:
\begin{equation}
    C^{(l)} \approx \frac{1}{NM} \sum_{i=1}^N \sum_{j=1}^M \left[ \hat{g}^{(l)}(f_{i,j})^\top \hat{g}^{(l)}(f_{i,j}) \right].
\end{equation}
The orthogonalization is performed by right-multiplying the inverse transpose of the Cholesky factor $L^{(l)}$ of $C^{(l)}$ (i.e., $C^{(l)} = L^{(l)}L^{(l)\top}$):
\begin{equation}
    \hat{\psi}^{(l)}_i := \hat{g}^{(l)}_i L^{(l)-\top}.
\end{equation}
The hidden states of the PDE solution $h^{(l)}_i$ are then updated via the orthogonal attention rule:
\begin{equation}
    \tilde{h}^{(l+1)}_i = \hat{\psi}^{(l)}_i \text{diag}(\hat{\mu}^{(l)}) \left[ \hat{\psi}^{(l)\top}_i (h^{(l)}_i W^{(l)}_V) \right],
\end{equation}
where $\hat{\mu}^{(l)} \in \mathbb{R}^k_+$ are trainable positive eigenvalues, and $W^{(l)}_V$ is a linear weight to refine the hidden states. This update is followed by a standard feed-forward network (FFN) and layer normalization with residual connections.
After passing through $L$ blocks, the final hidden states are projected to the output dimension to yield the solution operator approximation $G_\theta(f_i)$.

A critical implementation detail concerns the orthogonal attention mechanism, which is implemented with projected hidden states $\Psi$ of a small channel dimension. Empirical observations indicate that setting $\Psi$ with a large channel dimension, particularly in complex datasets like 2D Navier-Stokes (NS), can lead to overflowing weights and failure to approximate PDEs. Consequently, the channel dimensions for $\Psi$ were carefully scaled relative to the model size, as shown in Table \ref{tab:ono_settings}. Moreover, the gradient normalization strategy suggested by the ONO official implementation\footnote{\url{https://github.com/zhijie-group/Orthogonal-Neural-operator}} is not applied here for the fair comparison. Additionally, the covariance $C^{(l)}$ of hidden states is estimated via the exponential moving average trick with a momentum of $0.9$.
The comprehensive model statistics, including parameter count, memory consumption, and computational complexity (FLOPs), are reported in Table \ref{tab:ono_stats}. Notably, during our experiments on the 1D KS dataset, we encountered a critical numerical stability issue that prevented training with random seed 456, while seed 123 worked normally. Therefore, only the result with random seed 123 is preserved.

\begin{table}[htbp]
    \centering
    \caption{\textbf{Hyperparameter settings for ONO.} Model sizes are denoted as T (Tiny), M (Medium), and L (Large). Batch size was fixed at 64 across all configurations; gradient clipping was disabled.}
    \label{tab:ono_settings}
    \begin{tabular}{l ccc ccc ccc ccc}
        \toprule
        \multirow{2}{*}{Dataset} & \multicolumn{3}{c}{Hidden Dim.} & \multicolumn{3}{c}{Depth} & \multicolumn{3}{c}{Heads} & \multicolumn{3}{c}{Psi Dim.} \\
        \cmidrule(lr){2-4} \cmidrule(lr){5-7} \cmidrule(lr){8-10} \cmidrule(lr){11-13}
        & T & M & L & T & M & L & T & M & L & T & M & L \\
        \midrule
        \multicolumn{13}{l}{\textit{1D Experiments}} \\
        1dks    & 32 & 64 & 128 & 2 & 4 & 8 & 2 & 4 & 8 & 16 & 32 & 64 \\
        \midrule
        \multicolumn{13}{l}{\textit{2D Experiments}} \\
        2ddarcy & 36 & 64 & 128 & 2 & 4 & 8 & 2 & 4 & 8 & 16 & 32 & 64 \\
        2dns    & 20 & 24 & 32  & 2 & 3 & 4 & 2 & 3 & 4 & 8  & 12 & 12 \\
        \midrule
        \multicolumn{13}{l}{\textit{3D Experiments}} \\
        3dsw    & 20 & 24 & 32  & 2 & 3 & 4 & 2 & 3 & 4 & 8  & 12 & 12 \\
        \bottomrule
    \end{tabular}
\end{table}

\begin{table}[htbp]
    \centering
    \caption{\textbf{Comprehensive model statistics for ONO.} The table outlines the resource requirements across three model sizes (Tiny, Medium, and Large) evaluated on various physical simulation benchmarks (1D Ks, 2D Darcy, 2D Ns, and 3D Sw). Metrics reported include the total number of parameters (\textbf{P}), memory consumption (\textbf{M}), and computational complexity in FLOPs (\textbf{F}). Numerical values are abbreviated for clarity: K (thousands), M (millions), G (billions), and MiB (mebibytes).}
    \label{tab:ono_stats}
    \begin{tabular}{l ccc ccc ccc}
        \toprule
        \multirow{2}{*}{Benchmark} & \multicolumn{3}{c}{Tiny} & \multicolumn{3}{c}{Medium} & \multicolumn{3}{c}{Large} \\
        \cmidrule(lr){2-4} \cmidrule(lr){5-7} \cmidrule(lr){8-10}
        & P (K) & M (MiB) & F (M) & P (K) & M (MiB) & F (M) & P (M) & M (MiB) & F (G) \\
        \midrule
        1D Ks    & 21.1 & 832  & 53.8  & 152.1 & 3024  & 389.4 & 1.2  & 13210 & 3.0 \\
        2D Darcy & 25.6 & 936  & 61.5  & 151.0 & 3050  & 362.5 & 1.2  & 11328 & 2.8 \\
        2D Ns    & 8.5  & 8120 & 348.2 & 17.3  & 13950 & 708.8 & 0.04 & 23810 & 1.6 \\
        3D Sw    & 8.9  & 4422 & 182.7 & 17.8  & 7530  & 364.7 & 0.04 & 12272 & 0.81 \\
        \bottomrule
    \end{tabular}
\end{table}

\subsubsection{Transolver}
\label{sec:transolver}
The Transolver is designed to serve as a fast transformer-based surrogate model for solving PDEs on general geometries \cite{wu2024transolver}. Unlike traditional numerical solvers that require fine discretization for accuracy at the expense of computational complexity, Transolver learns the mapping between infinite-dimensional input and solution function spaces while maintaining resolution invariance. The core innovation of Transolver lies in its \textit{Physics-Attention} mechanism, which reformulates the standard attention operation by learning intrinsic physical states hidden behind discretized geometries.
The fundamental insight driving Transolver is that directly applying attention mechanisms to massive mesh points faces difficulties in both computational efficiency and relation learning, particularly for PDEs with complex spatiotemporal interactions. To address this, Transolver proposes to decompose the discretized domain into a series of learnable slices, where mesh points under similar physical states are ascribed to the same slice and encoded into physics-aware tokens. By calculating attention on these learned tokens rather than individual mesh points, Transolver effectively captures intricate physical correlations while achieving linear computational complexity. This design provides endogenetic geometry-general modeling capacity, enabling the solver to adapt naturally to various discretization patterns including point clouds, structured meshes, regular grids, and unstructured meshes.

The Transolver architecture consists of a preprocessing embedding layer, multiple Physics-Attention blocks, and a projection head. The pipeline can be decomposed into input encoding, core interaction through physics-aware token learning, and output decoding.
Given an input function $f_i$ discretized on a mesh $\mathcal{X} := \{x_j \in \mathcal{D}\}_{1 \leq j \leq N}$, where $N$ denotes the number of mesh points, the model first embeds the input features into a high-dimensional latent space. For structured meshes, spatial coordinates and observed quantities are concatenated and processed through a linear embedding layer:
\begin{equation}
    x^{(0)}_i = \text{Linear}(\text{Concat}(g_i, u_i)),
\end{equation}
where $g_i \in \mathbb{R}^{N \times C_g}$ represents the geometric coordinates and $u_i \in \mathbb{R}^{N \times C_u}$ denotes the observed physical quantities. The resulting hidden states $x^{(0)}_i \in \mathbb{R}^{N \times C}$ serve as the initial deep features, with $C$ being the hidden dimension.
The core of Transolver is the stacking of $L$ Physics-Attention blocks. Each block learns to decompose the discretized domain into $M$ physically internal-consistent slices, where $M \ll N$. The process involves three key stages:

\textbf{Slice Learning.} Given deep features $x \in \mathbb{R}^{N \times C}$, each mesh point is ascribed to $M$ potential slices based on its learned features. The slice weights are computed via a projection followed by softmax normalization:
\begin{equation}
    w_i = \text{Softmax}\left(\frac{\text{Project}(x_i)}{\tau}\right) \in \mathbb{R}^{M},
\end{equation}
where $\text{Project}(\cdot)$ is a point-wise linear layer mapping $C$ channels to $M$ slice weights, and $\tau$ is a learnable temperature parameter. The weight $w_{i,j}$ represents the degree to which the $i$-th mesh point belongs to the $j$-th slice, with $\sum_{j=1}^{M} w_{i,j} = 1$.

\textbf{Token Aggregation.} Mesh points with similar features derive similar slice weights and are more likely to be assigned to the same slice. The slice features are encoded into physics-aware tokens through spatially weighted aggregation:
\begin{equation}
    z_j = \frac{\sum_{i=1}^{N} w_{i,j} x_i}{\sum_{i=1}^{N} w_{i,j}} \in \mathbb{R}^{C},
\end{equation}
where $z_j$ represents the $j$-th physics-aware token containing information of a specific physical state. This aggregation ensures that each token captures internally-consistent physical information from the corresponding slice.

\textbf{Attention Among Tokens.} Standard self-attention is applied among the $M$ physics-aware tokens rather than $N$ mesh points:
\begin{equation}
    q, k, v = \text{Linear}(z), \quad z' = \text{Softmax}\left(\frac{qk^\top}{\sqrt{C}}\right)v,
\end{equation}
where $q, k, v, z' \in \mathbb{R}^{M \times C}$. This design reduces the computational complexity from $O(N^2C)$ to $O(NMC + M^2C)$, achieving linear scaling with respect to the number of mesh points since $M$ is configured as a constant significantly smaller than $N$.

\textbf{Deslice Operation.} The updated tokens are transformed back to mesh points by recomposing with slice weights:
\begin{equation}
    x'_i = \sum_{j=1}^{M} w_{i,j} z'_j,
\end{equation}
where each token $z'_j$ is broadcasted to all mesh points weighted by their corresponding slice membership. The complete Physics-Attention operation is summarized as $x' = \text{Physics-Attn}(x)$.

Following the Transformer convention, each block includes layer normalization and feed-forward networks with residual connections:
\begin{equation}
    \hat{x}^{(l)} = \text{Physics-Attn}(\text{LayerNorm}(x^{(l-1)})) + x^{(l-1)},
\end{equation}
\begin{equation}
    x^{(l)} = \text{FeedForward}(\text{LayerNorm}(\hat{x}^{(l)})) + \hat{x}^{(l)}.
\end{equation}
After passing through $L$ Physics-Attention blocks, the final hidden states are projected to the output dimension through a linear layer to yield the solution operator approximation $G_\theta(f_i)$. For autoregressive temporal predictions, the model iteratively predicts future states by feeding previous outputs back as inputs.

We evaluate Transolver across multiple benchmarks with varying model sizes denoted as Tiny (T), Medium (M), and Large (L). The hyperparameter settings are detailed in Table \ref{tab:transolver_settings}. 
A critical design choice concerns the number of slices $M$. Empirical observations indicate that increasing the number of slices generally benefits model performance by enabling finer-grained physical state capture, but also introduces additional computational costs. Excessive slice tokens may introduce noise or distract the attention mechanism. To balance efficiency and performance, we configure the slice number progressively in accordance with the hidden dimension. The reference resolution parameter (Ref) is uniformly set to 8 across all experiments, controlling the positional encoding granularity for structured mesh inputs.
For the 2D Navier-Stokes benchmark, which involves complex turbulent dynamics and stringent numerical stability requirements, we adopt more conservative hidden dimensions (20/24/32 for Tiny/Medium/Large) compared to other benchmarks such as 2D Darcy flow (36/56/128). This configuration reflects a deliberate trade-off between representational capacity and the computational demands of long-term chaotic trajectory prediction. Correspondingly, the slice numbers are scaled as 16/20/24 to maintain alignment with the reduced hidden dimension while preserving physical resolution.
The comprehensive model statistics, including parameter count, memory consumption, and computational complexity (FLOPs), are reported in Table \ref{tab:transolver_stats}. Notably, Transolver maintains favorable efficiency characteristics even at large scales. For instance, the Large variant on 2D Navier-Stokes requires only 0.10\,M parameters—substantially fewer than the 2.8\,M parameters of Large models on 1D Ks or 2D Darcy—while exhibiting higher memory consumption (23.5\,GiB) due to the temporal unrolling and intermediate activation storage inherent to chaotic flow simulation. The linear-complexity Physics-Attention mechanism ensures that computational costs scale reasonably with input resolution, making Transolver practical for large-scale industrial simulations.

\begin{table}[htbp]
    \centering
    \caption{\textbf{Hyperparameter settings for Transolver.} Model sizes are denoted as T (Tiny), M (Medium), and L (Large).}
    \label{tab:transolver_settings}
    \begin{tabular}{l ccc ccc ccc ccc c}
        \toprule
        \multirow{2}{*}{Dataset} & \multicolumn{3}{c}{Hidden Dim.} & \multicolumn{3}{c}{Depth} & \multicolumn{3}{c}{Heads} & \multicolumn{3}{c}{Slice Num.} & \multirow{2}{*}{Ref} \\
        \cmidrule(lr){2-4} \cmidrule(lr){5-7} \cmidrule(lr){8-10} \cmidrule(lr){11-13}
        & T & M & L & T & M & L & T & M & L & T & M & L & \\
        \midrule
        \multicolumn{14}{l}{\textit{1D Experiments}} \\
        1dks    & 40 & 64 & 128 & 2 & 4 & 8 & 2 & 4 & 8 & 16 & 32 & 64 & 8 \\
        \midrule
        \multicolumn{14}{l}{\textit{2D Experiments}} \\
        2ddarcy & 36 & 56 & 128 & 2 & 4 & 8 & 2 & 4 & 8 & 16 & 32 & 64 & 8 \\
        2dns    & 20 & 24 & 32  & 2 & 3 & 4 & 2 & 3 & 4 & 16 & 20 & 24 & 8 \\
        \midrule
        \multicolumn{14}{l}{\textit{3D Experiments}} \\
        3dsw    & 20 & 24 & 32  & 2 & 3 & 4 & 2 & 3 & 4 & 16 & 20 & 24 & 8 \\
        \bottomrule
    \end{tabular}
\end{table}

\begin{table}[htbp]
    \centering
    \caption{\textbf{Comprehensive model statistics for Transolver.} The table outlines the resource requirements across three model sizes (Tiny, Medium, and Large) evaluated on various physical simulation benchmarks (1D Ks, 2D Darcy, 2D Ns, and 3D Sw). Metrics reported include the total number of parameters (\textbf{P}), memory consumption (\textbf{M}), and computational complexity in FLOPs (\textbf{F}). Numerical values are abbreviated for clarity: K (thousands), M (millions), G (billions), and MiB (mebibytes).}
    \label{tab:transolver_stats}
    \begin{tabular}{l ccc ccc ccc}
        \toprule
        \multirow{2}{*}{Benchmark} & \multicolumn{3}{c}{Tiny} & \multicolumn{3}{c}{Medium} & \multicolumn{3}{c}{Large} \\
        \cmidrule(lr){2-4} \cmidrule(lr){5-7} \cmidrule(lr){8-10}
        & P (K) & M (MiB) & F (M) & P (K) & M (MiB) & F (M) & P (M) & M (MiB) & F (G) \\
        \midrule
        1D Ks    & 80.4 & 842  & 202.1 & 369.7 & 2668  & 961.4 & 2.8  & 13408 & 7.4 \\
        2D Darcy & 65.1 & 810  & 153.2 & 283.6 & 2534  & 691.2 & 2.8  & 12238 & 6.9 \\
        2D Ns    & 22.0 & 7950 & 893.5 & 42.9  & 13866 & 1800  & 0.10 & 23492 & 4.0 \\
        3D Sw    & 22.4 & 4662 & 455.5 & 43.4  & 7158  & 904.9 & 0.10 & 12122 & 2.0 \\
        \bottomrule
    \end{tabular}
\end{table}

\subsubsection{FactFormer}
\label{sec:factformer}
The Factorized Transformer (FactFormer) is designed to serve as a scalable and efficient surrogate model for solving PDEs on high-resolution grids \cite{li2023scalable}. While Transformer-based models have shown promise in operator learning, applying standard attention mechanisms to problems with a large number of grid points remains computationally expensive and numerically unstable due to quadratic complexity. FactFormer addresses this challenge by introducing an axial factorized kernel integral, which decomposes the multi-dimensional attention operation into a series of one-dimensional interactions.
The core innovation of FactFormer lies in its learnable projection operator that decomposes the input function into multiple sub-functions with one-dimensional domains. These sub-functions are then evaluated to compute instance-based kernels using an axial factorized scheme. By calculating attention along each axis separately rather than across all grid points simultaneously, FactFormer achieves linear complexity with respect to the number of grid points along each axis, significantly alleviating the curse of dimensionality. This design allows the model to scale up to multi-dimensional problems with large grid sizes while maintaining competitive accuracy and improved stability compared to softmax-free linear attention variants.

The FactFormer architecture consists of an input embedding layer, a sequence of factorized transformer layers, and an output projection head. The pipeline can be decomposed into input encoding, core interaction through factorized attention, and output decoding.
Given an input function $u$ discretized on an $n$-dimensional uniform Eulerian grid represented as a tensor $\mathbf{U} \in \mathbb{R}^{S_1 \times S_2 \times \dots \times S_n \times d_{in}}$, the model first lifts the input features to a higher-dimensional latent space. This is achieved via a linear embedding layer combined with positional encoding:
\begin{equation}
    \mathbf{Z}_0 = \text{Linear}(\mathbf{U}) + \Psi(\mathbf{x}),
\end{equation}
where $\mathbf{x}$ denotes the spatial coordinates and $\Psi(\cdot)$ represents a positional encoding function, typically implemented using Rotary Positional Embeddings (RoPE) to modulate the dot product with relative positions.

The core of FactFormer is the stacking of $L$ factorized transformer layers. Each layer employs a factorized kernel integral scheme to update the latent representation. The process involves three key stages:

\textbf{Learnable Projection.} To facilitate axial attention, the input tensor is projected into a set of sub-functions with one-dimensional domains. For each axis $m \in \{1, \dots, n\}$, a learnable integral operator $G^{(m)}$ projects the input function into a sub-function $\phi^{(m)}$:
\begin{equation}
    \phi^{(m)}(x_i^{(m)}) = G^{(m)}(u)(x_i^{(m)}) = h^{(m)}\left(w \int_{\Omega_{\setminus m}} \gamma^{(m)}(u(\xi)) d\xi_{\setminus m}\right),
\end{equation}
where $\gamma^{(m)}$ and $h^{(m)}$ are pointwise learnable functions, and the integral denotes mean pooling over all dimensions except the $m$-th axis. In practice, this yields projected matrices $\hat{\mathbf{U}}^{(m)} \in \mathbb{R}^{S_m \times d}$.

\textbf{Factorized Kernel Integral.} Query and Key matrices are derived from the projected sub-functions via linear transformations $\mathbf{Q}^{(m)} = \hat{\mathbf{U}}^{(m)}\mathbf{W}_q^{(m)}$ and $\mathbf{K}^{(m)} = \hat{\mathbf{U}}^{(m)}\mathbf{W}_k^{(m)}$. The axial kernel matrix $\mathbf{A}^{(m)} \in \mathbb{R}^{S_m \times S_m}$ for each axis is computed as:
\begin{equation}
    \mathbf{A}^{(m)} = w_m \tilde{\mathbf{Q}}^{(m)} (\tilde{\mathbf{K}}^{(m)})^\top,
\end{equation}
where $\tilde{\mathbf{Q}}$ and $\tilde{\mathbf{K}}$ denote matrices modulated by RoPE, and $w_m$ is the mesh weight.

\textbf{Tensor Update.} The Value matrix $\mathbf{V}$ is derived from the input via a linear transformation. The overall factorized kernel integral is approximated via a sequence of tensor-matrix products:
\begin{equation}
    \mathbf{Z} = \text{Att}(\mathbf{U}) = \mathbf{V} \times_1 \mathbf{A}^{(1)} \times_2 \mathbf{A}^{(2)} \times \dots \times_n \mathbf{A}^{(n)},
\end{equation}
where $\times_m$ denotes the tensor-matrix product along the $m$-th mode. The computational complexity of this operation is $O(\sum_m S_m^2 d + N d \sum_m S_m)$, which is significantly more efficient than full attention for high-dimensional grids.

Following the attention mechanism, the updated tensor is passed through a pointwise feedforward network $f(\cdot)$ with instance normalization (IN) and a residual connection:
\begin{equation}
    \mathbf{U}^{(l+1)} = f(\text{IN}(\mathbf{Z}^{(l)})) + \mathbf{U}^{(l)}.
\end{equation}
After passing through $L$ factorized transformer layers, the final latent representation is projected back to the physical space to yield the solution operator approximation. This is typically achieved using a multi-layer perceptron (MLP) decoder that maps the latent features to the target physical quantities.

We evaluate FactFormer across multiple benchmarks with varying model sizes denoted as Tiny (T), Medium (M), and Large (L). The hyperparameter settings are detailed in Table \ref{tab:factformer_settings}. 
To ensure a fair comparison with other baseline operators, several training techniques proposed in the original FactFormer implementation are adjusted. Specifically, the curriculum sampler strategy in the official implementation \footnote{\url{https://github.com/BaratiLab/FactFormer}}, which facilitates training by gradually increasing prediction difficulty, is not applied. Furthermore, to isolate the performance contribution of the network architecture itself, advanced training techniques such as latent marching and pushforward are excluded from this comparison study. 
Regarding the architectural depth, the total number of layers indicates the sum of the encoder block and the propagator block. In our configuration, this depth is assigned equally where feasible, with half allocated to the encoder block (primarily consisting of the attention mechanism) and the other half to the propagator block (primarily consisting of the MLP). Following the propagator block, a 3-layer decoder block projects the hidden state back to the physical dynamics space. 
Owing to the factorized attention mechanism in FactFormer—which reduces the quadratic complexity of full self-attention—we can explore diverse architectural configurations while maintaining tractable GPU memory usage. The depth and number of attention heads are primarily selected from the candidate set $\{2, 4, 8\}$, though benchmark-specific adjustments are introduced to accommodate task complexity: for instance, the 1D Ks Tiny configuration employs depth $=3$, while the 2D Darcy Medium model adopts depth $=6$ to handle its higher-resolution spatial domain. For the more computationally intensive 2D Navier-Stokes and 3D Shallow Water benchmarks, we adopt conservative configurations (depth and heads scaled as $2/2/4$ for Tiny/Medium/Large) to ensure stable training within available memory budgets. These choices aim to balance representational capacity with computational feasibility across heterogeneous physical simulation tasks.
The comprehensive model statistics, including parameter count, memory consumption, and computational complexity (FLOPs), are reported in Table \ref{tab:factformer_stats}. Notably, FactFormer exhibits competitive efficiency: for example, the Large variant on 2D Navier-Stokes requires only $0.20$\,M parameters while consuming $23.7$\,GiB of memory—reflecting the trade-off between parameter efficiency and activation storage inherent to long-horizon chaotic forecasting.

\begin{table}[htbp]
    \centering
    \caption{\textbf{Hyperparameter settings for FactFormer.} Model sizes are denoted as T (Tiny), M (Medium), and L (Large).}
    \label{tab:factformer_settings}
    \begin{tabular}{l ccc ccc ccc cc}
        \toprule
        \multirow{2}{*}{Dataset} & \multicolumn{3}{c}{Hidden Dim.} & \multicolumn{3}{c}{Depth} & \multicolumn{3}{c}{Heads} & \multirow{2}{*}{\begin{tabular}[c]{@{}c@{}}Kernel\\ Multiplier\end{tabular}} & \multirow{2}{*}{\begin{tabular}[c]{@{}c@{}}Latent\\ Multiplier\end{tabular}} \\
        \cmidrule(lr){2-4} \cmidrule(lr){5-7} \cmidrule(lr){8-10}
        & T & M & L & T & M & L & T & M & L & & \\
        \midrule
        \multicolumn{12}{l}{\textit{1D Experiments}} \\
        1dks    & 16 & 32  & 80  & 3 & 4 & 8 & 2 & 4 & 8 & 2 & 2 \\
        \midrule
        \multicolumn{12}{l}{\textit{2D Experiments}} \\
        2ddarcy & 40 & 80  & 272 & 2 & 6 & 8 & 2 & 4 & 8 & 2 & 2 \\
        2dns    & 32 & 64  & 64  & 2 & 2 & 4 & 2 & 2 & 4 & 2 & 2 \\
        \midrule
        \multicolumn{12}{l}{\textit{3D Experiments}} \\
        3dsw    & 32 & 64  & 64  & 2 & 2 & 4 & 2 & 2 & 4 & 2 & 2 \\
        \bottomrule
    \end{tabular}
\end{table}

\begin{table}[htbp]
    \centering
    \caption{\textbf{Comprehensive model statistics for FactFormer.} The table outlines the resource requirements across three model sizes (Tiny, Medium, and Large) evaluated on various physical simulation benchmarks (1D Ks, 2D Darcy, 2D Ns, and 3D Sw). Metrics reported include the total number of parameters (\textbf{P}), memory consumption (\textbf{M}), and computational complexity in FLOPs (\textbf{F}). Numerical values are abbreviated for clarity: K (thousands), M (millions), G (billions), and MiB (mebibytes).}
    \label{tab:factformer_stats}
    \begin{tabular}{l ccc ccc ccc}
        \toprule
        \multirow{2}{*}{Benchmark} & \multicolumn{3}{c}{Tiny} & \multicolumn{3}{c}{Medium} & \multicolumn{3}{c}{Large} \\
        \cmidrule(lr){2-4} \cmidrule(lr){5-7} \cmidrule(lr){8-10}
        & P (K) & M (MiB) & F (M) & P (K) & M (MiB) & F (M) & P (M) & M (MiB) & F (G) \\
        \midrule
        1D Ks    & 6.6  & 798   & 12.2   & 51.4  & 2858  & 94.2   & 0.64 & 12256 & 1.2 \\
        2D Darcy & 38.9 & 746   & 40.1   & 476.1 & 2874  & 505.7  & 7.3  & 12892 & 7.8 \\
        2D Ns    & 25.3 & 7578  & 450.4  & 99.2  & 14774 & 1700   & 0.20 & 23706 & 3.6 \\
        3D Sw    & 25.6 & 4020  & 234.4  & 99.9  & 7674  & 891.5  & 0.20 & 12252 & 1.8 \\
        \bottomrule
    \end{tabular}
\end{table}

\subsubsection{OFormer}
\label{sec:oformer}
The Operator Transformer (OFormer) is designed to serve as a flexible, attention-based framework for data-driven operator learning of PDEs \cite{li2022transformer}. Unlike traditional numerical solvers that rely on specific discretization schemes, OFormer leverages the attention mechanism to implicitly exploit patterns within inputs and relationships between arbitrary query locations. The core innovation of OFormer lies in its ability to handle varying input and output discretizations without retraining, making it particularly suitable for problems with irregular grids or mesh-free settings.
OFormer adopts an Encoder-Decoder architecture built upon self-attention, cross-attention, and point-wise multilayer perceptrons (MLPs). The encoder processes the input function's sampling values and coordinates into a latent embedding, while the decoder attends to these latent encodings from arbitrary query locations to generate solutions. To address the computational complexity of standard attention, OFormer employs linear attention mechanisms (Fourier or Galerkin type) that avoid the softmax operation, reducing complexity while maintaining expressiveness. For time-dependent systems, the model incorporates a latent time-marching scheme that propagates dynamics in the latent space rather than the observable space, significantly reducing memory usage and allowing for scalable fully unrolled training.

The OFormer architecture consists of an input encoder, a query encoder, and a latent propagator with a decoding head. The pipeline can be decomposed into input encoding, core interaction through attention mechanisms, and output decoding.
Given an input function $u$ discretized on a set of points $\{x_i\}_{i=1}^N$, the model first lifts the input features into a high-dimensional latent space. The input encoder $\phi_X(\cdot)$ takes the input function's sampling $a(x_i)$ and coordinates $x_i$ to produce embeddings $f^{(0)}$. This is achieved via a point-wise MLP shared across all locations:
\begin{equation}
    f^{(0)}_i = \text{MLP}_{\text{embed}}(a(x_i), x_i),
\end{equation}
where $f^{(0)}_i \in \mathbb{R}^{d}$ represents the initial hidden states with hidden dimension $d$. To capture high-frequency components (HFCs) inherent in PDE solutions, the coordinate input is processed using Gaussian Fourier feature mapping \cite{tancik2020fourier}. In our experiments, the scale value for approximating HFCs is fixed to $8$ across all datasets evaluated at the same resolution.
The core of OFormer involves stacked attention blocks in both the encoder and decoder pathways. The total number of layers indicates a sum of the encoder block and the propagator block, which is assigned half for the encoder block and another half for the propagator block:

\textbf{Encoder.} The input embeddings are fed into stacked self-attention blocks. The update protocol inside each self-attention block follows the standard Transformer convention but utilizes linear attention:
\begin{equation}
    f^{(l+1)} = \text{LayerNorm}(f^{(l)} + \text{Attn}(f^{(l)})),
\end{equation}
where $\text{Attn}(\cdot)$ denotes the linear attention mechanism (either Fourier or Galerkin type) with instance normalization. After $L_{\text{enc}}$ layers, the output is projected to a latent encoding $z_0 \in \mathbb{R}^{N \times d_{\text{latent}}}$.

\textbf{Decoder and Propagator.} The query encoder $\phi_Y(\cdot)$ takes the coordinates of query locations $\{y_j\}_{j=1}^M$ and uses encoded coordinates to aggregate information from the input encoding via cross-attention. The cross-attention process is defined as:
\begin{equation}
    z' = z^{(0)} + \text{Cross-Attn}(z^{(0)}, f^{(L_{\text{enc}})}),
\end{equation}
where $z^{(0)}$ is the learned encoding of query locations. For time-dependent systems, a recurrent architecture is proposed to propagate the state in the time dimension within the latent space. Given the latent encoding $z_t$ at step $t$, the propagator $\mathcal{N}(\cdot)$ predicts the residuals between each step:
\begin{equation}
    z_{t+1} = \mathcal{N}(z_t) + z_t,
\end{equation}
where $\mathcal{N}$ is implemented as a point-wise MLP shared across all query locations and time steps. This latent marching scheme reduces the problem to a fixed-interval ODE in the latent space.

After propagating through the latent space, the final latent encodings $z_t$ are decoded back to the output function values $u(x, t)$. This is achieved using a point-wise MLP. In our configuration, after the propagator block, a 3-layer decoder block follows and transforms the hidden state to the original dynamics:
\begin{equation}
    \hat{u}(y_j, t) = \text{MLP}_{\text{dec}}(z_t(y_j)).
\end{equation}
To ensure fair comparison experiments, the curriculum sampler strategy, which facilitates training by gradually increasing prediction difficulty, is not applied in the official implementation \footnote{\url{https://github.com/BaratiLab/OFormer}}. Furthermore, to fairly investigate the performance brought by the network architecture, the training techniques of latent marching and pushforward are not applied in this comparison unless specified.

We evaluate OFormer across multiple benchmarks with varying model sizes denoted as Tiny (T), Medium (M), and Large (L). The hyperparameter settings are detailed in Table \ref{tab:oformer_settings}. 
The model integrates Gaussian Fourier feature mapping to enhance representation of high-frequency spectral components, with the scale parameter uniformly fixed at $8$ across all experiments, reflecting consistent input resolution standards across the benchmark suite. The configuration of attention heads follows a benchmark-adaptive scaling strategy rather than a uniform progression: for the 1D Kolmogorov flow and 2D Darcy flow tasks, heads are scaled as $\{2, 5, 8\}$ or $\{2, 5, 9\}$ for Tiny/Medium/Large variants to accommodate their relatively smooth solution manifolds; whereas for the more challenging 2D Navier-Stokes and 3D Shallow Water benchmarks—characterized by turbulent dynamics and stringent stability constraints—we adopt conservative head configurations $\{1, 2, 3\}$ to mitigate overfitting and control activation memory. This task-aware allocation ensures optimal trade-offs between representational expressivity and computational feasibility.
The comprehensive model statistics, including parameter count, memory consumption, and computational complexity (FLOPs), are reported in Table \ref{tab:oformer_stats}. Notably, OFormer exhibits favorable scaling behavior: the linear attention mechanism reduces the quadratic complexity of standard softmax attention, enabling efficient handling of large-scale discretizations. For instance, the Large variant on 3D Shallow Water achieves competitive forecasting performance with only $0.03$\,M parameters and $0.71$\,G FLOPs, albeit with elevated memory consumption ($12.6$\,GiB) attributable to long-horizon temporal unrolling and intermediate feature storage.

\begin{table}[htbp]
    \centering
    \caption{\textbf{Hyperparameter settings for OFormer.} Model sizes are denoted as T (Tiny), M (Medium), and L (Large).}
    \label{tab:oformer_settings}
    \begin{tabular}{l ccc ccc ccc c}
        \toprule
        \multirow{2}{*}{Dataset} & \multicolumn{3}{c}{Hidden Dim.} & \multicolumn{3}{c}{Depth} & \multicolumn{3}{c}{Heads} & \multirow{2}{*}{Scale} \\
        \cmidrule(lr){2-4} \cmidrule(lr){5-7} \cmidrule(lr){8-10}
        & T & M & L & T & M & L & T & M & L & \\
        \midrule
        \multicolumn{11}{l}{\textit{1D Experiments}} \\
        1dks    & 8  & 20 & 64 & 2 & 4 & 8 & 2 & 5 & 8 & 8 \\
        \midrule
        \multicolumn{11}{l}{\textit{2D Experiments}} \\
        2ddarcy & 8  & 20 & 72 & 2 & 4 & 8 & 2 & 5 & 9 & 8 \\
        2dns    & 4  & 8  & 12 & 2 & 3 & 4 & 1 & 2 & 3 & 8 \\
        \midrule
        \multicolumn{11}{l}{\textit{3D Experiments}} \\
        3dsw    & 4  & 8  & 12 & 2 & 3 & 4 & 1 & 2 & 3 & 8 \\
        \bottomrule
    \end{tabular}
\end{table}

\begin{table}[htbp]
    \centering
    \caption{\textbf{Comprehensive model statistics for OFormer.} The table outlines the resource requirements across three model sizes (Tiny, Medium, and Large) evaluated on various physical simulation benchmarks (1D Ks, 2D Darcy, 2D Ns, and 3D Sw). Metrics reported include the total number of parameters (\textbf{P}), memory consumption (\textbf{M}), and computational complexity in FLOPs (\textbf{F}). Numerical values are abbreviated for clarity: K (thousands), M (millions), G (billions), and MiB (mebibytes).}
    \label{tab:oformer_stats}
    \begin{tabular}{l ccc ccc ccc}
        \toprule
        \multirow{2}{*}{Benchmark} & \multicolumn{3}{c}{Tiny} & \multicolumn{3}{c}{Medium} & \multicolumn{3}{c}{Large} \\
        \cmidrule(lr){2-4} \cmidrule(lr){5-7} \cmidrule(lr){8-10}
        & P (K) & M (MiB) & F (M) & P (K) & M (MiB) & F (M) & P (M) & M (MiB) & F (G) \\
        \midrule
        1D Ks    & 11.6 & 904  & 31.2  & 92.0  & 2698  & 239.9 & 1.4  & 11224 & 3.5 \\
        2D Darcy & 11.6 & 876  & 29.1  & 91.9  & 2588  & 224.6 & 1.7  & 12290 & 4.1 \\
        2D Ns    & 3.0  & 7098 & 134.7 & 11.6  & 13618 & 498.7 & 0.03 & 24940 & 1.4 \\
        3D Sw    & 3.1  & 3720 & 68.2  & 11.7  & 6988  & 251.0 & 0.03 & 12582 & 0.71 \\
        \bottomrule
    \end{tabular}
\end{table}

\subsubsection{GNOT}
\label{sec:gnot}
The General Neural Operator Transformer (GNOT) is designed to serve as a scalable and flexible transformer-based framework for learning solution operators of PDEs \cite{hao2023gnot}. Unlike traditional numerical solvers that require specific discretization schemes, GNOT leverages the attention mechanism to handle irregular meshes, multiple input functions, and multi-scale problems simultaneously. The core innovation of GNOT lies in its \textit{Heterogeneous Normalized Attention} (HNA) block, which provides a general encoding interface for different input functions while maintaining linear complexity with respect to the sequence length.

GNOT adopts an architecture inspired by MIONet \cite{jin2022mionet}, consisting of a trunk network and multiple branch networks, followed by elaborated attention blocks and a decoder MLP. The trunk network encodes query point coordinates into latent embeddings, while branch networks independently embed distinct input components (e.g., boundary shapes, source functions, global parameters) into context embeddings. To address multi-scale problems, GNOT introduces a geometric gating mechanism based on mixture-of-experts (MoE), which can be viewed as a soft domain decomposition. This design allows the model to adaptively weight different expert networks based on the geometric coordinates of input points, effectively handling physically distinct subdomains within the same problem.

The GNOT architecture consists of input encoding through trunk and branch networks, core interaction through heterogeneous normalized attention blocks with geometric gating, and output decoding via an MLP. The pipeline can be decomposed into input encoding, core interaction, and output decoding.
Given an input function discretized on a mesh $\mathcal{X} := \{x_j \in \mathcal{D}\}_{1 \leq j \leq N}$, the model first encodes the input features into high-dimensional latent spaces through separate networks. The trunk network $\phi_{\text{trunk}}(\cdot)$ takes the coordinates of query locations and produces embeddings $z_0 \in \mathbb{R}^{N \times d}$:
\begin{equation}
    z_0 = \text{MLP}_{\text{trunk}}(x_{\text{coord}}),
\end{equation}
where $d$ denotes the hidden dimension. Multiple branch networks $\phi_{\text{branch}}^{(i)}(\cdot)$ independently embed different input components $c^{(i)}$ (e.g., boundary conditions, source terms, global parameters) into context embeddings $e^{(i)} \in \mathbb{R}^{M_i \times d}$:
\begin{equation}
    e^{(i)} = \text{MLP}_{\text{branch}}^{(i)}(c^{(i)}), \quad i = 1, \dots, K,
\end{equation}
where $K$ is the number of different input types and $M_i$ is the number of points for the $i$-th input component. To capture high-frequency components, coordinates are processed using horizontal Fourier embedding before being fed into the trunk network.
The core of GNOT is the stacking of $L$ MIOE (Multiple Input Output Expert) Cross Attention blocks. Each block updates the trunk embeddings by fusing information from branch embeddings through heterogeneous normalized cross-attention, followed by self-attention and geometric gating:

\textbf{Heterogeneous Normalized Cross-Attention.} The cross-attention layer aggregates information from multiple branch embeddings into the trunk embedding. For query $Q$ from trunk and keys/values $K^{(i)}, V^{(i)}$ from branches, the linear cross-attention is computed as:
\begin{equation}
    \text{CrossAttn}(Q, \{K^{(i)}, V^{(i)}\}) = \sum_{i=1}^K \frac{(\phi(Q) \cdot \phi(K^{(i)})^\top) V^{(i)}}{\sum_{j=1}^K \phi(Q) \cdot \phi(K^{(j)})^\top},
\end{equation}
where $\phi(\cdot) = \text{softmax}(\cdot)$ serves as the kernel feature map. This linear attention mechanism reduces the quadratic complexity $O(N^2)$ to linear complexity $O(N)$ by avoiding explicit materialization of the attention matrix.

\textbf{Self-Attention.} After cross-attention, a linear self-attention layer captures long-range spatial dependencies within the trunk embedding itself:
\begin{equation}
    \text{SelfAttn}(Z) = \frac{(\phi(Z) \cdot \phi(Z)^\top) Z}{\phi(Z) \cdot \phi(Z)^\top},
\end{equation}
where $Z$ represents the trunk embeddings after cross-attention fusion.

\textbf{Geometric Gating (MoE).} To handle multi-scale problems, each block incorporates a geometric gating mechanism that computes weighted averages of multiple expert MLPs. Given geometric coordinates $x_{\text{coord}}$, the gating network produces scores $G(x_{\text{coord}}) \in \mathbb{R}^{N_{\text{experts}}}$:
\begin{equation}
    p_i(x) = \frac{\exp(G_i(x))}{\sum_{j=1}^{N_{\text{experts}}} \exp(G_j(x))}, \quad i = 1, \dots, N_{\text{experts}},
\end{equation}
where $p_i(x)$ are the weights for averaging $N_{\text{experts}}$ expert networks. The updated embedding is:
\begin{equation}
    z_{\text{out}} = z_{\text{in}} + \sum_{i=1}^{N_{\text{experts}}} p_i(x) \cdot \text{MLP}_{\text{expert}}^{(i)}(z_{\text{in}}),
\end{equation}
where each expert MLP has an inner dimension double the embedding dimension.

After passing through $L$ attention blocks, the final trunk embeddings $z_L$ are projected to the target domain dimension through an output MLP:
\begin{equation}
    \hat{u} = \text{MLP}_{\text{out}}(z_L),
\end{equation}
where $\hat{u}$ represents the predicted solution field. For time-dependent problems, the model employs a sequence-to-sequence non-autoregressive strategy, predicting the full trajectory of future temporal solutions simultaneously to enhance memory efficiency.

We evaluate GNOT across multiple benchmarks with varying model sizes denoted as Tiny (T), Medium (M), and Large (L). The hyperparameter settings are detailed in Table \ref{tab:gnot_settings}.
A critical implementation detail concerns the GPU memory requirement. Each attention block in GNOT comprises both a self-attention and a cross-attention module, effectively doubling the memory footprint relative to a standard single-pathway attention layer. Consequently, to balance representational capacity with computational feasibility under constrained GPU budgets, the number of attention blocks (depth) is primarily selected from $\{1, 2, 4\}$ for benchmarks with moderate complexity (e.g., 1D Ks, 2D Darcy). For more memory-intensive tasks such as 2D Navier-Stokes and 3D Shallow Water, we adopt conservative depth configurations $\{1, 2, 3\}$ for the Large variant to ensure stable training without exceeding hardware limits.
The number of experts is uniformly fixed at 4 across all benchmarks and model scales, consistent with the original GNOT formulation. The inner dimension within each expert is set to twice the embedding dimension (multiplier = 2), preserving expressive capacity while maintaining parameter efficiency. The number of MLP layers follows a progressive scaling strategy: $1/2/3$ layers for Tiny/Medium/Large configurations, respectively. The MLP hidden dimension is likewise set to twice the embedding dimension, aligning with the maximum embedding scales employed in other baseline architectures to ensure comparability.
Architectural hyperparameters exhibit benchmark-dependent adaptation. For instance, attention heads follow a $\{2, 4, 8\}$ progression for 1D Ks and 2D Darcy, but are reduced to $\{1, 2, 2\}$ for 2D Navier-Stokes and 3D Shallow Water to accommodate their higher spatial resolution and temporal complexity. Similarly, hidden dimensions vary substantially across tasks: while 1D Ks Large employs a hidden dimension of 72, the corresponding configuration for 2D Navier-Stokes uses only 10, reflecting a deliberate trade-off between model capacity and the computational demands of chaotic flow prediction.
The comprehensive model statistics, including parameter count, memory consumption, and computational complexity (FLOPs), are reported in Table \ref{tab:gnot_stats}. Notably, GNOT maintains linear complexity with respect to the number of spatial points due to its linear attention mechanism, enabling efficient scaling to large-scale discretizations compared to standard softmax attention. However, the dual-pathway architecture (trunk + branches) incurs non-negligible memory overhead: for example, the Large variant on 2D Navier-Stokes requires $23.8$\,GiB of memory despite having only $0.02$\,M parameters, underscoring the activation storage costs inherent to long-horizon trajectory forecasting.

\begin{table}[htbp]
    \centering
    \caption{\textbf{Hyperparameter settings for GNOT.} Model sizes are denoted as T (Tiny), M (Medium), and L (Large).}
    \label{tab:gnot_settings}
    \begin{tabular}{l ccc ccc ccc ccc cc}
        \toprule
        \multirow{2}{*}{Dataset} & \multicolumn{3}{c}{Hidden Dim.} & \multicolumn{3}{c}{Depth} & \multicolumn{3}{c}{MLP Layers} & \multicolumn{3}{c}{Heads} & \multirow{2}{*}{N Experts} & \multirow{2}{*}{N Inner} \\
        \cmidrule(lr){2-4} \cmidrule(lr){5-7} \cmidrule(lr){8-10} \cmidrule(lr){11-13}
        & T & M & L & T & M & L & T & M & L & T & M & L & & \\
        \midrule
        \multicolumn{15}{l}{\textit{1D Experiments}} \\
        1dks    & 16 & 24 & 72 & 1 & 2 & 4 & 1 & 2 & 3 & 2 & 4 & 8 & 4 & 2 \\
        \midrule
        \multicolumn{15}{l}{\textit{2D Experiments}} \\
        2ddarcy & 12 & 32 & 64 & 1 & 2 & 4 & 1 & 2 & 3 & 2 & 4 & 8 & 4 & 2 \\
        2dns    & 8  & 8  & 10 & 1 & 2 & 3 & 1 & 2 & 3 & 1 & 2 & 2 & 4 & 2 \\
        \midrule
        \multicolumn{15}{l}{\textit{3D Experiments}} \\
        3dsw    & 8  & 8  & 10 & 1 & 2 & 3 & 1 & 2 & 3 & 1 & 2 & 2 & 4 & 2 \\
        \bottomrule
    \end{tabular}
\end{table}

\begin{table}[htbp]
    \centering
    \caption{\textbf{Comprehensive model statistics for GNOT.} The table outlines the resource requirements across three model sizes (Tiny, Medium, and Large) evaluated on various physical simulation benchmarks (1D Ks, 2D Darcy, 2D Ns, and 3D Sw). Metrics reported include the total number of parameters (\textbf{P}), memory consumption (\textbf{M}), and computational complexity in FLOPs (\textbf{F}). Numerical values are abbreviated for clarity: K (thousands), M (millions), G (billions), and MiB (mebibytes).}
    \label{tab:gnot_stats}
    \begin{tabular}{l ccc ccc ccc}
        \toprule
        \multirow{2}{*}{Benchmark} & \multicolumn{3}{c}{Tiny} & \multicolumn{3}{c}{Medium} & \multicolumn{3}{c}{Large} \\
        \cmidrule(lr){2-4} \cmidrule(lr){5-7} \cmidrule(lr){8-10}
        & P (K) & M (MiB) & F (M) & P (K) & M (MiB) & F (M) & P (M) & M (MiB) & F (G) \\
        \midrule
        1D Ks    & 14.1 & 884  & 34.6  & 59.2  & 3008  & 147.8 & 0.99 & 14630 & 2.5 \\
        2D Darcy & 8.0  & 804  & 18.3  & 103.5 & 3142  & 243.7 & 0.79 & 11544 & 1.9 \\
        2D Ns    & 3.8  & 7504 & 145.8 & 7.3   & 13988 & 279.5 & 0.02 & 23816 & 0.63 \\
        3D Sw    & 3.9  & 3922 & 74.7  & 7.4   & 6972  & 141.6 & 0.02 & 12280 & 0.32 \\
        \bottomrule
    \end{tabular}
\end{table}

\subsubsection{DynFormer}
We evaluate DynFormer across multiple benchmarks with varying model sizes denoted as Tiny (T), Medium (M), and Large (L). The hyperparameter settings are detailed in Table \ref{tab:dynformer_settings}.
A distinguishing architectural feature of DynFormer is its hybrid spectral-attention formulation, wherein global dependencies are captured via truncated spectral modes while local dynamics are modeled through attention mechanisms. The spectral truncation strategy is adapted to input dimensionality: for the 1D Kolmogorov flow benchmark, we retain $[64, 1]$ Fourier modes to resolve fine-scale temporal oscillations along the periodic domain; for 2D and 3D benchmarks (Darcy flow, Navier-Stokes, Shallow Water), an isotropic truncation $[12, 12]$ is applied to balance spectral expressiveness with computational tractability. 
Architectural hyperparameters exhibit deliberate benchmark-dependent scaling. Hidden dimensions follow a progressive schedule for most tasks (e.g., $24/32/128$ for 2D Darcy), yet adopt conservative configurations for chaotic regimes: the 2D Navier-Stokes Large model employs a hidden dimension of $32$ (identical to Medium) to mitigate overfitting under limited trajectory data, while the 3D Shallow Water Large configuration uses $24$ to accommodate the increased cost of 3D spectral transforms. Depth and attention heads are scaled jointly to preserve the capacity-to-cost ratio: while 1D Ks and 2D Darcy follow a $\{2, 4, 8\}$ head progression for Large models, the 2D Navier-Stokes and 3D Shallow Water benchmarks adopt a reduced $\{2, 2, 4\}$ schedule to constrain the quadratic complexity of spatial attention. Notably, the Medium 2D Darcy configuration employs depth $=6$—intermediate between Tiny ($2$) and Large ($8$)—to handle its heterogeneous permeability field without incurring the full cost of the deepest architecture.
The comprehensive model statistics, including parameter count, memory consumption, and computational complexity (FLOPs), are reported in Table \ref{tab:dynformer_stats}. DynFormer exhibits a distinctive efficiency profile characterized by low parameter counts but elevated memory requirements: for instance, the Large variant on 2D Navier-Stokes comprises only $0.11$\,M parameters yet consumes $24.6$\,GiB of memory, reflecting the activation storage overhead inherent to spectral transformations and long-horizon unrolling. Computational complexity scales sub-quadratically with spatial resolution due to the spectral attention decomposition, enabling feasible inference on high-resolution meshes: the Large 2D Darcy model achieves $3.1$\,G FLOPs with $3.4$\,M parameters, representing a favorable trade-off for engineering-scale surrogate modeling. These characteristics position DynFormer as a parameter-efficient architecture for physics-informed learning, albeit with memory demands that necessitate careful hardware-aware configuration for large-scale chaotic forecasting tasks.

\begin{table}[htbp]
    \centering
    \caption{\textbf{Hyperparameter settings for DynFormer.} Model sizes are denoted as T (Tiny), M (Medium), and L (Large).}
    \label{tab:dynformer_settings}
    \begin{tabular}{l ccc ccc ccc c}
        \toprule
        \multirow{2}{*}{Dataset} & \multicolumn{3}{c}{Hidden Dim.} & \multicolumn{3}{c}{Depth} & \multicolumn{3}{c}{Heads} & \multirow{2}{*}{Spectral Modes} \\
        \cmidrule(lr){2-4} \cmidrule(lr){5-7} \cmidrule(lr){8-10}
        & T & M & L & T & M & L & T & M & L & \\
        \midrule
        \multicolumn{11}{l}{\textit{1D Experiments}} \\
        1dks    & 4  & 8  & 32  & 2 & 3 & 6 & 2 & 4 & 8 & [64, 1] \\
        \midrule
        \multicolumn{11}{l}{\textit{2D Experiments}} \\
        2ddarcy & 24 & 32 & 128 & 2 & 6 & 8 & 2 & 4 & 8 & [12, 12] \\
        2dns    & 16 & 32 & 32  & 2 & 2 & 4 & 2 & 2 & 4 & [12, 12] \\
        \midrule
        \multicolumn{11}{l}{\textit{3D Experiments}} \\
        3dsw    & 8  & 20 & 24  & 2 & 2 & 3 & 2 & 2 & 4 & [12, 12] \\
        \bottomrule
    \end{tabular}
\end{table}

\begin{table}[htbp]
    \centering
    \caption{\textbf{Comprehensive model statistics for DynFormer.} The table outlines the resource requirements across three model sizes (Tiny, Medium, and Large) evaluated on various physical simulation benchmarks (1D Ks, 2D Darcy, 2D Ns, and 3D Sw). Metrics reported include the total number of parameters (\textbf{P}), memory consumption (\textbf{M}), and computational complexity in FLOPs (\textbf{F}). Numerical values are abbreviated for clarity: K (thousands), M (millions), G (billions), and MiB (mebibytes).}
    \label{tab:dynformer_stats}
    \begin{tabular}{l ccc ccc ccc}
        \toprule
        \multirow{2}{*}{Benchmark} & \multicolumn{3}{c}{Tiny} & \multicolumn{3}{c}{Medium} & \multicolumn{3}{c}{Large} \\
        \cmidrule(lr){2-4} \cmidrule(lr){5-7} \cmidrule(lr){8-10}
        & P (K) & M (MiB) & F (M) & P (K) & M (MiB) & F (M) & P (M) & M (MiB) & F (G) \\
        \midrule
        1D Ks    & 1.2  & 962  & 2.1   & 5.9   & 2776  & 10.3  & 0.16 & 12848 & 0.29 \\
        2D Darcy & 32.6 & 726  & 32.7  & 162.9 & 2362  & 151.8 & 3.4  & 12676 & 3.1 \\
        2D Ns    & 15.1 & 6894 & 263.6 & 57.9  & 13206 & 999.1 & 0.11 & 24578 & 1.8 \\
        3D Sw    & 7.0  & 3184 & 51.9  & 39.7  & 7068  & 274.7 & 0.08 & 11850 & 0.55 \\
        \bottomrule
    \end{tabular}
\end{table}

\newpage
\subsection{Visualization of baselines across benchmarks}
\subsubsection{1D Kuramoto-Sivashinsky}
\begin{figure}[htbp]
    \centering
    \includegraphics[width=0.98\linewidth]{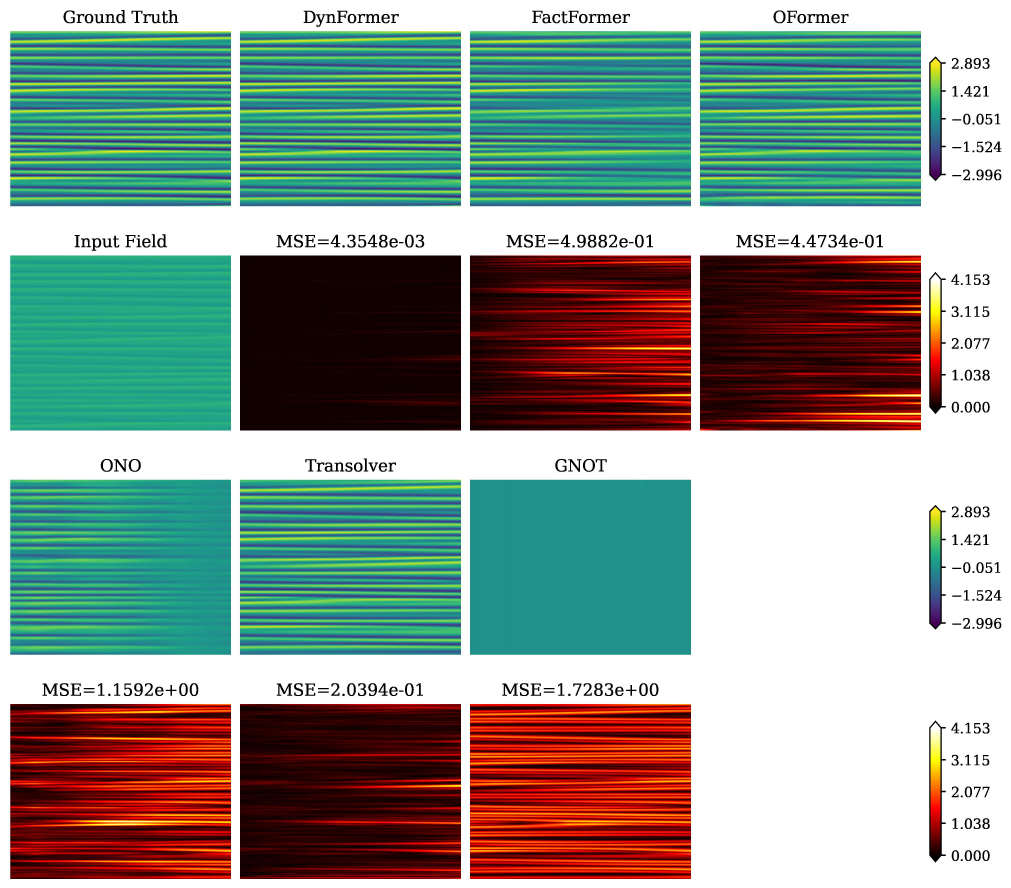}
    \caption{Visual comparison on 1D Kuramoto-Sivashinsky for the first sample. Each panel displays the evolution of the field $u(x,t)$ over space (horizontal axis) and time (vertical axis). }
    \label{fig:1dks_batch0}
\end{figure}

\begin{figure}[htbp]
    \centering
    \includegraphics[width=0.98\linewidth]{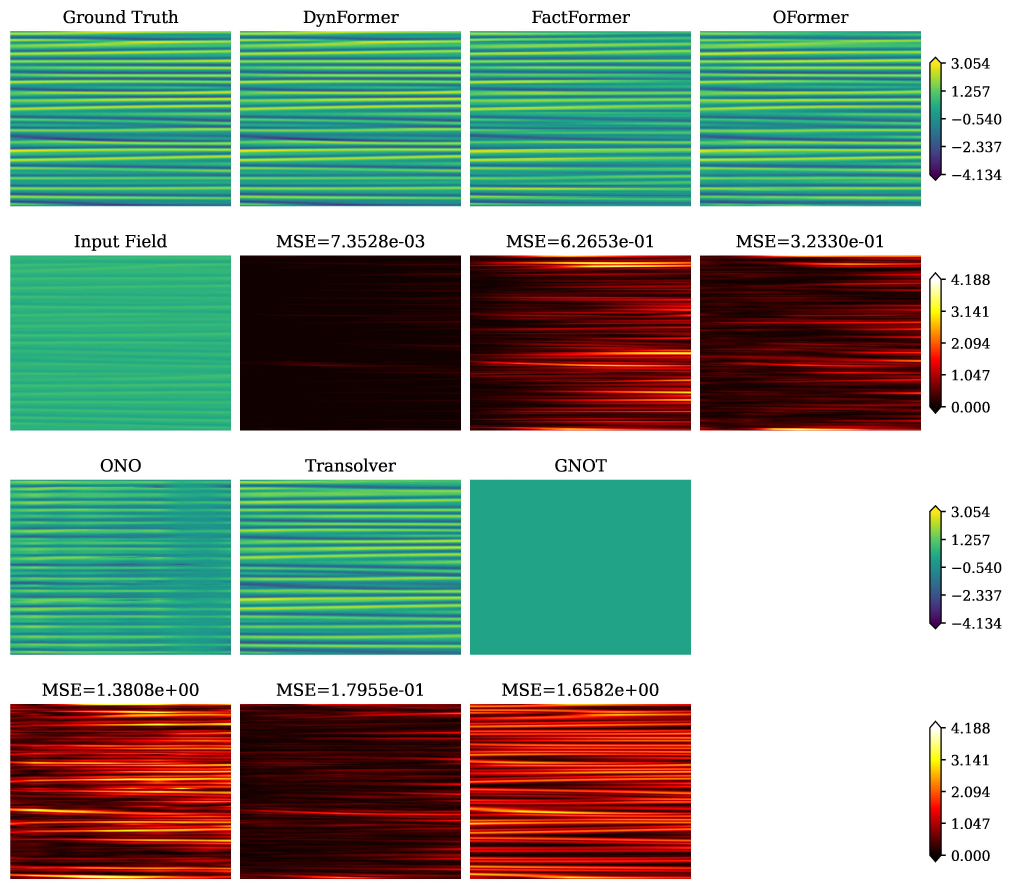}
    \caption{Visual comparison on 1D Kuramoto-Sivashinsky for the last sample. Each panel displays the evolution of the field $u(x,t)$ over space (horizontal axis) and time (vertical axis). }
    \label{fig:1dks_batch1}
\end{figure}

The visualizations of all baselines and DynFormer for the field $u(x,t)$ at the prediction horizon of the first and last samples in the test dataset are shown in Fig.~\ref{fig:1dks_batch0} and Fig.~\ref{fig:1dks_batch1}. Each figure organizes the comparison in a grid layout with Ground Truth and DynFormer in the top row, and FactFormer, Input Field, ONO, Transolver, and GNOT in the bottom row. The vertical axis represents the spatial coordinate $x$ over the domain $L=64\pi$, while the horizontal axis represents temporal evolution. All panels share a consistent color mapping to enable direct visual comparison of field magnitudes.
From a physical perspective, the Kuramoto-Sivashinsky equation governs spatiotemporal chaos arising from the competition between destabilizing negative diffusion ($u_{xx}$), stabilizing hyperdiffusion ($u_{xxxx}$), and nonlinear energy transfer ($u u_x$). The ground-truth solutions exhibit characteristic cellular structures with sharp gradients and irregular wave patterns—hallmarks of fully developed spatiotemporal chaos in dissipative systems. These structures emerge from the balance between large-scale energy injection and small-scale dissipation, producing complex, turbulent-like dynamics that challenge long-term predictive stability.

DynFormer produces output fields that retain the spatial topology of the ground truth across both samples (MSE = $4.35\times10^{-3}$ and $7.35\times10^{-3}$, respectively). The characteristic cellular structures and wave propagation patterns are visually preserved, with sharp gradients and irregular oscillations matching the ground-truth morphology. In contrast, GNOT (MSE = $1.73$ and $1.66$) exhibit substantial structural degradation, with smoothed or distorted patterns that fail to capture the fine-grained chaotic fluctuations inherent in the KS dynamics.
The error maps reveal that baseline models concentrate errors precisely along regions of steep spatial gradients and wavefront interactions. ONO (MSE = $1.16$ and $1.38$) and GNOT show pronounced smearing of sharp features, suggesting inadequate resolution of the hyperdiffusive term $u_{xxxx}$ which governs small-scale dissipation. DynFormer's predictions maintain crisp gradient transitions, indicating superior capture of the fourth-order stabilization mechanism that dampens high-frequency oscillations in the physical system.
Both samples demonstrate that DynFormer maintains temporal coherence throughout the 10-step prediction window. The wave propagation direction and speed align with ground-truth trajectories, reflecting accurate modeling of the nonlinear advection term $u u_x$ responsible for convective momentum transport. Transolver (MSE = $2.04\times10^{-1}$ and $1.80\times10^{-1}$) shows moderate temporal fidelity but exhibits phase drift and amplitude decay, while other baselines display complete loss of temporal structure by the prediction horizon.
The performance gap between DynFormer and baselines remains consistent across the first sample (Fig.~\ref{fig:1dks_batch0}) and last sample (Fig.~\ref{fig:1dks_batch1}), despite different initial conditions from the $\mathcal{U}(-1,1)$ distribution. DynFormer's MSE remains within the same order of magnitude ($\sim\!10^{-3}$), whereas baseline errors vary by up to one order of magnitude. This suggests DynFormer generalizes robustly across the chaotic attractor landscape rather than overfitting to specific trajectory patterns.

In summary, the figures demonstrate that DynFormer achieves superior visual fidelity in capturing the spatiotemporal chaos characteristic of the Kuramoto-Sivashinsky system. DynFormer's superior visual fidelity—particularly in preserving sharp gradient structures and maintaining temporal coherence—is physically consistent with accurate resolution of the competing energy injection, nonlinear transfer, and hyperdiffusive dissipation mechanisms that define the KS equation's chaotic dynamics. The substantial error reduction (2--3 orders of magnitude) compared to baselines indicates that DynFormer effectively learns the underlying physical operators governing this stiff, fourth-order nonlinear PDE.

\subsubsection{2D Darcy Flow}
\begin{figure}[htbp]
    \centering
    \includegraphics[width=0.98\linewidth]{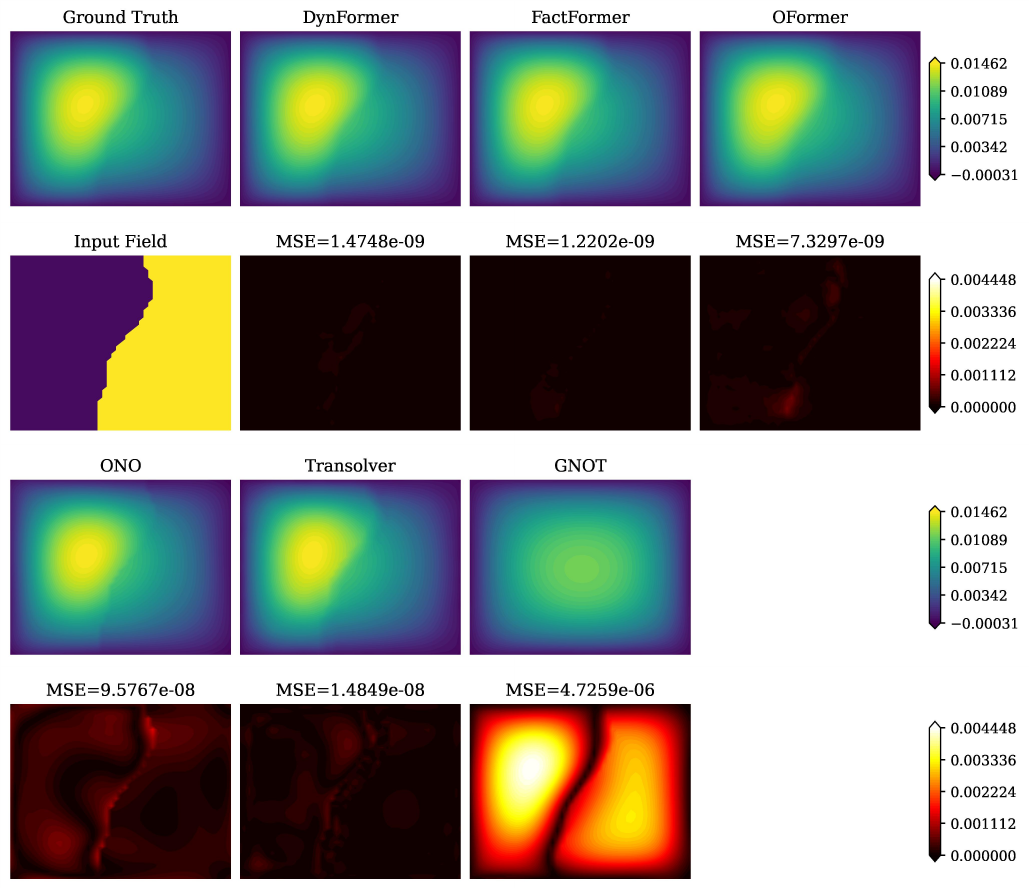}
    \caption{Visual comparison on 2D Darcy for the first sample. Each panel displays a 2D scalar field (e.g., pressure or concentration) over a square domain}
    \label{fig:2ddarcy_batch0}
\end{figure}

\begin{figure}[htbp]
    \centering
    \includegraphics[width=0.98\linewidth]{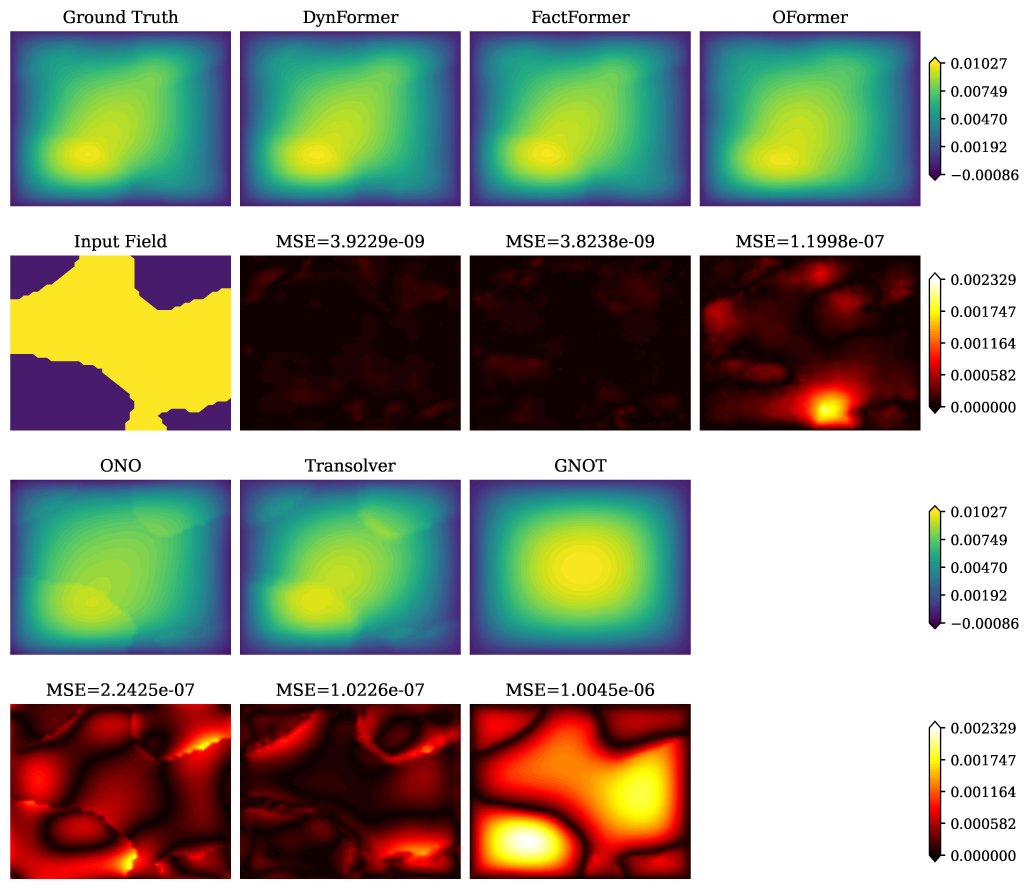}
    \caption{Visual comparison on 2D Darcy for the last sample. Each panel displays a 2D scalar field (e.g., pressure or concentration) over a square domain}
    \label{fig:2ddarcy_batch1}
\end{figure}

The visualizations of all baselines and DynFormer for the steady-state pressure field at the final equilibrium state of the first and last samples in the test dataset are shown in Fig. \ref{fig:2ddarcy_batch0} and Fig. \ref{fig:2ddarcy_batch1}.
Each figure is organized to present the ground-truth pressure field alongside the predictions from DynFormer, FactFormer, and OFormer in the top row. The second row displays the binary-thresholded input permeability field on the far left, followed by the spatial Mean Squared Error (MSE) maps for DynFormer, FactFormer, and OFormer. The third row shows the prediction fields for ONO, Transolver, and GNOT, while the fourth row displays their corresponding MSE maps. The error maps use a consistent colormap to highlight spatial regions of high deviation.
From a physical perspective, the 2D Darcy flow describes the stationary flow of a Newtonian fluid through a highly heterogeneous porous medium, driven by a uniform source. The system represents a state of global equilibrium where mass conservation must be maintained across sharp material interfaces. The ground-truth solutions exhibit a generally smooth global pressure gradient that features localized, sharp transitions and flow channeling—characteristic of the fluid navigating the high-contrast permeability zones ($k=12$ vs $k=3$) dictated by the input field.

Models failing to resolve the sharp permeability interfaces exhibit large errors precisely along the boundaries of the distinct material zones. As seen in the bottom-row error maps of both Fig. \ref{fig:2ddarcy_batch0} and Fig. \ref{fig:2ddarcy_batch1}, GNOT demonstrate significant error concentrations that perfectly trace the complex, thresholded geometries of the input field. This indicates inadequate resolution of the localized flux continuity constraints inherent in the Darcy equation across sharp material contrasts.
DynFormer and FactFormer produce output fields that retain the spatial topology of the ground truth with exceptionally high fidelity (both achieving MSE on the order of $10^{-9}$). Their corresponding error maps are nearly uniformly dark across the entire domain, indicating that these models successfully capture both the global pressure equilibrium and the micro-adjustments required by the discontinuous permeability field without accumulating localized boundary errors.
OFormer and Transolver exhibit intermediate predictive performance (MSE on the order of $10^{-8}$ to $10^{-7}$). Their error maps reveal that deviations are not randomly distributed but are systematically concentrated in specific geometric sub-regions, particularly where the high-permeability channels narrow or change direction abruptly. This suggests these models capture the smooth global trend but struggle with the high-frequency spatial components dictated by the sharp material boundaries.

In summary, the figures demonstrate that successfully learning the Darcy flow operator requires accurately capturing both the global equilibrium state and the localized, sharp flow transitions caused by highly heterogeneous permeability. DynFormer's superior visual fidelity—particularly its near-zero error along complex material interfaces—is physically consistent with a rigorous adherence to local flux conservation and global mass balance for this highly constrained physical system.

\subsubsection{2D Navier-Stokes}
\begin{figure}[htbp]
    \centering
    \includegraphics[width=0.98\linewidth]{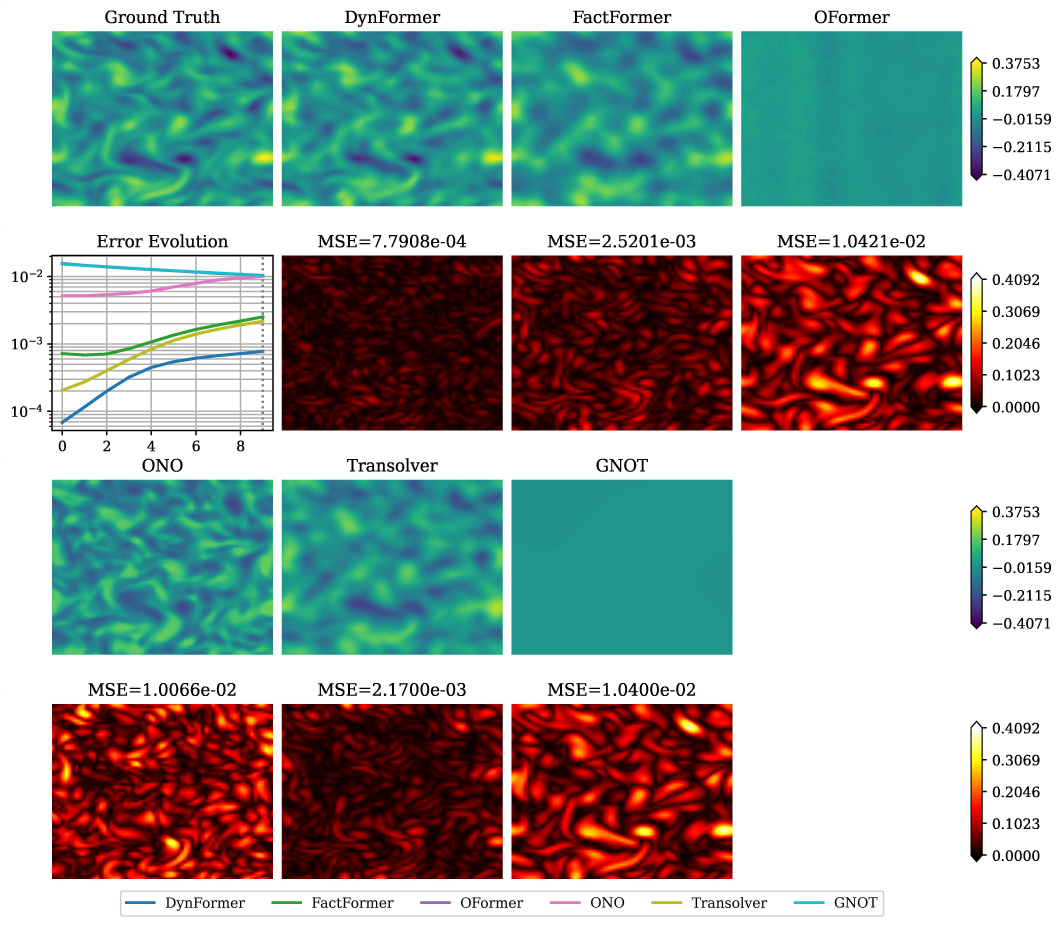}
    \caption{Visual comparison on 2D Navier-Stokes for the first sample at the final step. The vorticity variable $\omega$ is presented here.}
    \label{fig:2dns_batch0}
\end{figure}

\begin{figure}[htbp]
    \centering
    \includegraphics[width=0.98\linewidth]{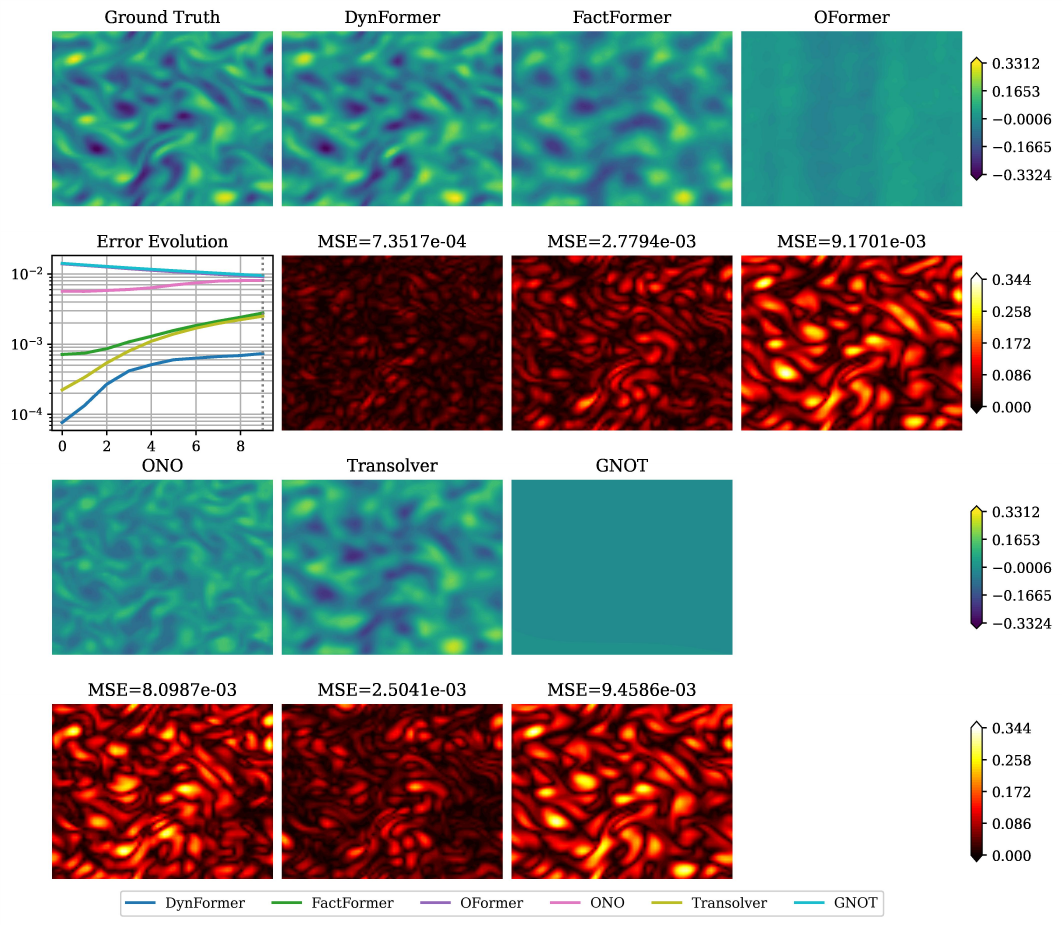}
    \caption{Visual comparison on 2D Navier-Stokes for the last sample at the final step. The vorticity variable $\omega$ is presented here.}
    \label{fig:2dns_batch1}
\end{figure}

The visualizations of all baselines and DynFormer for the scalar vorticity field $\omega$ at the final predicted time step (step 10) of the first and last samples in the test dataset are shown in Fig. \ref{fig:2dns_batch0} and Fig. \ref{fig:2dns_batch1}.
Each figure is organized to present the ground-truth vorticity field alongside the predictions from DynFormer, FactFormer, and OFormer in the top row. The second row displays the temporal Error Evolution plot over the rollout steps, followed by the spatial Mean Squared Error (MSE) maps for DynFormer, FactFormer, and OFormer. The third row shows the prediction fields for ONO, Transolver, and GNOT, while the fourth row displays their corresponding MSE maps.
From a physical perspective, the 2D Navier-Stokes equations model the evolution of incompressible, viscous flows. Operating in a high Reynolds number regime ($\nu = 10^{-5}$), the system's dynamics are heavily dominated by the nonlinear advection term rather than diffusion. The ground-truth solutions exhibit highly complex spatiotemporal behavior, characterized by the transfer of enstrophy, the formation of high-gradient vorticity filaments, and the interaction of coherent vortices—characteristic of nearly inviscid, two-dimensional turbulence.

Models that fail to accurately capture the nonlinear advection term exhibit severe artificial numerical dissipation, overwhelming the physical viscosity of the system. This is starkly visible in the predictions of OFormer and GNOT (Fig. \ref{fig:2dns_batch0} and Fig. \ref{fig:2dns_batch1}), which produce nearly featureless, homogenized output fields. Their corresponding error maps are uniformly saturated with high values ($MSE \sim 10^{-2}$), indicating a complete failure to maintain the flow's enstrophy and small-scale structures over the integration period.
FactFormer and Transolver exhibit stronger predictive capacity ($MSE \sim 2 \times 10^{-3}$), successfully capturing the macroscopic placement of the primary vortices. However, their spatial error maps reveal that deviations are heavily localized along the edges of the coherent structures and the thin filamentary regions. This physically suggests that while these models capture the general convective flow, they struggle to resolve the sharp, high-gradient phase boundaries dictated by the advective transport, leading to localized smearing of the vorticity field.
The Error Evolution plots in both figures demonstrate that temporal error accumulation is a major challenge for the baselines as the simulation rolls out. DynFormer uniquely maintains the lowest error trajectory across all forecast steps. Visually, DynFormer is the only model that preserves the crisp, high-frequency filamentary structures of the ground truth without introducing artificial smoothing. Its error maps ($MSE \sim 7 \times 10^{-4}$) remain uniformly dark, confirming that it accurately tracks both the position and intensity of the chaotic vortex interactions.

In summary, the figures demonstrate that forecasting high Reynolds number Navier-Stokes flow requires strict suppression of artificial numerical diffusion to prevent the smoothing of fine-scale turbulent structures. DynFormer's superior visual fidelity, particularly its ability to maintain high-gradient vorticity filaments and limit temporal error accumulation, is physically consistent with a highly accurate resolution of the nonlinear advection term dominating the fluid system.

\subsubsection{3D Shallow Water}
\begin{figure}[htbp]
    \centering
    \includegraphics[width=0.98\linewidth]{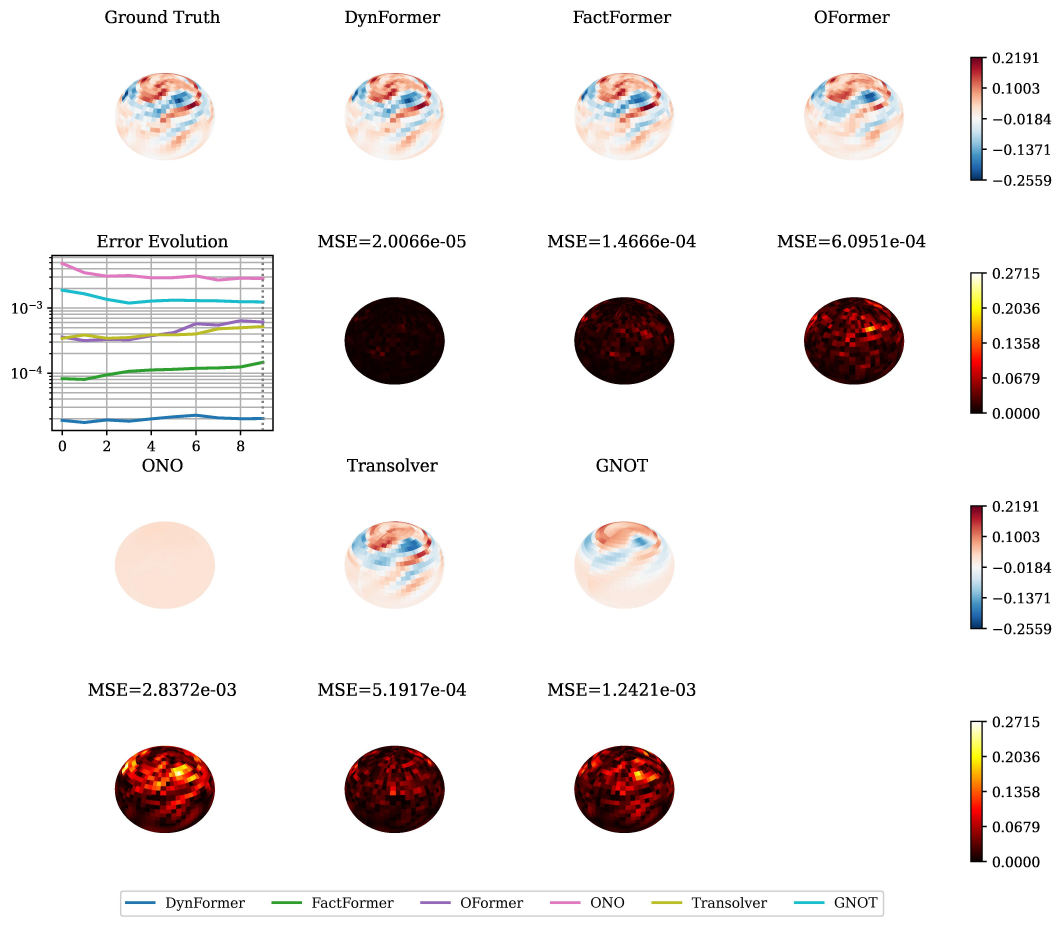}
    \caption{Visual comparison on 3D Shallow Water for the first sample at the final step. The vorticity variable $\omega$ is presented here.}
    \label{fig:3dsw_batch0}
\end{figure}

\begin{figure}[htbp]
    \centering
    \includegraphics[width=0.98\linewidth]{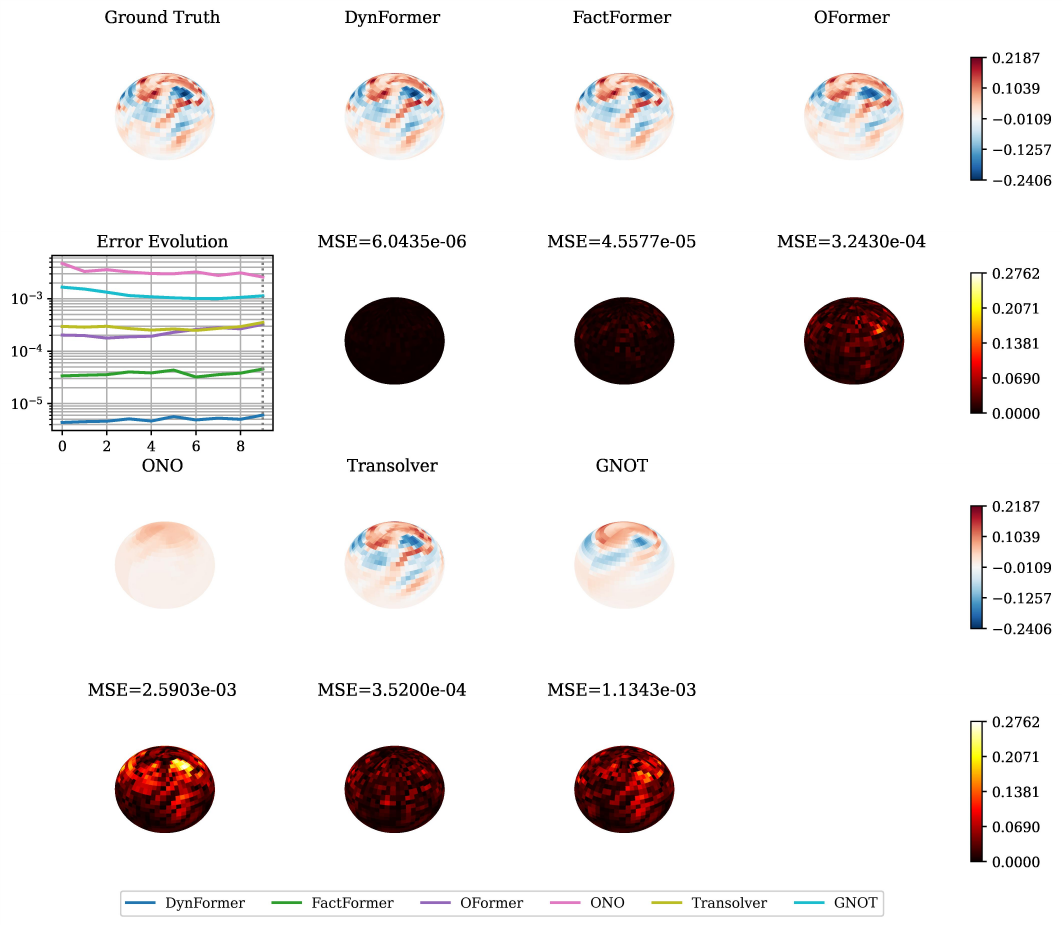}
    \caption{Visual comparison on 3D Shallow Water for the last sample at the final step. The vorticity variable $\omega$ is presented here.}
    \label{fig:3dsw_batch1}
\end{figure}

The visualizations of all baselines and DynFormer for the scalar vorticity field $\omega$ at the final predicted time step (step 10) of the first and last samples in the test dataset are shown in Fig. \ref{fig:3dsw_batch0} and Fig. \ref{fig:3dsw_batch1}.
Each figure is organized to present the ground-truth vorticity field alongside the predictions from DynFormer, FactFormer, and OFormer in the top row. The second row displays the temporal Error Evolution plot over the rollout steps, followed by the spatial Mean Squared Error (MSE) maps for DynFormer, FactFormer, and OFormer. The third row shows the prediction fields for ONO, Transolver, and GNOT, while the fourth row displays their corresponding MSE maps.
From a physical perspective, the Shallow Water equations model free-surface fluid flows subject to rotational and gravitational effects, serving as a foundational model for large-scale atmospheric and oceanic circulation. In this specific benchmark, the system dynamics are dictated by the geostrophic balance on a two-dimensional spherical manifold, where a localized perturbation triggers barotropic instability. This leads to the nonlinear transition from a stable zonal jet to a chaotic state characterized by breaking Rossby waves and coherent eddy formation. The ground-truth solutions exhibit this highly structured flow field, featuring distinct vortex patches, sharp rotational gradients, and complex filamentary interactions across the curvilinear domain.

Models that struggle with the complex advective dynamics and the spherical geometry, such as GNOT, ONO, and OFormer, exhibit severe artificial numerical dissipation. As seen in the prediction panels of Fig. \ref{fig:3dsw_batch0} and Fig. \ref{fig:3dsw_batch1}, their forecasted vorticity fields are heavily homogenized, failing to resolve the rotational structures of the breaking waves. Their corresponding error maps are highly saturated (with MSEs reaching up to $10^{-3}$), indicating a complete loss of the physical vortex dynamics over the integration period.
FactFormer and Transolver display intermediate predictive capabilities ($MSE \sim 10^{-4}$), successfully capturing the macroscopic locations of the primary swirling structures. However, their spatial error maps reveal that deviations are systematically concentrated along the outer edges and high-gradient boundaries of the vortices. This physically suggests that while these models capture the bulk fluid transport, they introduce localized smearing and fail to maintain the sharp gradients necessary for accurate eddy boundary resolution in the chaotic regime.
The temporal Error Evolution plots (second row, left) illustrate that most baselines suffer from progressive error accumulation as the jet breaks down over the rollout steps. DynFormer uniquely resists this trend, maintaining a flat, strictly bounded, and significantly lower error trajectory across all forecast steps. Visually, DynFormer is the only model that retains the crisp, high-frequency rotational filaments of the ground-truth vorticity field. Its nearly blank error maps ($MSE \sim 10^{-5}$) confirm that it accurately tracks the chaotic interactions of the shallow water eddies without succumbing to artificial diffusion.

In summary, the figures demonstrate that forecasting shallow water dynamics on a spherical manifold requires preserving complex rotational structures and strictly avoiding artificial numerical diffusion over long temporal horizons. DynFormer's superior visual fidelity—particularly its ability to track high-gradient vortex boundaries and maintain exceptional temporal stability during barotropic instability—is physically consistent with a highly accurate resolution of the nonlinear advective and geostrophic forces governing the planetary-scale system.

\subsection{Limitations and Discussion}
\subsubsection{Dependency on Uniform Grids}
A primary limitation of the current DynFormer implementation stems from its reliance on the parameterized Fourier transform to establish the spectral embedding. Because this approach fundamentally depends on the Fast Fourier Transform (FFT), it strictly requires uniform Euclidean grids. Consequently, the architecture encounters difficulties when applied directly to unstructured meshes or spatial domains with complex, irregular boundaries. To extend the model's applicability, future iterations could incorporate coordinate mapping techniques to transform irregular physical domains into regular computational grids prior to spectral embedding \cite{li2023fourier}.

\subsubsection{Performance Trade-offs in Linear, Steady-State Regimes}
While DynFormer is highly optimized for transient, multi-scale nonlinearities, its performance exhibits minor trade-offs when applied to strictly linear, steady-state physical regimes. Empirical evaluations, such as the 2D Darcy flow benchmark, indicate that DynFormer can be marginally outperformed by architectures specifically optimized for spatial factorizations (e.g., FactFormer). This behavior is a direct consequence of the model's architectural priors: the Local-Global-Mixing (LGM) module is explicitly designed to model the cross-frequency energy cascades characteristic of turbulent and chaotic systems. In purely elliptic PDEs governed by static global equilibrium, this nonlinear mixing mechanism introduces representational overhead without a proportional predictive benefit. A promising avenue for future optimization involves adapting the LGM module to dynamically bypass or attenuate nonlinear mixing when encountering purely linear physical systems.

\subsubsection{Comparison with Non-Transformer Neural Operators}
Under strict memory-aligned constraints, Transformer-based neural operators generally exhibit lower predictive accuracy compared to highly optimized non-Transformer architectures, such as the standard Fourier Neural Operator (FNO). This performance gap primarily stems from the inherent parameter overhead and representational footprint associated with self-attention mechanisms. However, drawing parallels to the evolutionary trajectory of Vision Transformers (ViTs) in computer vision \cite{dosovitskiy2020image}, the true representational advantage of Transformer-based architectures typically emerges only when trained on massive, large-scale datasets. Consequently, the empirical evaluations in this work deliberately focus on benchmarking DynFormer exclusively against other state-of-the-art Transformer-based models. This targeted comparison highlights DynFormer's superior scalability and dynamics-informed feature extraction within the Transformer paradigm. Ultimately, we position DynFormer not as an immediate replacement for lightweight solvers in data-scarce regimes, but rather as a highly scalable, powerful backbone candidate for future large-scale foundation models in scientific computing.

\subsubsection{More Physical Constraints Ensuring Separable Space Assumption}
A core theoretical pillar of DynFormer is the reduction of spatial complexity from $O(N^4)$ to $O(N^3)$ via Kronecker-structured attention. This efficiency relies heavily on the physical intuition that large-scale, smooth components of the PDE solution admit a separation of variables, thereby possessing a structured low-rank representation along coordinate axes. However, a current limitation is that this separability is treated as an architectural ansatz rather than a constrained objective. While the network is designed to process features assuming they are separable, the projection from the physical input space to the latent space $\mathbb{R}^{N_1 \times N_2 \times d_{\text{in}}} \rightarrow \mathbb{R}^{N_1 \times N_2 \times d_n}$ is learned freely without explicit penalties for non-separable structures.In regimes with highly anisotropic dynamics or complex diagonal couplings where the separation of variables does not strictly hold, the learned latent representation may deviate from the assumed Kronecker structure. Consequently, forcing these non-separable features through axis-wise factorized attention could result in information loss or approximation errors regarding off-axis correlations.To address this in future work, we propose incorporating explicit physical constraints to enforce the validity of the separable space assumption. This could be achieved by introducing a "separability regularization" term in the loss function, mathematically penalizing the high-rank components of the latent feature tensors via nuclear norm minimization \cite{liu2010interior, lu2019low} or maximizing alignment with rank-1 tensor decompositions \cite{shi2017tensor, cao2024learning}. By explicitly constraining the latent space to lie on a separable manifold, the model would ensure that the efficient Kronecker-structured attention mechanism is applied to features that rigorously satisfy the underlying mathematical prerequisites, thereby improving both theoretical consistency and predictive robustness in complex regimes.

\bibliographystyle{elsarticle-num}
\bibliography{references}
\newpage